\documentclass[twoside]{article}

\usepackage[accepted]{aistats2023}

\usepackage[utf8]{inputenc} 
\usepackage[T1]{fontenc}    
\usepackage{hyperref}       
\usepackage{url}            
\usepackage{booktabs}       
\usepackage{amsfonts}       
\usepackage{nicefrac}       
\usepackage{microtype}      
\usepackage{xcolor,pifont}        
\usepackage{amsthm}
\usepackage{graphicx}
\usepackage{subfigure}
\usepackage{amsmath}
\usepackage{amssymb}
\usepackage{bbm}
\usepackage{mathtools}
\usepackage{array,adjustbox}
\usepackage{natbib}
\usepackage{multirow}
\usepackage{caption}
\usepackage{changes}

\usepackage{algorithm,algorithmic}
\usepackage{cleveref}
\hypersetup{
    colorlinks,
    linkcolor={blue},
    citecolor={blue},
    urlcolor={blue}
}
\theoremstyle{plain}
\newtheorem{theorem}{Theorem}[section]
\newtheorem{proposition}[theorem]{Proposition}
\newtheorem{lemma}[theorem]{Lemma}

\newtheorem{definition}[theorem]{Definition}
\newtheorem{assumption}[theorem]{Assumption}
\newtheorem{remark}[theorem]{Remark}
\newcommand{\R}{DR_{p,\varepsilon}}

\renewcommand{\S}{\mathbb{S}^{d-1}}

\newcommand{\supS}{\underset{u\in \mathbb{S}^{d-1}}{\sup}}
\newcommand{\intR}{\int_{0}^{1-\varepsilon}}

\newcommand{\mbX}{{\mathbf{X}}}
\newcommand{\mbY}{{\mathbf{Y}}}
\newcommand{\mbM}{{\mathbf{M}}}
\newcommand{\reals}{\mathbb{R}}

\begin{document}

%

%

\twocolumn[\aistatstitle{A Pseudo-Metric between Probability Distributions \\ based on Depth-Trimmed Regions}

\aistatsauthor{ Guillaume Staerman$^{1}$ \And Pavlo Mozharovskyi$^{1}$ \And  Pierre Colombo$^{1,2}$ \AND Stéphan Clémencon$^{1}$ \And Florence d'Alché-Buc$^{1}$ }

\aistatsaddress{ $^{1}$LTCI, Télécom Paris, Institut Polytechnique de Paris \\ $^{2}$ IBM GBS France}  ]

\begin{abstract}
The design of a metric between probability distributions is a longstanding problem motivated by numerous applications in machine learning. Focusing on continuous probability distributions in the Euclidean space $\mathbb{R}^d$, we introduce a novel pseudo-metric between probability distributions by leveraging the extension of univariate quantiles to multivariate spaces. Data depth is a nonparametric statistical tool that measures the centrality of any element $x\in\mathbb{R}^d$ with respect to (w.r.t.) a probability distribution or a dataset. It is a natural median-oriented extension of the cumulative distribution function (cdf) to the multivariate case. Thus, its upper-level sets---the depth-trimmed regions---give rise to a definition of multivariate quantiles. The new pseudo-metric relies on the average of the  Hausdorff distance between the depth-based quantile regions w.r.t. each distribution. Its good behavior w.r.t. major transformation groups, as well as its ability to factor out translations, are depicted. Robustness, an appealing feature of this pseudo-metric, is studied through the finite sample breakdown point. Moreover, we propose an efficient approximation method with linear time complexity w.r.t. the size of the dataset and its dimension. The quality of this approximation, as well as the performance of the proposed approach, are illustrated in numerical experiments.
\end{abstract}

\section{INTRODUCTION}\label{intro}
Metrics or pseudo-metrics between probability distributions have attracted a long-standing interest in information theory \citep{kullback1959,renyi1961,csiszar,stummer2012}, probability theory and statistics \citep{billingsley1999,sriperumbudurIPM,panaretosW,rachev1991probability}. While they serve many purposes in machine learning \citep{cha2002,mackay2003}, they are of crucial importance in automatic evaluation of natural language generation (see e.g. \citealp{kusner2015word,zhang2019bertscore}), especially when leveraging deep contextualized embeddings such as the popular BERT \citep{bert}.
 Yet designing a measure to compare two probability distributions is a challenging research field. This is certainly due to the inherent difficulty in capturing in a single measure typical desired properties such as: (i) metric or pseudo metric properties, (ii) invariance under specific geometric transformations, (iii) efficient computation, and (iv) robustness to contamination. 

One can find in the literature a vast collection of discrepancies between probability distributions that rely on different principles. The $f$-divergences \citep{csiszar} are defined as the weighted average by a well-chosen function $f$ of the odds ratio between the two distributions. They are widely used in statistical inference but they are by design ill-defined when the supports of both distributions do not overlap, which appears to be a significant limitation in many applications. IPMs \citep{sriperumbudurIPM} are based on a variational definition of the metric, i.e. the maximum difference in expectation for both distributions calculated over a class of measurable functions and give rise to various metrics (Maximum Mean Discrepancy (MMD), Dudley's metric, $L_1$-Wassertein Distance) depending on the choice of this class. However, except in the case of MMD, which appears to enjoy a closed-form solution, the variational definition raises issues in computation. From the side of Optimal transport (OT) (see \citealp{Villani,Peyre}),  the $L_p$-Wasserstein distance is based on a ground metric able to take into account the geometry of the space on which the distributions are defined. Its ability to handle non-overlapping support and appealing theoretical properties make OT a powerful tool, mainly when applied to generative models \citep{arjovsky2017} or automatic text evaluation \citep{zhao2019moverscore,clark2019sentence,colombo2021automatic}.


%



In this work, we adopt another angle. Focusing on continuous probability distributions in the Euclidean space $\mathbb{R}^d$, we propose to consider a new metric between probability distributions by leveraging the extension of univariate quantiles to multivariate spaces. The notion of quantile function is an interesting ground to build a comparison between two probability measures as illustrated by the closed-form of the Wasserstein distance defined over $\reals$. However, given the lack of natural ordering on $\mathbb{R}^d$ as soon as $d>1$, extending the concept of univariate quantiles to the multivariate case raises a real challenge. Many extensions have been proposed in the literature, such as minimum volume sets \citep{mason}, spatial quantiles \citep{kol} or data depth \citep{Tukey75}. The latter offers different ways of ordering multivariate data w.r.t a probability distribution. Precisely, \textit{data depths} are non parametric statistics that determine the centrality of any element $x\in \mathbb{R}^d$ w.r.t. a probability measure. They provide a multivariate ordering based on topological properties of the distribution, allowing it to be characterized by its location, scale or shape (see e.g. \citealp{mosler} or \citealp{phdguigui} for a review). Several data depths were subsequently proposed such as convex hull peeling depth \citep{barnett}, simplicial depth \citep{liu1990}, Oja depth \citep{Oja} or zonoid depth \citep{koshevoy1997} differing in their properties and applications. With a substantial body of literature devoted to its computation, recent advances allow for fast exact~\citep{PokotyloMD19} and approximate~\citep{DyckerhoffMN20} computation of several depth notions. The desirable properties of data depth, such as affine invariance, continuity w.r.t. its arguments, and robustness \citep{ZuoSerfling00} make it an important tool in many fields. Today, in its variety of notions and applications, data depth constitutes a versatile methodology \citep{MoslerM21} that has been successfully employed in a variety of machine learning tasks such as regression \citep{rousseeuw1999,hallin2010}, classification \citep{LI,LangeMM14}, anomaly detection \citep{ser2006,rousseeuw2018,FIF,staerman2020,staerman2022functional} and clustering \citep{jornsten}.

This paper presents a new discrepancy measure between probability distributions, well-defined for non-overlapping supports,  that leverages the interesting features of data depths. This measure is studied through the lens of the previously stated properties, yielding the contributions listed below.

\vspace{0.8cm}
\noindent \textbf{Contributions: } 
\begin{itemize}
\item A new discrepancy measure between probability distributions involving the upper-level sets of data depth is introduced. We show that this measure defines a pseudo-metric in general. Its good behavior regarding major transformation groups, as well as its ability to factor out translations, are depicted. Its robustness is investigated through the concept of finite sample breakdown point.
\item An efficient approximation of the depth-trimmed regions-based pseudo-metric is proposed for convex depth functions such as halfspace, projection and integrated rank-weighted depths. This approximation relies on a nice feature of the Hausdorff distance when computed between convex bodies.

\item The behavior of this algorithm regarding its parameters is studied through numerical experiments, which also highlight the by-design robustness of the depth-trimmed regions based pseudo-metric. Applications to robust clustering of images and automatic evaluation of natural language generation (NLG) show the benefits of this approach when benchmarked with state-of-the-art probability metrics.

\end{itemize}

\section{BACKGROUND ON DATA DEPTH}\label{sec:background}

In this section, we recall the concept of statistical \textit{data depth} function and its attractive theoretical properties for clarity. 
Here and throughout, the space of all continuous probability measures on $\mathbb{R}^d$ with $d\in \mathbb{N}^*$ is denoted by $\mathcal{M}_1(\mathbb{R}^d)$. By $g_{\sharp}$ we denote the push-forward operator of the function $g$.
Introduced by \citet{Tukey75}, the concept of data depth initially extends the notion of median to the multivariate setting. In other words, it measures the centrality of any element $x\in \mathbb{R}^d$ w.r.t. a probability distribution (respectively, a dataset). Formally, a data depth is defined as follows:

\begin{equation}\label{equ:depthdef}
\begin{tabular}{lrcl}
$D:$ & $\mathbb{R}^d\times \mathcal{M}_1(\mathbb{R}^d)$ & $\longrightarrow$ & $[0,1]$\,,
\\ &  $( x, \; \rho)$ & $\longmapsto$ & $D(x,\; \rho)$.
\end{tabular}
\end{equation}

We denote by $D(x,\rho)$ (or $D_{\rho}(x)$ for brevity) the depth of  $x\in \mathbb{R}^d$ w.r.t. $\rho\in\mathcal{M}_1(\mathbb{R}^d)$. The higher $D(x,\rho)$, the deeper it is in $\rho$. The depth-induced median of $\rho$ is then defined by the set attaining $\sup_{x\in \mathbb{R}^d}\; D(x,\rho)$. Since data depth naturally and in a nonparametric way defines a pre-order on $\mathbb{R}^d$ w.r.t. a probability distribution, it can be seen as a centrality-based alternative to the cumulative distribution function (cdf) for multivariate data. For any $\alpha\in [0,1]$, the associated $\alpha$-depth region of a depth function is defined as its upper-level set:

\begin{equation*}
	D_{\rho}^{\alpha}=\left\{x\in \mathbb{R}^d,\,D_{\rho}(x)\geq \alpha\right\}\hspace*{-0.1cm}.
\end{equation*}

 It follows that depth regions are nested, i.e. $D_{\rho}^{\alpha'} \subseteq D_{\rho}^{\alpha}$  for any $\alpha<\alpha'$.
 These depth regions generalize the notion of quantiles to a multivariate distribution. 

A depth function's relevance to capturing information about a distribution relies on the statistical properties it satisfies. Such properties have been thoroughly investigated in \citet{liu1990,ZuoSerfling00} and \cite{rainer2004} with slightly different sets of axioms (or postulates) to be satisfied by a proper depth function. In this paper, we restrict to \emph{convex depth functions} \citep{rainer2004} mainly motivated by recent algorithmic developments including theoretical results \citep{NagyDM20} as well as implementation guidelines \citep{DyckerhoffMN20}.

The general formulation \eqref{equ:depthdef} opens the door to various possible definitions. While these differ in theoretical and practically related properties such as robustness or computational complexity (see \citealp{MoslerM21} for a detailed discussion), several postulates have been developed throughout the recent decades the ``good'' depth function should satisfy. Formally, a function $D$ is called a \emph{convex depth function} if it satisfies the following postulates:

\begin{itemize}
\item[\textbf{D1}] (\textsc{Affine invariance}) $D(g(x), g_{\sharp} \rho)=D(x,\rho)$ holds 
for $g: x\in \mathbb{R}^d \mapsto Ax+b$ with any non-singular matrix $A\in \mathbb{R}^{d\times d}$ and any vector $b\in \mathbb{R}^d$.\label{pr:d1}
\item[\textbf{D2}] (\textsc{Vanishing at infinity}) $ \underset{||x|| \rightarrow \infty}{\lim}D_{\rho}(x)=0$.
\item[\textbf{D3}] (\textsc{Upper semicontinuity}) $\left\{ x\in \mathbb{R}^d \; \; D_{\rho}(x)<\alpha\right\}$ is an open set for every $\alpha\in (0,1]$.
\item[\textbf{D4}] (\textsc{Quasiconcavity})  For every $\lambda \in [0,1]$ and  $x,y\in \mathbb{R}^d$, $D_{\rho}(\lambda x+(1-\lambda) y)\geq \min \{D_{\rho}(x),D_{\rho}(y) \}$. 
\end{itemize}

While (\textbf{D1}) is useful in applications providing independence w.r.t. measurement units and coordinate system, (\textbf{D2}) and (\textbf{D3}) appear as natural properties since data depth is a (center-outward) generalization of cdf. Limit values vanish due to median-oriented construction. (\textbf{D4}) allows to preserve the original center-outward ordering goal of data depth and induces convexity of the depth regions. Furthermore, it is easy to see that (\textbf{D1--D4}) respectively yield properties of affine equivariance, boundedness, closedness  and convexity  of the central regions $D_{\rho}^{\alpha}$  \citep{rainer2004}. 
Thanks to (\textbf{D2--D4}), if $\alpha>0$,  non-empty regions associated to convex depth functions are convex bodies (compact convex set in $\mathbb{R}^d$).

Below we recall three convex depth functions satisfying (\textbf{D1--D4}) that will be used throughout the paper: halfspace depth \citep{Tukey75}, which is probably the most studied in the literature, projection depth \citep{Liu92}, and the (affine-invariant) integrated rank-weighted (AI-IRW) depth \citep{IRW,AIIRW}.  For this, let $\mathbb{S}^{d-1}$ be the unit sphere in $\mathbb{R}^d$ and $X$ a random variable defined on a certain probability space $(\Omega,\mathcal{A},\mathbb{P})$ that takes values in $\mathcal{X}\subset \mathbb{R}^d$ following distribution $\rho$. The halfspace depth of a given $x\in \mathbb{R}^d$ w.r.t. $\rho$ is defined as the smallest probability mass that can be contained in a closed halfspace containing $x$:

\vspace*{-0.1cm}
\begin{equation*}
HD_{\rho}(x)=\underset{u\in \mathbb{S}^{d-1}}{\inf } \; \;\mathbb{P}\left(\langle u,X \rangle \leq \langle u,x \rangle \right)\hspace*{-0.05cm}.
\end{equation*}

Projection depth, being a monotone transform of the Stahel-Donoho outlyingness \citep{DonohoGasko,stahel}, is defined as follows:
\begin{align*}
	PD_{\rho}(x)&\hspace{-0.05cm}= \hspace{-0.05cm}\left( 1+ \hspace{-0.1cm} \underset{u\in \mathbb{S}^{d-1}}{\sup } \frac{|\langle u,x \rangle - \text{med}(\langle u,X \rangle)|}{\text{MAD}(\langle u,X \rangle)} \right)^{-1}\,, 
\end{align*}

\noindent where $\text{med}$ and $\text{MAD}$ stand for the univariate median and median absolute deviation from the median, respectively. The affine-invariant integrated rank-weighted  (AI-IRW) depth of a given $x\in \mathbb{R}^d$ relative to a square-integrable random vector $X$ with probability distribution $\rho$ on $\mathbb{R}^d$ and positive definite variance-covariance matrix $\Sigma$, named $AD_{\rho}(x)$, is given by:

\begin{align}\label{eq:AIIRW}
 \mathbb{E}\left[ \min \Big \{  \mathbb{P}\left(\langle V,X \rangle \leq \langle V,x \rangle \right),  \mathbb{P}\left(\langle V,X \rangle > \langle V,x \rangle \right) \Big \}\right], 
\end{align}

where $V=\Sigma^{{\scriptscriptstyle - \top/2}} U/\vert\vert \Sigma^{{\scriptscriptstyle - \top/2}} U \vert\vert$ and $U$ is uniformly distributed on the hypersphere $\mathbb{S}^{d-1}$. 

\begin{remark} Data depth functions have connections with the density function in particular cases. Indeed, for elliptical distributions, the level sets of any data depth satisfying (\textbf{D1--D4}) are concentric ellipsoids with the same center, and orientation as the density level sets \citep{LiuSingh}. The density is a local measure assigning the score of an element as the probability mass in an infinitesimal neighborhood. In contrast, data depths are global measures of ordering taking into account the whole distribution to assign a score to an element and are thus not equivalent to the density for general distributions. However,  they provide interesting alternatives in many applications, such as anomaly detection (see e.g. \citealp{AIIRW}). For example, the density will assign a zero score to every $x\in \mathbb{R}^d$ far from a concentrated group of observations regardless of the distance. At the same time, the projection depth described above will be able to rank these ‘‘outliers'' depending on how it moves away from them.
\end{remark}

\section{A PSEUDO-METRIC BASED ON DEPTH-TRIMMED REGIONS}
In this section, we introduce the depth-based pseudo-metric and study its properties. We consider depth regions possessing the same probability mass to compare depth regions from different probability distributions fairly.
 Following \citet{localdepth}, we denote by $\alpha : (\beta, \rho)\in [0,1] \times \mathcal{M}_1(\mathbb{R}^d) \longmapsto \alpha(\beta,\rho) \in [0,1] \ $ the highest level such that the probability mass of the depth-trimmed region at this level is at least $\beta$. Precisely, for any pair $(\beta,\rho) \in [0,1) \times \mathcal{M}_1(\mathbb{R}^d)$:

 \begin{equation}\label{alphadef}
     \alpha(\beta,\rho)=\sup \{ \gamma \in [0,1]: \; \; \rho\left( D^{\gamma}_{\rho} \right) > \beta \}.
 \end{equation}
 
 In the remainder of this paper,  when the quantity $\alpha (\beta, \rho)$ will be associated with depth regions of $\rho$, the second argument of the function $\alpha(\cdot,\cdot)$ will be omitted, for notation simplicity. It is worth mentioning that $D_{\rho}^{\alpha(\beta')} \subseteq D_{\rho}^{\alpha(\beta)}$  for any $\beta>\beta'$, since $\beta \mapsto \alpha(\beta, \rho)$ is a monotone decreasing function. Thus, $D_{\rho}^{\alpha(\beta)}$ is the smallest depth region with probability larger than or equal to $\beta$ and can be defined in an identical way as:
 \begin{align*}
     D_{\rho}^{\alpha(\beta)}= \underset{\gamma \in \Gamma_{\rho}(\beta)}{\bigcap} D_{\rho}^{\gamma},
 \end{align*}
 
 \vspace{-0.3cm}
\noindent where $\Gamma_{\rho}(\beta) = \{\zeta \in [0,1]: \; \; \rho \left(D^{\zeta}_{\rho} \right) > \beta \}$. The strict inequalities in (\ref{alphadef}) and in the definition of $\Gamma_{\rho}(\beta)$ eliminate cases where the supremum does not exist. Indeed, when $\beta=0$, the depth region is then an infinitesimal set with a probability higher than zero. To the best of our knowledge, the supremum exists (without necessarily being unique) in the case of the halfspace depth \citep{rousseeuwruts} and the projection depth \citep{zuo} under mild assumptions. Still, no results have been derived for AI-IRW depth yet. The set $ \{ D_{\rho}^{\alpha(\beta)}, \; \beta \in [0,1-\varepsilon], \; \varepsilon \in (0,1] \}$  where each region probability mass is equal to $\beta$  then defines quantile regions of $\rho$.

 
 
 

 Let $\mu,\nu$ be two absolutely continuous probability measures (w.r.t. the Lebesgue measure) on $\mathcal{X},\mathcal{Y} \subset \mathbb{R}^d$ respectively. Denote by $d_{\mathcal{H}}(A,B)$ the Hausdorff distance between the sets $A$ and $B$. The pseudo-metric between probability distributions $\mu$ and $\nu$ based on the depth-trimmed regions is defined as follows:

\begin{definition}\label{def2}
 Let $\varepsilon \in (0,1]$ and $p\in (0,\infty)$, for all pairs $(\mu,\nu)$ in $\mathcal{M}_1(\mathcal{X})\times \mathcal{M}_1(\mathcal{Y})$, the  depth-trimmed regions ($DR_{p,\varepsilon}$) discrepancy measure between  $\mu$ and $\nu$  is  defined as

\begin{align}
DR_{p,\varepsilon}^p(\mu,\nu) =  \int_{0}^{1-\varepsilon} d_{\mathcal{H}} \left( D_{\mu}^{\alpha(\beta)},D_{\nu}^{\alpha(\beta)} \right)^p~\mathrm{d}\beta.\label{DRW}
\end{align}
\end{definition}

 Our discrepancy measure relies on the Hausdorff distance averaged over depth-trimmed regions with the same probability mass w.r.t. each distribution. Properties (\textbf{D2--D3}) ensure that for every $0 \leq \beta < 1$, $D^{\alpha(\beta)}_{\mu} $ is a non-empty compact subset of $\mathbb{R}^d$ leading to a well-defined discrepancy measure. Observe that the parameter $\varepsilon$ can be considered as a robustness tuning parameter. Indeed, choosing higher $\varepsilon$ amounts to ignore the larger upper-level sets of data depth function, i.e. the tails of the distributions.
 



\begin{remark}
Data depths provide robustness to (\ref{DRW}) together with the $\varepsilon$-trimming. Indeed, data depths such as the three previously introduced in Section~\ref{sec:background} exhibit attractive robustness properties. The asymptotic breakdown point of the halfspace and the integrated rank-weighted medians are higher than $1/(d+1)$. In contrast, the projection median is known to have a breakdown point equal to $1/2$ \citep{DonohoGasko,IRW}.
\end{remark}
\begin{remark}When $d=1$, the $L_p$-Wasserstein distance enjoys an explicit expression involving quantile and distribution functions. Let $X^{1}\sim \mu_1,\;Y^{1} \sim \nu_1$ be two random variables where $\mu_1,\nu_1$ are univariate probability distributions. Denoting by $F^{-1}_{X^1}$ the quantile function of $X^1$, the $L_p$-Wasserstein distance can be written as
\begin{align}
\label{wass:dim1:quantile}
 W_p^p(\mu_1,\nu_1)= \int_{0}^{1} | F^{-1}_{X^1} (q) -F^{-1}_{Y^1}(q) |^p\; \mathrm{d}q.
\end{align}
Since data depth and its central regions are extensions of cdf and quantiles to dimension $d>1$, $DR_{p,\varepsilon}$ is then a possible (center-outward) generalization of (\ref{wass:dim1:quantile})  to higher dimensions. When $DR_{p,\varepsilon}$ is associated with the halfspace depth, a simple calculus (see Lemma~\ref{dimensionone} in the Appendix for mathematical details) leads to
\vspace{-0.7cm}

\begin{align*}
    DR_{p,\varepsilon}^p(\mu_1,\nu_1)= 2\int_{\varepsilon/2}^{1/2} &\max \Big\{ | F^{-1}_{X^1} (q) -F^{-1}_{Y^1}(q) |^p, \; \\& \hspace{-0.2cm}| F^{-1}_{X^1} (1-q) -F^{-1}_{Y^1}(1-q) |^p \Big \}~\mathrm{d}q.
\end{align*}

Thus, $W_p^p (\mu_1,\nu_1) \leq \underset{\varepsilon \rightarrow 0}{\lim} \;  DR_{p,\varepsilon}^p (\mu_1,\nu_1)$ in general where the equality holds for symmetric distributions.


\end{remark}

\subsection{Metric properties}  
  
We now investigate to which extent the proposed discrepancy measure satisfies the metric axioms. As a first go, we show that $DR_{p,\varepsilon}$ fulfills most conditions. However, it does not define distance in general. 

\begin{proposition}[\textsc{Metric properties}]\label{distance} For any convex data depth,   $\R$ is positive, symmetric and satisfies triangular inequality but the entailment $ DR_{p,\varepsilon}(\mu,\nu)=0 \Longrightarrow \mu=\nu $ does not hold in general. 
\end{proposition}

Thus, $DR_{p,\varepsilon}$  defines a pseudo-metric rather than a distance.
Based on distance, the proposed discrepancy measure preserves isometry invariance, as stated in the following proposition.

\begin{proposition}[\textsc{Isometry invariance}]\label{affine}
Let $A \in \mathbb{R}^{d\times d}$ be a non-singular matrix and $b\in \mathbb{R}^d$. Define the isometry mapping $g: x\in \mathbb{R}^d\mapsto Ax+b$ with $AA^\top=I_d$, then it holds:
\begin{equation*}
DR_{p,\varepsilon}(g_{\sharp} \mu, g_{\sharp} \nu) = DR_{p,\varepsilon}(\mu, \nu),
\end{equation*}

where $g_{\sharp}\mu$ is the push-forward of $\mu$ by $g$. In particular, it ensures invariance of $DR_{p,\varepsilon}$ under translations and rotations.
\end{proposition}
 
Although formulas \eqref{DRW} and \eqref{wass:dim1:quantile} are based on the same spirit, there are no apparent reasons why the proposed pseudo-metric should have the same behavior as the Wasserstein distance. It is the purpose of Proposition~\ref{general} to investigate 
the ability to factor out translations,  for $DR_{2,\varepsilon}$ associated with the halfspace depth, giving a positive answer for the case of two Gaussian distributions with equal covariance matrices.


\begin{proposition}[\textsc{Translation characterization}]\label{general} Consider $X,Y$ two random variables following $\mu\in \mathcal{M}_1(\mathcal{X})$ and $\nu\in \mathcal{M}_1(\mathcal{Y})$ with expectations $\mathbf{m}_1,\mathbf{m}_2$ and variance-covariance matrices $\mathbf{\Sigma}_1,\mathbf{\Sigma}_2$ respectively.  Denoting by $\mu^*,\nu^*$ the centered versions of $\mu,\nu$,  it holds:

\begin{align*}
 \Big \vert DR^2_{2,\varepsilon}(\mu,\nu) - DR^2_{2,\varepsilon}&(\mu^*,\nu^*)  - ||\mathbf{m}_1-\mathbf{m}_2||^2 \Big \vert \\& \leq  2~DR_{1,\varepsilon}(\mu^*,\nu^*) ||\mathbf{m}_1-\mathbf{m}_2|| .\nonumber
\end{align*}%
Now, let $\mu \sim \mathcal{N}(\mathbf{m}_1, \mathbf{\Sigma}_1)$ and $\nu \sim \mathcal{N}(\mathbf{m}_2, \mathbf{\Sigma}_2)$. Then it holds:

\vspace{-0.6cm}
{\small
\begin{align*}
 \Big| DR_{1,\varepsilon}(\mu,\nu)\hspace{-0.05cm}  -\hspace{-0.05cm} || \mathbf{m}_1\hspace{-0.05cm}  -\hspace{-0.05cm}  \mathbf{m}_2 ||  \Big|\hspace{-0.05cm}  \leq C_{\varepsilon}\hspace{-0.1cm} \supS \hspace{-0.1cm}\big| \sqrt{u^\top\mathbf{\Sigma}_1 u} \hspace{-0.05cm}-\hspace{-0.05cm} \sqrt{u^\top\mathbf{\Sigma}_2 u} \Big|,
\end{align*}}%
%
where $C_{\varepsilon}=\int_{0}^{1-\varepsilon} \big|\Phi^{-1}(1-\alpha(\beta))\big| ~\mathrm{d}\beta$ with $\Phi$  the cdf of the univariate standard Gaussian distribution.

\end{proposition}
Following Proposition~\ref{general}: when $\mathbf{\Sigma}_1=\mathbf{\Sigma}_2$, one has $DR_{2,\varepsilon}(\mu,\nu)=DR_{1,\varepsilon}(\mu,\nu)=||\mathbf{m}_1-\mathbf{m}_2||$ for any $\mu \sim \mathcal{N}(\mathbf{m}_1, \mathbf{\Sigma}_1)$ and $\nu \sim \mathcal{N}(\mathbf{m}_2, \mathbf{\Sigma}_2)$ providing a closed-form expression in the Gaussian case.


\subsection{Robustness}\label{sec:robust}
In this part, we explore the robustness of the proposed distance, associated with the halfspace depth, given the finite sample breakdown point (BP; \citealp{donohophd,donohohubert}). This notion investigates the smallest contamination fraction under which the estimation breaks down in the worst case. Considering a sample $\mathcal{S}_n=\{X_1,\ldots, X_n \}$ composed of i.i.d. observations drawn from a distribution $\mu$ with empirical measure  $\hat{\mu}_n=(1/n)\sum_{i=1}^{n} \delta_{X_i}$, the finite sample breakdown point of $DR_{p,\varepsilon}$ w.r.t. $\mathcal{S}_n$, denoted by $BP(DR_{p,\varepsilon}, \mathcal{S}_n)$  is defined as 

\begin{equation*}
 \min \left\{ \dfrac{o}{n+ o}: \; \underset{Z_1,\ldots, Z_o }{\sup } \;  DR_{p,\varepsilon} (\hat{\mu}_{n+o}, \hat{\mu}_n) = + \infty  \right\}\hspace*{-0.1cm},
\end{equation*}

where $\hat{\mu}_{n+o}=\frac{1}{n+o} \left( \sum_{i=1}^{n}\delta_{X_i} + \sum_{j=1}^{o}\delta_{Z_j} \right)$ is the ‘‘concatenate'' empirical measure between $X_1,\ldots, X_n$ and the contamination sample $Z_1,\ldots, Z_o$ with $o \in \mathbb{N}^*$. It is well known that the extremal regions of the halfspace depth are not robust while its central regions are rather stable under contamination \citep{DonohoGasko}. Fortunately, by construction, the parameter $\varepsilon$ allows us to ignore these extremal depth regions and thus ensure the robustness of the depth-trimmed regions distance. Based on the results of \citet{DonohoGasko} and \citet{illumination}, the following proposition provides a lower bound on the finite sample breakdown point of $DR_{p,\varepsilon}$, which highlights the robustness of the proposed distance (as well as its dependence on $\varepsilon$).

\begin{proposition} [\textsc{Breakdown Point}]\label{Breakdown} For the halfspace depth function, for any $\beta \in [0,1-\varepsilon]$ such that $\alpha(\beta,\hat{\mu}_n) < \alpha_{\max}(\hat{\mu}_n)$, it holds:
{\small
\begin{align*}
& BP(DR_{p,\varepsilon}, \mathcal{S}_n) \geq \\& \\&  \left\{
  \begin{array}{@{}ll@{}}
    \frac{\lceil n\alpha(1-\varepsilon,\hat{\mu}_n)/ (1-\alpha(1-\varepsilon,\hat{\mu}_n))\rceil}{n +\lceil n\alpha(1-\varepsilon,\hat{\mu}_n) / (1-\alpha(1-\varepsilon,\hat{\mu}_n))\rceil } & \! \! \! \!\text{if} \;  \alpha(1-\varepsilon,\hat{\mu}_n) \leq  \frac{\alpha_{\max}(\hat{\mu}_n) }{1+\alpha_{\max}(\hat{\mu}_n) },\\[0.4cm]
    \frac{\alpha_{\max}(\hat{\mu}_n) }{1+\alpha_{\max}(\hat{\mu}_n) } & \! \! \! \! \text{otherwise},
  \end{array}\right. 
\end{align*}}

where $\alpha_{\mathrm{max}}( \hat{\mu}_n)=\underset{x\in \mathbb{R}^d}{\max} \; \;  HD_{\hat{\mu}_n}(x)$.
 
\end{proposition}
Thus, at least a proportion $\alpha(1-\varepsilon,\hat{\mu}_n)/ (1-\alpha(1-\varepsilon,\hat{\mu}_n))$ of outliers must be added to break down $DR_{p,\varepsilon}$ when considering larger regions, while central regions are robust independently of $\varepsilon$. For two datasets,
$DR_{p,\varepsilon}$ breaks down if depth regions for at least one of the datasets do. The breakdown point is then the minimum between the breakdown points of each dataset. However, the breakdown point considers the worst case, i.e. the supremum over all possible contaminations, and is often pessimistic. Indeed the proposed pseudo-metric can handle more outliers in certain cases, as experimentally illustrated in Section~\ref{NUM}.

\section{EFFICIENT APPROXIMATE COMPUTATION} \label{APPROX}
Exact computation of $DR_{p,\varepsilon}$ can appear time-consuming due to the high time complexity of the algorithms that calculate depth-trimmed regions (c.f. \citealp{LiuZ14} and \citealp{pavlo2018} for projection and halfspace depths, respectively) rapidly growing with dimension. However, we design a universal approximate algorithm that achieves (log-) linear time complexity in $n$. Since properties (\textbf{D2--D4}) ensure that depth regions are convex bodies in $\mathbb{R}^d$, they can be characterized by their support functions defined by $h_{\mathcal{K}}(u)=\sup \{ \langle x,u \rangle, \; x\in \mathcal{K} \}$ for any $u\in \mathbb{S}^{d-1}$ where $\mathcal{K}$ is a convex compact of $\mathbb{R}^d$. Following \citet{schneider1993}, for two (convex) regions $D^{\alpha(\beta)}_{\mu}$ and $D^{\alpha(\beta)}_{\nu}$, the Hausdorff distance between them can be calculated as: 

\begin{equation*}
	d_{\mathcal{H}}(D^{\alpha(\beta)}_{\mu},D^{\alpha(\beta)}_{\nu})=\supS \big | h_{D^{\alpha(\beta)}_{\mu}}(u)- h_{D^{\alpha(\beta)}_{\nu}}(u)\big |.
\end{equation*}

As we shall see in Section~\ref{NUM}, mutual approximation of $h_{D^{\alpha(\beta)}_{\cdot}}(u)$ by points from the sample and of $\sup$ by taking maximum over a finite set of directions allows for stable estimation quality. Recently, motivated by their numerous applications, many algorithms have been developed for the (exact and approximate) computation of data depths; see, e.g., Section~5 of \citet{MoslerM21} for a recent overview. Depths satisfying the projection property (which also include halfspace and projection depth, see \citet{rainer2004}) can be approximated by taking minimum over univariate depths; see e.g. \citet{RousseeuwS98,Chen13,LiuZ14}, \citet{NagyDM20} for theoretical guarantees, and \citet{DyckerhoffMN20} for an experimental validation. The case of AI-IRW is easier since the expectation in \Cref{eq:AIIRW} can be approximated through Monte-Carlo approximation.

Let $\mathbf{X},\mathbf{Y}$ be  two samples  $\mathbf{X}=\{X_1,\ldots,X_n \}$ and $\mathbf{Y}=\{Y_1,\ldots,Y_m \}$ from $\mu,\nu$. When calculating approximated depth of sample points $D^\mbX\triangleq\{D(X_i,\hat{\mu}_n)\}_{i=1}^n$ (respectively $D^\mbY$), a matrix $\mbM^\mbX \in \mathbb{R}^{n\times K}$ (respectively $\mbM^\mbY \in \mathbb{R}^{m\times K}$) of projections of sample points on (a common) set of $K\in \mathbb{N}^*$ directions (with its element $\mbM^\mbX_{i,k}=\langle  u_k, X_i \rangle$ for some $u_k\sim\mathcal{U}(\mathbb{S}^{d-1})$, where $\mathcal{U}(\cdot)$ is the uniform probability distribution) can be obtained as a side product. More precisely, $D^\mbX,D^\mbY,\mbM^\mbX,\mbM^\mbY$ are used in Algorithm~\ref{alg:DRWappr}, which implements the MC-approximation of the integral in~\eqref{DRW}. Particular cases of approximation algorithms for the halfspace depth, the projection depth and the AI-IRW depth are recalled in Section~\ref{approxalgo} in the Appendix. Time complexity of Algorithm~\ref{alg:DRWappr} is $O\bigl(K(\Omega_{\cdot}(n \lor m, d) \lor n_\alpha (n \lor m))\bigr)$, where $\Omega_{\cdot}(\cdot,\cdot)$ stands for the complete complexity of computing univariate depths---in projections on $u$---for all points of the sample. As a byproduct, projections on $u$ can be saved to be reused after for the approximation of $h_{D^{\alpha(\beta)}_{\cdot}}(u)$. E.g., for the halfspace depth $\Omega_{hsp}(n, d)=O\bigl(n(d \lor \log\,n)\bigr)$ composed of projection of the data onto $u$, ordering them, and passing to record the depths (see e.g. \citealp{MozharovskyiML15}). For the projection depth,  $\Omega_{prj}(n, d)=O(nd)$, where after projecting the data onto $u$, univariate median and MAD can be computed with complexity $O(n)$ (see e.g. \citealp{LiuZ14}). For the AI-IRW depth, $\Omega_{aiirw}(n, d)=O(d^3 \;  \lor \; n(d \lor \log(n))$ since it involves the computation of the square root of the precision matrix. However, $O(d^3)$ may be improved, which depends on the algorithm employed for computing the inverse of the covariance matrix \citep{AIIRW}. In comparison with popular distances, fixing $n=m$,  the Wasserstein distance is of order $O(n^2(d \lor n))$ with approximations in $O(n^2d)$ for Sinkhorn \citep{cuturi13} and in $O(Kn(d \lor \log(n)))$ for the Sliced-Wasserstein distance \citep{sliced}; the MMD \citep{MMD} is of order $O(n^2d)$. For example, the computational complexity of $DR_{p,\varepsilon}$ with the projection depth is only of $O(Kn(d \lor n_{\alpha}))$ and thus competes with the fastest (max) sliced-Wasserstein distance. 

\begin{algorithm}[htb]
\caption{Approximation of $DR_{p,\varepsilon}$}
\textit{Initialization:} $\mathbf{X}$,$\mathbf{Y}$,$n_{\alpha}$,$K$
      \begin{algorithmic}[1]
      \STATE $H=0$; compute $D^\mbX,D^\mbY,\mbM^\mbX,\mbM^\mbY$
      \FOR{$\ell=1,\ldots,n_{\alpha}$}
      \STATE Draw $\beta_\ell\sim\mathcal{U}([0,1-\varepsilon])$
      \STATE Compute $\hat{\alpha}_{\ell}(\cdot):=\hat{\alpha}(\beta_{\ell}, \cdot)$ 
      \STATE Determine points inside $\alpha_{\ell}(\cdot)$-regions:\\  $\mathcal{I}_{\ell}^{\mathbf{X}}=\{i: D_i^{\mathbf{X}} > \hat{\alpha}_{\ell}(\mathbf{X})\}$; $\; \mathcal{I}_{\ell}^{\mathbf{Y}}=\{j: D_j^{\mathbf{Y}} > \hat{\alpha}_{\ell}(\mathbf{Y})\}$
      \FOR{$k=1,\ldots,K$}
      \STATE  Compute approximation of support functions:  $h_k^{\mathbf{X}}\hspace*{-0.1cm}=\hspace*{-0.1cm} \max \; \mathbf{M}^{\mathbf{X}}_{\mathcal{I}_{\ell}^{\mathbf{X}},k}$; $h_k^{\mathbf{Y}}= \max \; \mathbf{M}^{\mathbf{Y}}_{\mathcal{I}_{\ell}^{\mathbf{Y}},k}$
      \ENDFOR
      \STATE Increase cumulative Hausdorff distance: \\$H\mathrel{{+}{=}}\underset{k \le K}{\max}\,|h_k^{\mathbf{X}}-h_k^{\mathbf{Y}}|^p$
      \ENDFOR \\
 \textbf{Output}: $\widehat{DR}_{p,\varepsilon}=(H/n_\alpha)^{1/p}$
      \end{algorithmic}
      \label{alg:DRWappr}
\end{algorithm}

\vspace*{-0.2cm}

\section{NUMERICAL EXPERIMENTS}\label{NUM}
In this section, we first measure the quality of the approximation introduced in Section~\ref{APPROX} and explore its dependency on the number of projections. Further, we present two studies on robustness of the proposed pseudo-metric $DR_{p,\varepsilon}$ to outliers. On synthetic datasets, we investigate how $DR_{p,\varepsilon}$ behaves under the presence of outliers using two different settings. On a real image dataset extracted from Fashion-MNIST where images are seen as bags of pixels, we evaluate the robustness of spectral clustering based on $DR_{p,\varepsilon}$. Finally, we analyze the relevance of using $DR_{p,\varepsilon}$ as an evaluation metric in natural language generation to compare the empirical distributions of words of a pair of texts. 
Where applicable, we include state-of-the-art methods for comparison. Due to space limitations, experiments on the influence of the parameters $n_{\alpha}$ and $\varepsilon$, as well as on the statistical rates, are deferred to the Appendix section.
\begin{figure}[!h]
\vspace{-0.2cm}
\begin{center}
\begin{tabular}{cc}
\includegraphics[trim=2cm 0cm 0cm 0cm, scale=0.14]{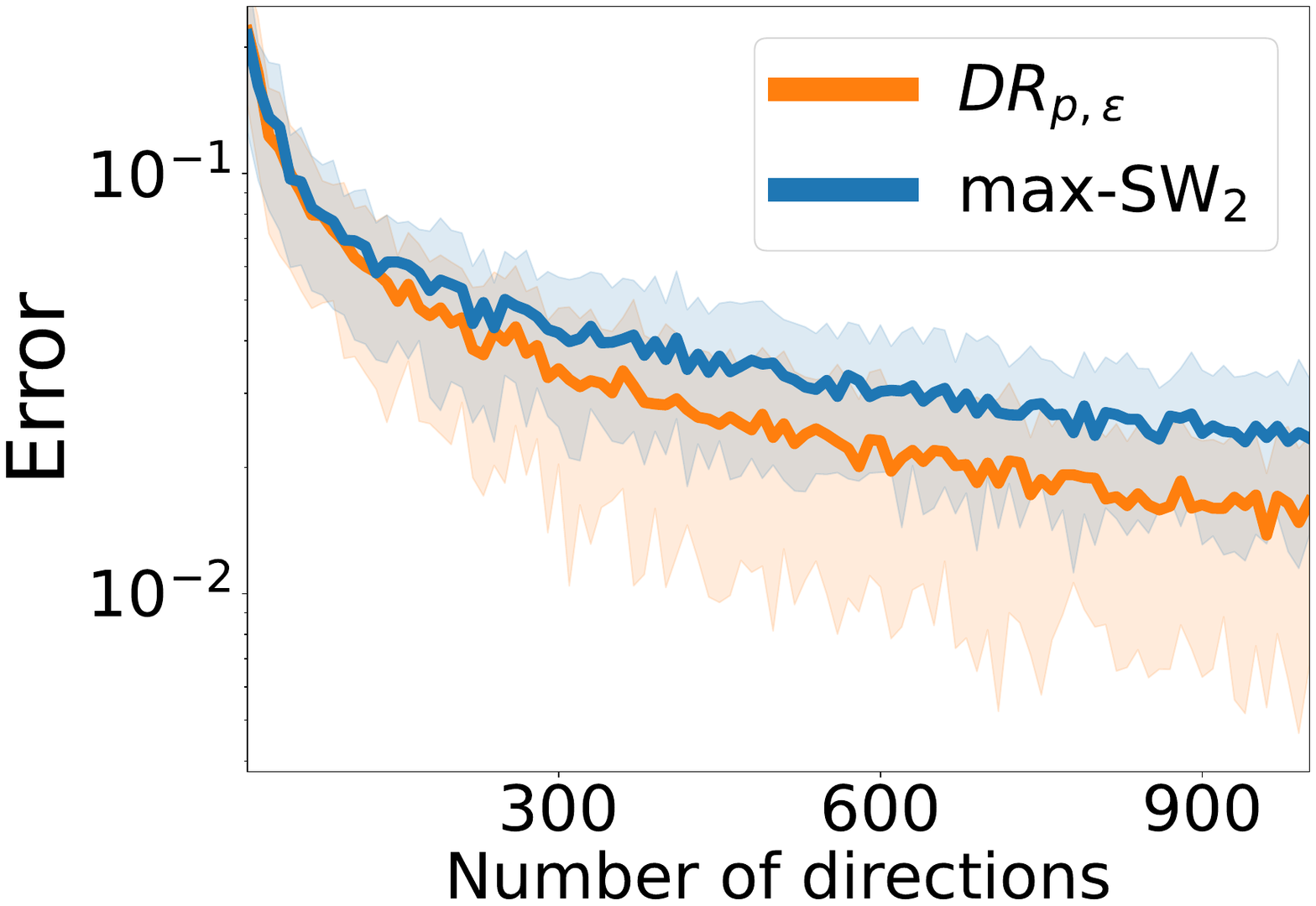}&  \includegraphics[trim=2.5cm 0cm 0cm 0cm, scale=0.14]{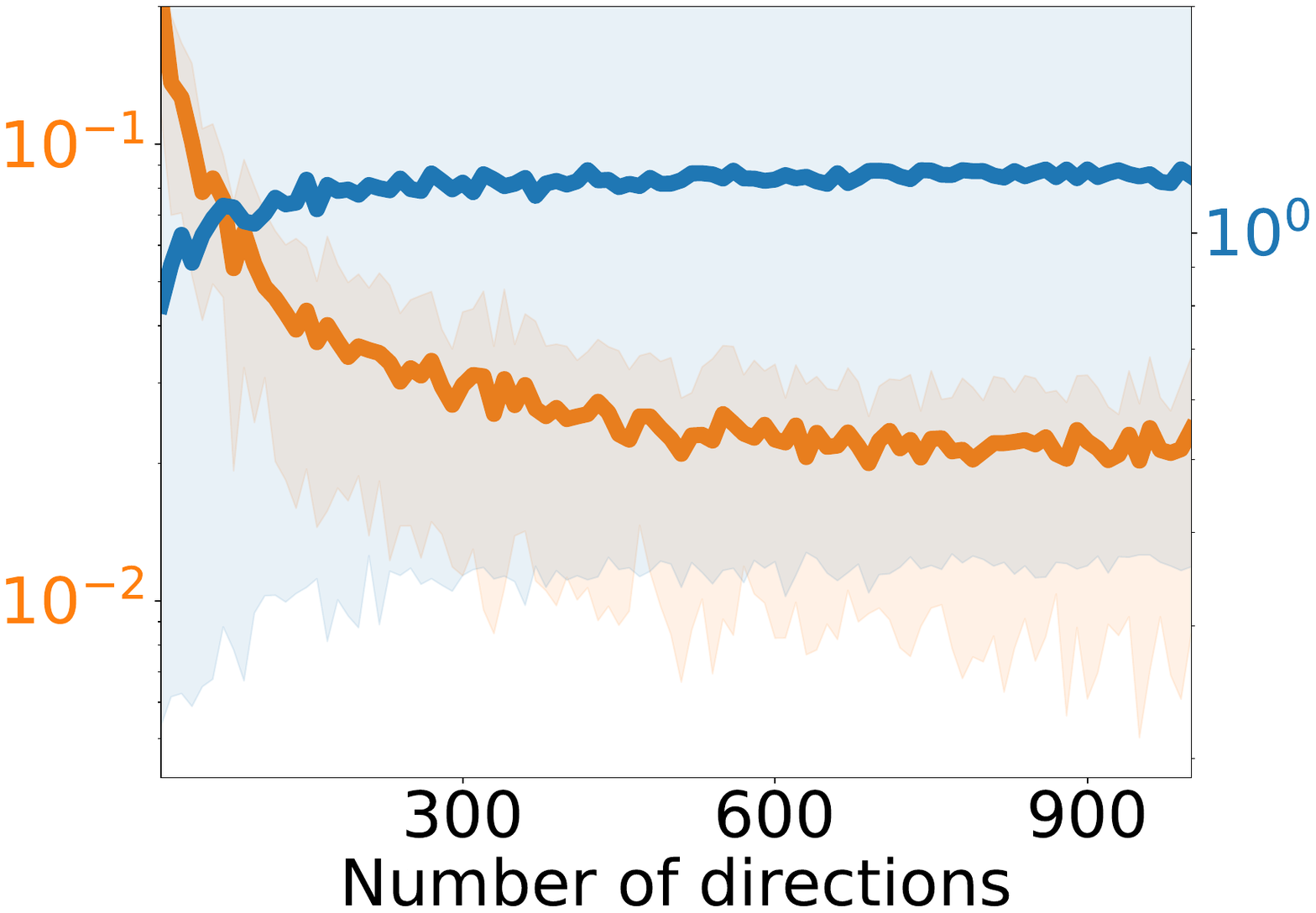}
\end{tabular}
\end{center}
\caption{Relative approximation error (averaged over $100$ runs) of $DR_{p,\varepsilon}$ and the max Sliced-Wassserstein for \emph{Gaussian} (left) and \emph{Cauchy} (right) sample with dimension $d=5$ for differing numbers of approximating directions.}
\label{fig:parameter1}
\end{figure}

\begin{figure}[!h]
\vspace{-0.2cm}
\begin{center}
\begin{tabular}{cc}
\includegraphics[trim=2.3cm 0cm 0cm 0cm, scale=0.14]{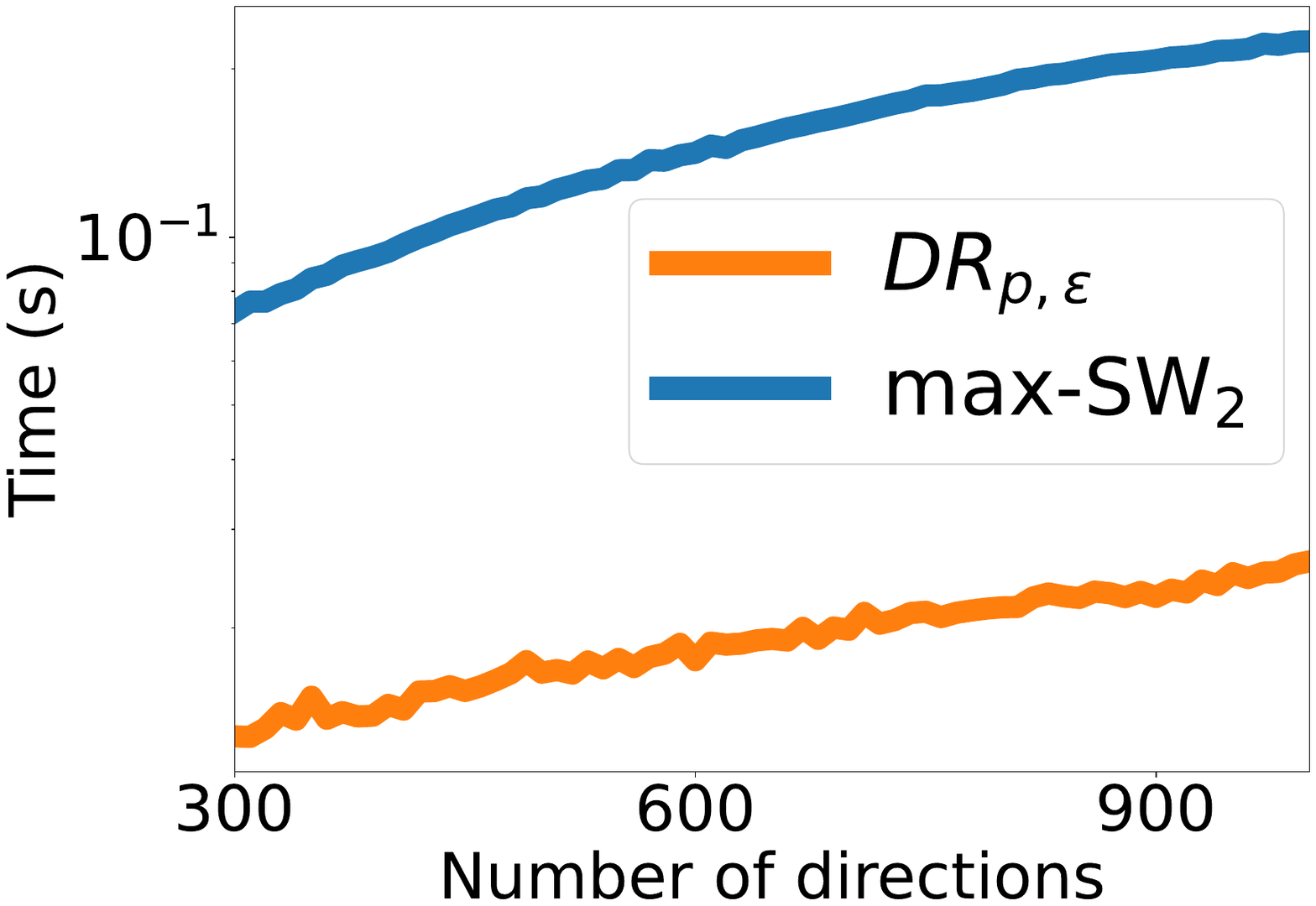}& \includegraphics[trim=2.5cm 0cm 0cm 0cm, scale=0.14]{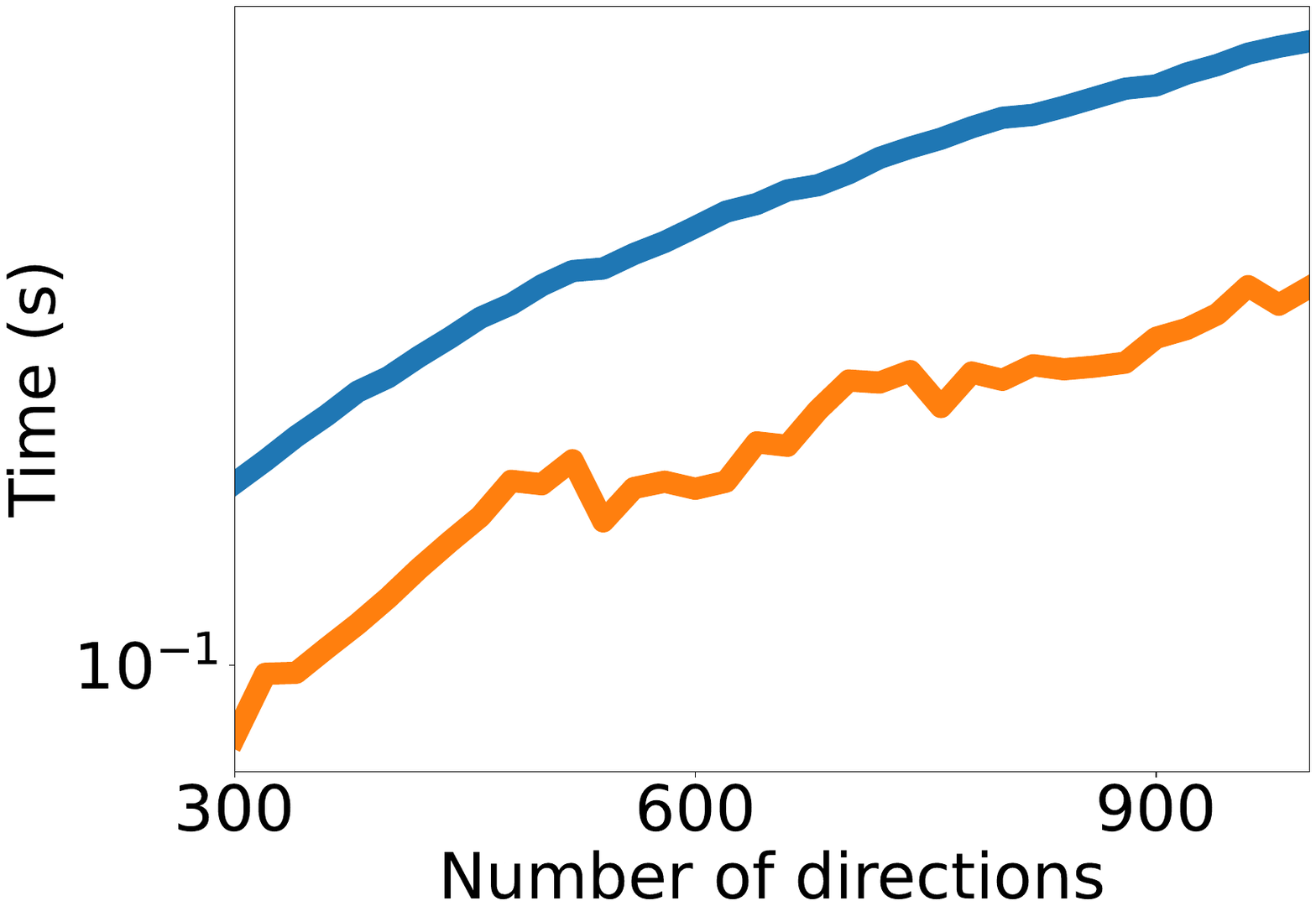}
\end{tabular}
\end{center}
\caption{Computation time (averaged over $100$ runs) of $DR_{p,\varepsilon}$ and the max Sliced-Wassserstein for \emph{Gaussian} with $n=100, \, d=5$ (left) and $n=1000, \, d=50$ (right) for differing numbers of approximating directions.}
\label{fig:computation}
\end{figure}

\noindent\textbf{Approximation error in terms of the number of projections.} 
Proposition~\ref{general} allows to derive a closed form expression for $DR_{2,\varepsilon}(\mu,\nu)$ when $\mu,\nu$ are Gaussian distributions with the same variance-covariance matrix. 
In order to investigate the quality of the approximation on light-tailed and heavy-tailed distributions, we focus on computing $DR_{p,\varepsilon}$ (with $p=2$, $\varepsilon=0.3$, $n_{\alpha}=20$ and using the halfspace depth) for varying number of random projections $K$ between a sample of 1000 points stemming from $\mu \sim \mathcal{N}(\mathbf{0}_d,I_d)$ for $d=5$ and two different samples. These two samples are constructed from 1000 observations stemming from \emph{Gaussian} and symmetrical \emph{Cauchy} distributions, both with a center equal to $\mathbf{7}_d$. Comparison with the approximation of max Sliced-Wasserstein (max-SW; see e.g. \citealp{kimia}), which shares the same closed-form as $DR_{2,\varepsilon}$, is also provided. Denoting by \text{max}-$\widehat{\text{SW}}$ the Monte-Carlo approximation of  the max-SW, the relative approximation errors, i.e., $(\widehat{DR}_{p,\varepsilon} -||\mathbf{7}_d||_2)/ ||\mathbf{7}_d||_2$ and (\text{max}-$\widehat{\text{SW}} -||\mathbf{7}_d||_2)/ ||\mathbf{7}_d||_2$, are computed investigating both the quality of the approximation and the robustness of these discrepancy measures. 
 Results that report the averaged approximation error and the 25-75\% empirical quantile intervals are depicted in Figure \ref{fig:parameter1}. They show that $DR_{p,\varepsilon}$ possesses the same behavior as max-SW when considering \emph{Gaussians} while it behaves advantageously for \emph{Cauchy} distribution. Computation times are depicted in Figure~\ref{fig:computation}, highlighting a constant-multiple improvement compared to the max-SW, which is already computationally fast.

\noindent \textbf{Robustness to outliers.}  We analyze the robustness of $DR_{p,\varepsilon}$ by measuring its ability to overcome outliers (its robustness regarding the influence of the parameter $\varepsilon$ are given in the Section~\ref{sec:app:eps} in the Appendix). In this benchmark, we naturally include existing robust extensions of the Wasserstein distance: Subspace Robust Wasserstein (SRW; \citealp{paty19}) searching for a maximal distance on lower-dimensional subspaces, ROBOT  \citep{mukherjee2020outlierrobust} and RUOT \citep{balaji2020robust} being robust modifications of the unbalanced optimal transport \citep{chizatunbalanced}. Medians-of-Means Wasserstein (MoMW; \citealp{staerman2020ot}) that replaces the empirical means in the Kantorivich duality formulae by the robust mean estimator MoM (see e.g.  \citealp{Lecue2017robust,laforgue2020}), is not employed due to high computational burden. Further, for completeness, we add the standard Wasserstein distance (W) and its approximation, the Sliced-Wasserstein (Sliced-W; \citealp{sliced}) distance, with the same number of projections ($K=1000$) as $DR_{p,\varepsilon}$. Since the scales of the compared methods differ, \emph{relative error} is used as a performance metric, i.e., the ratio of the absolute difference of the computed distance with and without anomalies divided by the latter. Two settings for a pair of distributions are addressed: (a) \emph{Fragmented hypercube} precedently studied in \cite{paty19}, where the source distribution is uniform in the hypercube $[-1,1]^2$ and the target distribution is transformed from the source via the map $T:x\mapsto  x+2\text{sign}(x)$ where $sign(.)$ is taken element-wisely. Outliers are drawn uniformly from $[-4,4]^2$. (b) Two multivariate standard \emph{Gaussian} distributions, one shifted by $\mathbf{10}_2$, with outliers drawn uniformly from $[-10,20]^2$. Our analysis is conducted over 500 sampled points from the distributions described above. 

\begin{figure}[!h]
\begin{center}

\begin{tabular}{cc}
\multicolumn{2}{c}{\includegraphics[scale=0.2, trim=0.8cm 1cm 0 0 ]{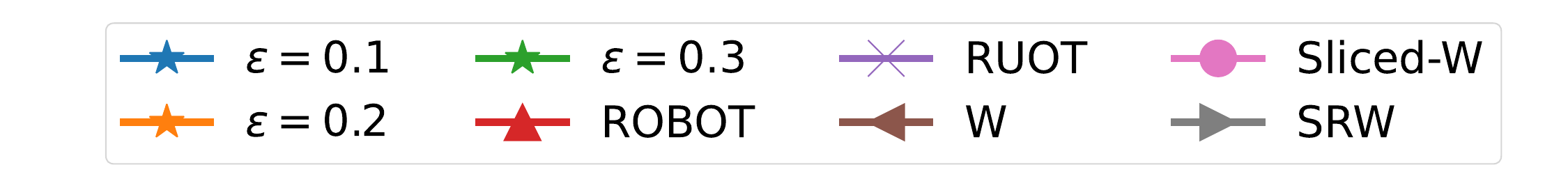}}\\
\includegraphics[trim=2.5cm 0cm 0cm 0cm, scale=0.142]{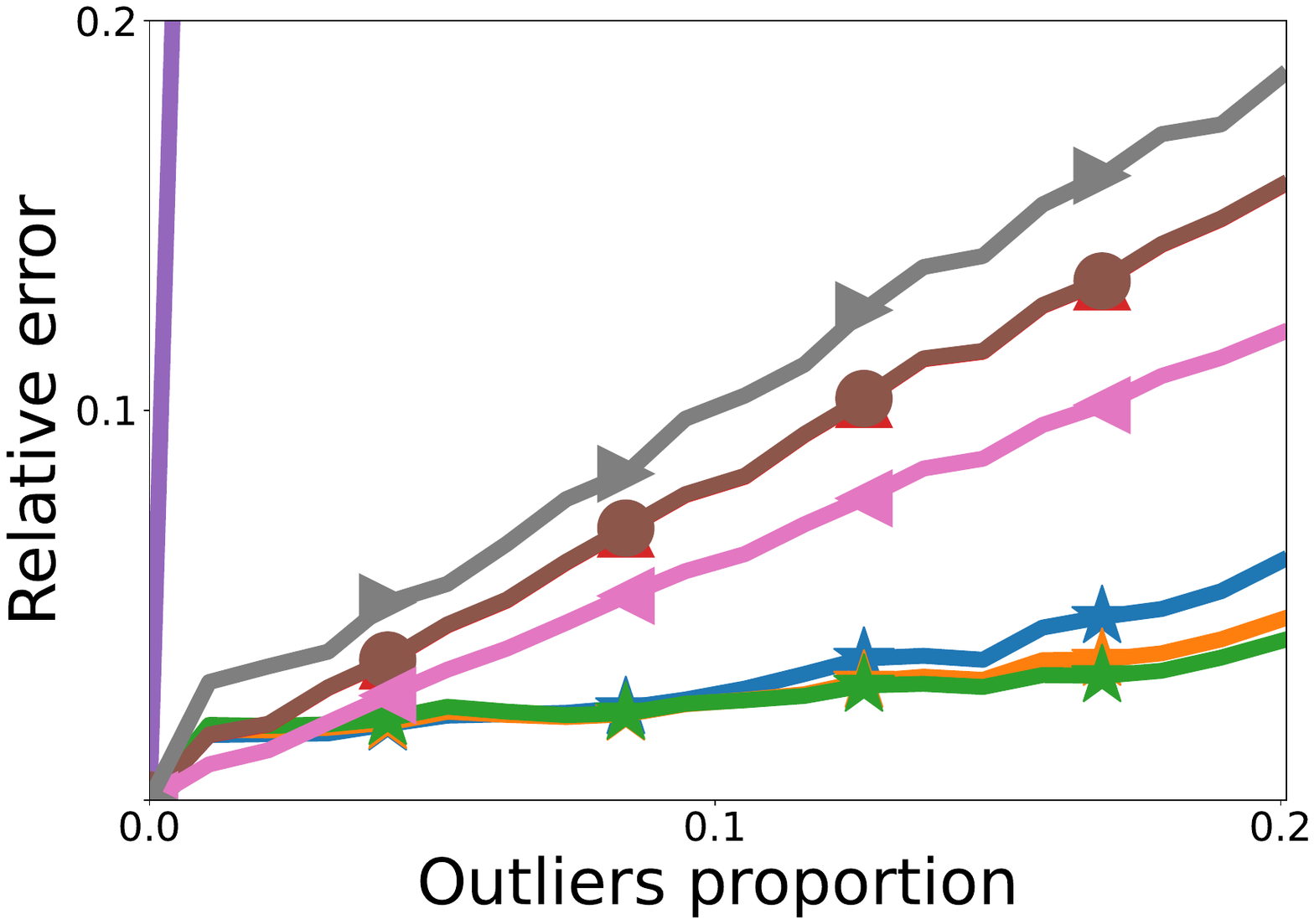} &
\includegraphics[trim=3cm 0cm 0cm 0cm, scale=0.142]{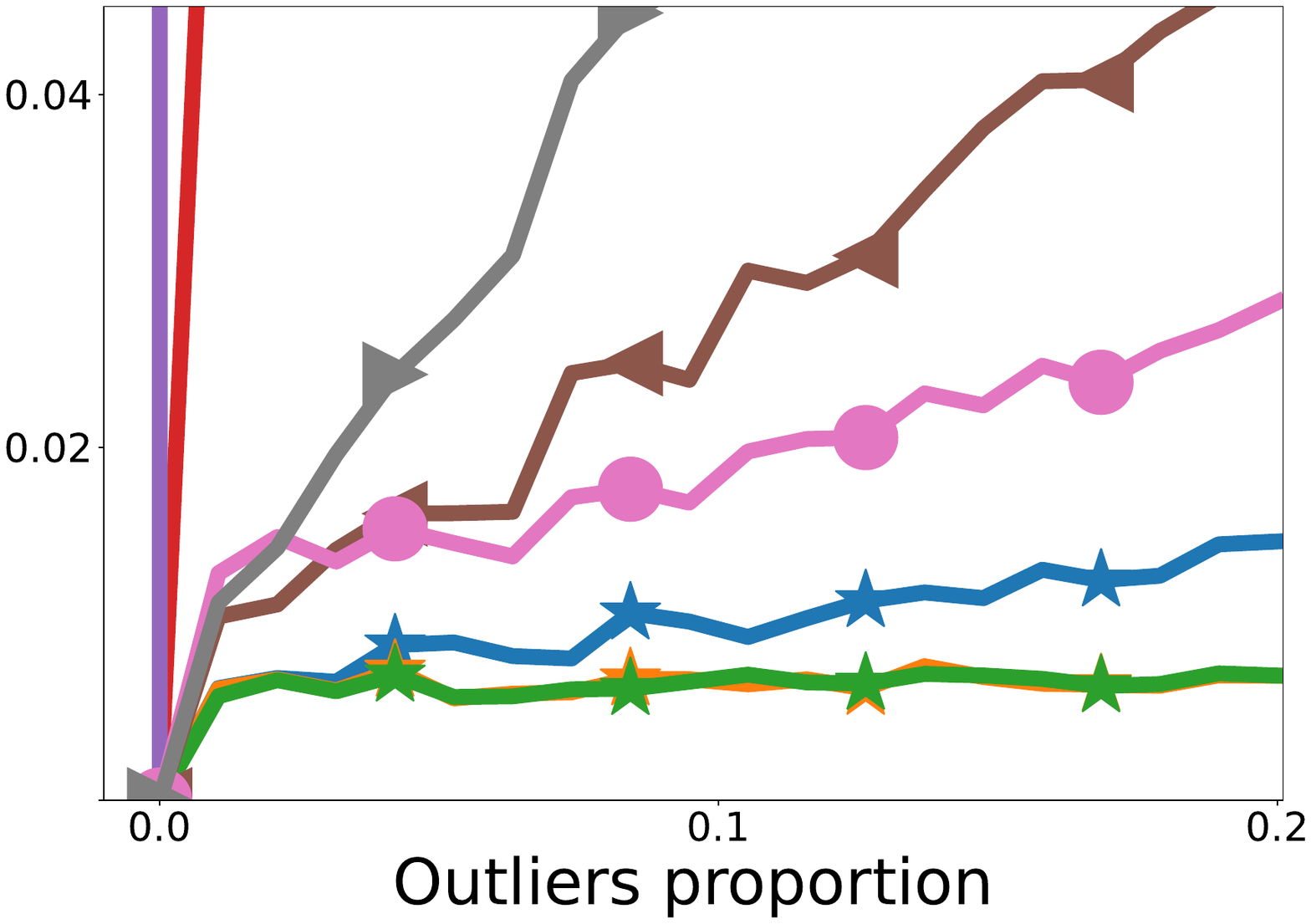} 
\end{tabular}
\end{center}
\caption{Relative error (averaged over $100$ runs) of different distances for increasing outliers proportion on \emph{fragmented hypercube} (left) and \emph{Gaussian} (right) data.}

\label{robustness}
\end{figure}

To investigate the robustness of $DR_{p,\varepsilon}$, we consider the following values of $\varepsilon$:  $0.1, 0.2, 0.3$ computed with the projection depth.
Thus, data depths are computed on source and target distributions such that $10\%$, $20\%$, $30\%$ of data with lower depth values w.r.t. each distribution are not used in computation of $DR_{p,0.1},DR_{p,0.2},DR_{p,0.3}$, respectively. Figure~\ref{robustness}, which plots the relative error depending on the portion of outliers varying up to $20\%$, illustrates advantageous behavior of $DR_{p,\varepsilon}$ (for $\varepsilon=0.1,0.2,0.3$) for reasonable (starting with $\approx 2.5\%$) contamination. It also confirms the pessimism of the breakdown point provided in Proposition~\ref{Breakdown} since $DR_{p,0.1}$ (represented by the blue curve) shows robustness to at least 20 \% of outliers.



\noindent \textbf{(Robust) Clustering on bags of pixels.} 
We demonstrate the relevance of the proposed pseudo-metric through an application to (robust) clustering. To that end, we perform spectral clustering \citep{spectralclustering} on two datasets derived from Fashion-MNIST (FM). Each grayscale image is seen as a bag of pixels \citep{bagofpixel}, i.e. as an empirical probability distribution over a 3-dimensional space (the two first dimensions indicate the pixel position and the third one, its intensity). The first dataset (FM) is constructed by taking the 100 first images in each class of the Fashion-MNIST dataset. The second dataset (Cont. FM), considered contaminated, is designed by introducing white patches on the left corner of 50 images drawn uniformly in the first dataset, which yields 5\% of contamination. We benchmark $DR_{p,\varepsilon}$ (using the projection depth) setting $p=2$ and $\varepsilon=0.1$ with the Wasserstein (W), the Sliced-Wasserstein (Sliced-W) and the Maximum Mean Discrepancy (MMD; \citealp{MMD}) distances. $DR_{p,\varepsilon}$ and the Sliced-Wasserstein are approximated by Monte-Carlo using 100 directions while the MMD distance is computed using a Gaussian kernel with a bandwidth equal to 1. As a baseline method, spectral clustering is also applied to images considered as vectors using Euclidean distance. Standard parameters of the \texttt{scikit-learn} spectral clustering implementation are employed with a number of clusters fixed to $10$. Performances of the benchmarked metrics are assessed by measuring the normalized mutual information (NMI; \citealp{shannon}) and the adjusted rank index (ARI; \citealp{ARI}), which are standard clustering evaluation measures when the ground truth class labels are available. Results presented in Table~\ref{table:clustering} show that for both cases, i.e. with or without contamination, spectral clustering based on $DR_{p,\varepsilon}$ outperforms spectral clustering based on the other metrics.

{\renewcommand{\arraystretch}{1.2} 
{\setlength{\tabcolsep}{0.2cm}
\begin{table}[h]
\begin{center}
\begin{tabular}{ccccc}\hline

 &\multicolumn{2}{c}{FM} &  \multicolumn{2}{c}{Cont. FM} \\
 \cmidrule(lr){2-3} \cmidrule(lr){4-5}
& NMI & ARI &NMI & ARI \\
$DR_{p,\varepsilon}$&  \textbf{0.58} & \textbf{0.43} & \textbf{0.55} & \textbf{0.42} \\
W & 0.50& 0.35 & 0.48 & 0.30 \\
Sliced-W & 0.55& 0.39& 0.47 &0.33  \\
MMD &0.54 &0.37& 0.50 & 0.36 \\
Euclidean & 0.50 &  0.32&  0.48 &  0.30 \\\hline
\end{tabular}
\end{center}
\caption{Spectral clustering performances.}
\label{table:clustering}
\end{table}
}}
\noindent \textbf{Automatic evaluation of natural language generation (NLG).} Collecting human annotations to evaluate NLG systems is both expensive and time-consuming. Thus, automatically assessing the similarity between two texts is highly interesting for the NLP community \citep{specia2010machine}. This task aims to build an evaluation metric that achieves a high correlation with the score given by a human annotator. String-based metrics (i.e. that compare the string representations of texts) such as BLEU \citep{bleu}, METEOR (MET.; \citealp{banerjee2005meteor}), ROUGE \citep{lin-2004-rouge}, TER \citep{snover2006study}, have been outperformed in many tasks by embedding-based metrics, i.e., that rely on continuous representations \citep{devlin2018bert}. Embedding-based metrics (e.g BertScore (BertS;  \citealp{zhang2019bertscore}) and MoverScore (MoverS; \citealp{zhao2019moverscore}) that are now the state-of-the-art domain, compare input and reference texts both represented as probability distributions and are both constructed similarly. The first step relies on a deep contextualized encoder (BERT in our case, see \citealp{devlin2018bert}) that maps texts into elements of a finite-dimensional space. Each text corresponds to a collection of words, where each word is represented by an element in $\mathbb{R}^d$, where $d$ is fixed by the encoder. The second step involves using a function that measures the similarity between the embedded texts. 

We follow previous BERT-based metrics and evaluate  performances of $DR_{p,\varepsilon}$ (with $p=2$, $\varepsilon=0.01$ and using the AI-IRW depth) on two different NLG tasks namely: data2text generation (using the WebNLG 2020 dataset \citealp{ferreira20202020}) and summarization. For the sake of place, summarization results and additional experimental details are reported in Section~\ref{sec:suplementary_nlp} in the Appendix. For WebNLG, we follow standard methods to assess the performance of NLG metrics (see e.g. \citealp{zhao2019moverscore}). We compute the correlation with the following annotation scores: \emph{correctness}, \emph{data coverage}, and \emph{relevance}. We report in Table~\ref{tab:web_nlg_sys} correlation results on the WebNLG task using Pearson ($r$), Spearman ($\rho$)  and Kendall ($\tau$) correlation coefficients. When performing a fair comparison between metrics, i.e. when $DR_{p,\varepsilon}$, W, Sliced-W, MMD are directly used on the output of BERT, we observe that $DR_{p,\varepsilon}$ achieves the best results on all configurations. It is worth noting that $DR_{p,\varepsilon}$ also compares favorably against existing state-of-the-art NLG methods in many different scenarios and shows promising results.






\begin{table}
\centering
\resizebox{0.48\textwidth}{!}{\begin{tabular}{lrrr|rrr|rrr}\hline  
  &  \multicolumn{3}{c}{Correctness} & \multicolumn{3}{c}{Data Coverage} & \multicolumn{3}{c}{Relevance} \\
 \cmidrule(lr){2-4} \cmidrule(lr){5-7} \cmidrule(lr){8-10}  &  $r$  &  $\tau$ &  $\rho$    &  $r$ &  $\tau$ &  $\rho$    &  $r$  & $\tau$ &  $\rho$    \\[0.5em]
  $DR_{p,\varepsilon}$& \textbf{89.4}  &  \textbf{80.0} & \textbf{92.6} & \textbf{84.2}  &   \underline{58.3} & \underline{72.3} & \textbf{86.2} & \underline{62.7} & \underline{72.9} \\[0.28em]
   W& 86.2 &  73.0 & 86.7   & 80.4 &  45.3 & 62.3  & 83.8 & 51.3 & 67.6 \\[0.35em]
  Sliced-W& 86.1  &  73.0 & 85.8 & 80.9  &  45.5 & 60.0& 82.0 &  51.3 & 68.2 \\[0.35em]
  MMD& 25.4  &  71.7 & 8.3 & 19.1  & 45.3 & 10.0& 26.1 &  51.3 & 15.0 \\[0.35em] \hline
BertS &      \underline{85.5} & \underline{73.3} &       83.4 &      
74.7 & \underline{53.3} &     \underline{68.2} &   
\underline{83.3}& \underline{65.0}  &     \textbf{79.4}    \\[0.35em]       
MoverS &      {84.1} &  \underline{73.3} &      \underline{84.1} &         
\underline{78.7} &   \underline{53.3}  &   66.2 &    
82.1 &   \underline{65.0} &   77.4  \\[0.35em] \hline
BLEU &      77.6 &   60.0 &     66.3 &       
55.7 &  36.6 &    50.2 &      
63.0 &  51.6 &    65.2    \\[0.35em]
ROUGE &      80.6 &  65.0&     65.0 &      
76.5 &  \textbf{60.3} &    \textbf{76.3} &     
64.3 &  56.7 &    {69.2}   \\[0.35em]
MET. &      \underline{86.5} &  \underline{70.0} &     {66.3} &     
\underline{77.3} &  46.6 &    50.2 &   
\underline{82.1} &  58.6 &    65.2   \\[0.35em]
TER &      79.6 &  58.0 &      \underline{78.3} &        69.7 & 38.0 &     58.2 &        75.0 &   \textbf{77.6} &   \underline{70.2}  \\[0.35em]
\bottomrule\end{tabular}}
\caption{Absolute correlation at the system level with three human judgment criteria. The best overall results are indicated in bold, best results in their group are underlined.}\label{tab:web_nlg_sys}
\end{table}


\section{DISCUSSION}

Leveraging the notion of statistical data depth function, a novel pseudo-metric between multivariate probability distributions---that meets the aforementioned requirements---was introduced.
The developed framework exhibits inherent versatility due to numerous data depth variants. The linear approximation algorithm and the robustness property make $DR_{p,\varepsilon}$ a promising tool for a large spectrum of applications beyond clustering and NLG, e.g. in generative adversarial networks (GANs) or information retrieval. Moreover, recent works extending the notion of data depth to further types of data such as functional and time-series data~\citep{NietoRB16,gibels2018general}, directional (or spherical) data~\citep{ley2014}, random matrices~\citep{paindaveine2018}, curves (or paths) data~\citep{lafaye2020}, and random sets~\citep{CascosLM21} shall allow for the use of the proposed pseudo-metric for a wide range of applications. 


%

\bibliographystyle{apalike}
\bibliography{DR_distance}

\begin{thebibliography}{}

\bibitem[Arjovsky et~al., 2017]{arjovsky2017}
Arjovsky, M., Chintala, S., and Bottou, L. (2017).
\newblock Wasserstein gan.
\newblock {\em arXiv preprint arXiv:1701.07875}.

\bibitem[Auer et~al., 2007]{auer2007dbpedia}
Auer, S., Bizer, C., Kobilarov, G., Lehmann, J., Cyganiak, R., and Ives, Z.
  (2007).
\newblock Dbpedia: A nucleus for a web of open data.
\newblock In {\em The semantic web}, pages 722--735. Springer.

\bibitem[Balaji et~al., 2020]{balaji2020robust}
Balaji, Y., Chellappa, R., and Feizi, S. (2020).
\newblock Robust optimal transport with applications in generative modeling and
  domain adaptation.
\newblock {\em arXiv preprint arXiv:2010.05862}.

\bibitem[Banerjee and Lavie, 2005]{banerjee2005meteor}
Banerjee, S. and Lavie, A. (2005).
\newblock Meteor: An automatic metric for mt evaluation with improved
  correlation with human judgments.
\newblock In {\em Proceedings of the acl workshop on intrinsic and extrinsic
  evaluation measures for machine translation and/or summarization}, pages
  65--72.

\bibitem[Barnett, 1976]{barnett}
Barnett, V. (1976).
\newblock The ordering of multivariate data.
\newblock {\em Journal of the royal society A}.

\bibitem[Bhandari et~al., 2020]{bhandari2020re}
Bhandari, M., Gour, P., Ashfaq, A., Liu, P., and Neubig, G. (2020).
\newblock Re-evaluating evaluation in text summarization.
\newblock {\em arXiv preprint arXiv:2010.07100}.

\bibitem[Billingsley, 1999]{billingsley1999}
Billingsley, P. (1999).
\newblock {\em Convergence of probability measures (2nd ed.)}.
\newblock John Wiley \& Sons.

\bibitem[Brunel, 2019]{bruneltukey}
Brunel, V.-E. (2019).
\newblock Concentration of the empirical level sets of tukey’s halfspace
  depth.
\newblock {\em Probability Theory and Related Fields}, 173(3):1165--1196.

\bibitem[Cascos et~al., 2021]{CascosLM21}
Cascos, I., Li, Q., and Molchanov, I. (2021).
\newblock Depth and outliers for samples of sets and random sets distributions.
\newblock {\em Australian \& New Zealand Journal of Statistics}, 63(1):55--82.

\bibitem[Cha and Srihari, 2002]{cha2002}
Cha, S. and Srihari, S.~N. (2002).
\newblock On measuring the distance between histograms.
\newblock {\em Pattern Recognit.}, 35(6):1355--1370.

\bibitem[Chapuis et~al., 2021]{chapuis2021code}
Chapuis, E., Colombo, P., Labeau, M., and Clave, C. (2021).
\newblock Code-switched inspired losses for generic spoken dialog
  representations.
\newblock {\em arXiv preprint arXiv:2108.12465}.

\bibitem[Chapuis et~al., 2020]{chapuis2020hierarchical}
Chapuis, E., Colombo, P., Manica, M., Labeau, M., and Clavel, C. (2020).
\newblock Hierarchical pre-training for sequence labelling in spoken dialog.
\newblock {\em arXiv preprint arXiv:2009.11152}.

\bibitem[Chatzikoumi, 2020]{chatzikoumi2020evaluate}
Chatzikoumi, E. (2020).
\newblock How to evaluate machine translation: A review of automated and human
  metrics.
\newblock {\em Natural Language Engineering}, 26(2):137--161.

\bibitem[Chen et~al., 2013]{Chen13}
Chen, D., Morin, P., and Wagner, U. (2013).
\newblock Absolute approximation of tukey depth: Theory and experiments.
\newblock {\em Computational Geometry}, 46(5):566 -- 573.

\bibitem[Chen and Bansal, 2018]{chen2018fast}
Chen, Y.-C. and Bansal, M. (2018).
\newblock Fast abstractive summarization with reinforce-selected sentence
  rewriting.
\newblock {\em arXiv preprint arXiv:1805.11080}.

\bibitem[Chizat et~al., 2018]{chizatunbalanced}
Chizat, L., Peyré, G., Schmitzer, B., and Vialard, F.-X. (2018).
\newblock Unbalanced optimal transport: dynamic and kantorovich formulations.
\newblock {\em Journal of Functional Analysis}, 274(11):3090 -- 3123.

\bibitem[Clark et~al., 2019]{clark2019sentence}
Clark, E., Celikyilmaz, A., and Smith, N.~A. (2019).
\newblock Sentence mover’s similarity: Automatic evaluation for
  multi-sentence texts.
\newblock In {\em Proceedings of the 57th Annual Meeting of the Association for
  Computational Linguistics}, pages 2748--2760.

\bibitem[Colombo et~al., 2020]{colombo2020guiding}
Colombo, P., Chapuis, E., Manica, M., Vignon, E., Varni, G., and Clavel, C.
  (2020).
\newblock Guiding attention in sequence-to-sequence models for dialogue act
  prediction.
\newblock In {\em Proceedings of the AAAI Conference on Artificial
  Intelligence}, volume~34, pages 7594--7601.

\bibitem[Colombo et~al., 2021a]{colombo2021automatic}
Colombo, P., Staerman, G., Clavel, C., and Piantanida, P. (2021a).
\newblock Automatic text evaluation through the lens of wasserstein
  barycenters.
\newblock {\em arXiv preprint arXiv:2108.12463}.

\bibitem[Colombo et~al., 2019]{colombo2019affect}
Colombo, P., Witon, W., Modi, A., Kennedy, J., and Kapadia, M. (2019).
\newblock Affect-driven dialog generation.
\newblock {\em arXiv preprint arXiv:1904.02793}.

\bibitem[Colombo et~al., 2021b]{colombo2021beam}
Colombo, P., Yang, C., Varni, G., and Clavel, C. (2021b).
\newblock Beam search with bidirectional strategies for neural response
  generation.
\newblock {\em arXiv preprint arXiv:2110.03389}.

\bibitem[Csiszàr, 1963]{csiszar}
Csiszàr, I. (1963).
\newblock Eine informationstheoretische ungleichung und ihre anwendung auf den
  bewis der ergodizität von markhoffschen kette.
\newblock {\em Magyer Tud. Akad. Mat. Kutato Int. Koezl}, 8:85--108.

\bibitem[Cuturi et~al., 2013]{cuturi13}
Cuturi, M., Teboul, O., and Vert, J.-P. (2013).
\newblock Sinkhorn distances: Lightspeed computation of optimal transportation.
\newblock In {\em Advances in Neural Information Processing Systems}.

\bibitem[Dang and Owczarzak, 2008]{dang2008overview}
Dang, H.~T. and Owczarzak, K. (2008).
\newblock Overview of the tac 2008 update summarization task.
\newblock In {\em Proceedings of the Text Analysis Conference (TAC)}.

\bibitem[Devlin et~al., 2018]{bert}
Devlin, J., Chang, M.-W., Lee, K., and Toutanova, K. (2018).
\newblock Bert: Pre-training of deep bidirectional transformers for language
  understanding.
\newblock {\em arXiv preprint arXiv:1810.04805}.

\bibitem[Devlin et~al., 2019]{devlin2018bert}
Devlin, J., Chang, M.-W., Lee, K., and Toutanova, K. (2019).
\newblock {BERT}: {P}re-{T}raining of {D}eep {B}idirectional {T}ransformers for
  {L}anguage {U}nderstanding.
\newblock In {\em Proceedings of the 2019 Conference of the North {A}merican
  Chapter of the Association for Computational Linguistics: Human Language
  Technologies, Volume 1 (Long and Short Papers)}, pages 4171--4186.

\bibitem[Dong et~al., 2019]{dong2019unified}
Dong, L., Yang, N., Wang, W., Wei, F., Liu, X., Wang, Y., Gao, J., Zhou, M.,
  and Hon, H.-W. (2019).
\newblock Unified language model pre-training for natural language
  understanding and generation.
\newblock {\em arXiv preprint arXiv:1905.03197}.

\bibitem[Donoho, 1982]{donohophd}
Donoho, D.~L. (1982).
\newblock Breakdown properties of location estimators.
\newblock {\em P.h.D., qualifying paper, Dept. Statistics, Hardvard
  University}.

\bibitem[Donoho and Gasko, 1992]{DonohoGasko}
Donoho, D.~L. and Gasko, M. (1992).
\newblock Breakdown properties of location estimates based on half space depth
  and projected outlyingness.
\newblock {\em The Annals of Statistics}, 20:1803--1827.

\bibitem[Donoho and Hubert, 1983]{donohohubert}
Donoho, D.~L. and Hubert, P.~J. (1983).
\newblock The notion of breakdown point.
\newblock {\em A Festschrift for Erich Lehman}, pages 157--184.

\bibitem[Dyckerhoff, 2004]{rainer2004}
Dyckerhoff, R. (2004).
\newblock Data depth satisfying the projection property.
\newblock {\em Allgemeines Statistisches Archiv}, 88(2):163--190.

\bibitem[Dyckerhoff et~al., 2021]{DyckerhoffMN20}
Dyckerhoff, R., Mozharovskyi, P., and Nagy, S. (2021).
\newblock Approximate computation of projection depths.
\newblock {\em Computational Statistics and Data Analysis}, 157:107166.

\bibitem[Einhmahl and Mason, 1992]{mason}
Einhmahl, J.~H. and Mason, D.~M. (1992).
\newblock Generalized quantile process.
\newblock {\em The annals of statistics}, 20(2):1062--1078.

\bibitem[Ferreira et~al., 2020]{ferreira20202020}
Ferreira, T., Gardent, C., Ilinykh, N., van~der Lee, C., Mille, S., Moussallem,
  D., and Shimorina, A. (2020).
\newblock The 2020 bilingual, bi-directional webnlg+ shared task overview and
  evaluation results (webnlg+ 2020).
\newblock In {\em Proceedings of the 3rd International Workshop on Natural
  Language Generation from the Semantic Web (WebNLG+)}.

\bibitem[Ferreira et~al., 2018]{ferreira2018enriching}
Ferreira, T.~C., Moussallem, D., Krahmer, E., and Wubben, S. (2018).
\newblock Enriching the webnlg corpus.
\newblock In {\em Proceedings of the 11th International Conference on Natural
  Language Generation}, pages 171--176.

\bibitem[Garcia et~al., 2019]{garcia2019token}
Garcia, A., Colombo, P., Essid, S., d'Alch{\'e} Buc, F., and Clavel, C. (2019).
\newblock From the token to the review: A hierarchical multimodal approach to
  opinion mining.
\newblock {\em arXiv preprint arXiv:1908.11216}.

\bibitem[Gardent et~al., 2017]{gardent2017creating}
Gardent, C., Shimorina, A., Narayan, S., and Perez-Beltrachini, L. (2017).
\newblock Creating training corpora for nlg micro-planning.
\newblock In {\em 55th annual meeting of the Association for Computational
  Linguistics (ACL)}.

\bibitem[Gehrmann et~al., 2018]{gehrmann-etal-2018-bottom}
Gehrmann, S., Deng, Y., and Rush, A. (2018).
\newblock Bottom-up abstractive summarization.
\newblock In {\em Proceedings of the 2018 Conference on Empirical Methods in
  Natural Language Processing}, pages 4098--4109.

\bibitem[Gijbels and Nagy, 2017]{gibels2018general}
Gijbels, I. and Nagy, S. (2017).
\newblock On a general definition of depth for functional data.
\newblock {\em Statistical Science}, 32(4):630--639.

\bibitem[Gretton et~al., 2007]{MMD}
Gretton, A., Borgwardt, K., Rasch, M., Sch\"{o}lkopf, B., and Smola, A. (2007).
\newblock A kernel method for the two-sample-problem.
\newblock {\em Advances in Neural Information Processing Systems}.

\bibitem[Hallin et~al., 2010]{hallin2010}
Hallin, M., Paindaveine, D., and Šiman, M. (2010).
\newblock Multivariate quantiles and multiple-output regression quantiles: From
  l1 optimization to halfspace depth.
\newblock {\em Ann. Statist.}, 38(2):635--669.

\bibitem[Hubert and Arabie, 1985]{ARI}
Hubert, L. and Arabie, P. (1985).
\newblock Comparing partitions.
\newblock {\em Journal of Classification}, 2(1):193--218.

\bibitem[Jebara, 2003]{bagofpixel}
Jebara, T. (2003).
\newblock Images as bags of pixels.
\newblock In {\em Proceedings of the Ninth IEEE International Conference on
  Computer Vision}, pages 265--272.

\bibitem[Jörnsten, 2004]{jornsten}
Jörnsten, R. (2004).
\newblock Clustering and classification based on the l1 data depth.
\newblock {\em Journal of Multivariate Analysis}, 90(1):67 -- 89.

\bibitem[Kedzie et~al., 2018]{kedzie2018content}
Kedzie, C., McKeown, K., and Daume~III, H. (2018).
\newblock Content selection in deep learning models of summarization.
\newblock {\em arXiv preprint arXiv:1810.12343}.

\bibitem[Kendall, 1938]{kendall1938new}
Kendall, M.~G. (1938).
\newblock A new measure of rank correlation.
\newblock {\em Biometrika}, 30(1/2):81--93.

\bibitem[Koehn, 2009]{koehn2009statistical}
Koehn, P. (2009).
\newblock {\em Statistical machine translation}.
\newblock Cambridge University Press.

\bibitem[Kolouri et~al., 2019]{kimia}
Kolouri, S., Nadjahi, K., Umut, S., Badeau, R., and Rohde~K., G. (2019).
\newblock Generalized sliced wasserstein distance.
\newblock In {\em Advances Neural Information Processing Systems}.

\bibitem[Koltchinskii and Dudley, 1996]{kol}
Koltchinskii, V.~I. and Dudley, R.~M. (1996).
\newblock On spatial quantiles.
\newblock {\em Unpublished manuscript}.

\bibitem[Koshevoy and Mosler, 1997]{koshevoy1997}
Koshevoy, G. and Mosler, K. (1997).
\newblock Zonoid trimming for multivariate distributions.
\newblock {\em The Annals of Statistics}, 25(5):1998--2017.

\bibitem[Kullback, 1959]{kullback1959}
Kullback, S. (1959).
\newblock {\em Information Theory and Statistics}.
\newblock John Wiley.

\bibitem[Kusner et~al., 2015]{kusner2015word}
Kusner, M., Sun, Y., Kolkin, N., and Weinberger, K. (2015).
\newblock From word embeddings to document distances.
\newblock In {\em International conference on machine learning}, pages
  957--966. PMLR.

\bibitem[Lafaye et~al., 2020]{lafaye2020}
Lafaye, P., Mozharovskyi, P., and Vimond, M. (2020).
\newblock Depth for curve data and applications.
\newblock {\em Journal of the American Statistical Association}, pages 1--17.
\newblock in press.

\bibitem[Laforgue et~al., 2021]{laforgue2020}
Laforgue, P., Staerman, G., and Cl{\'e}men{\c{c}}on, S. (2021).
\newblock Generalization bounds in the presence of outliers: a median-of-means
  study.
\newblock In {\em Proceedings of the 38th International Conference on Machine
  Learning}, volume 139, pages 5937--5947.

\bibitem[Lange et~al., 2014]{LangeMM14}
Lange, T., Mosler, K., and Mozharovskyi, P. (2014).
\newblock Fast nonparametric classification based on data depth.
\newblock {\em Statistical Papers}, 55(1):49--69.

\bibitem[Lecué and Lerasle, 2020]{Lecue2017robust}
Lecué, G. and Lerasle, M. (2020).
\newblock Robust machine learning by median-of-means: Theory and practice.
\newblock {\em The Annals of Statistics}, 48(2):906--931.

\bibitem[Leusch et~al., 2006]{leusch-etal-2006-cder}
Leusch, G., Ueffing, N., and Ney, H. (2006).
\newblock {CDER}: Efficient {MT} evaluation using block movements.
\newblock In {\em 11th Conference of the EACL}.

\bibitem[Leusch et~al., 2003]{leusch2003novel}
Leusch, G., Ueffing, N., Ney, H., et~al. (2003).
\newblock A novel string-to-string distance measure with applications to
  machine translation evaluation.
\newblock In {\em Proceedings of Mt Summit IX}, pages 240--247.

\bibitem[Lewis et~al., 2019]{lewis2019bart}
Lewis, M., Liu, Y., Goyal, N., Ghazvininejad, M., Mohamed, A., Levy, O.,
  Stoyanov, V., and Zettlemoyer, L. (2019).
\newblock Bart: Denoising sequence-to-sequence pre-training for natural
  language generation, translation, and comprehension.
\newblock {\em arXiv preprint arXiv:1910.13461}.

\bibitem[Ley et~al., 2014]{ley2014}
Ley, C., Sabbah, C., and Verdebout, T. (2014).
\newblock A new concept of quantiles for directional data and the angular
  {M}ahalanobis depth.
\newblock {\em Electronic Journal of Statistics}, 8(1):795--816.

\bibitem[Li et~al., 2012]{LI}
Li, J., Cuesta-Albertos, J.~A., and Liu, R.~Y. (2012).
\newblock Dd-classifier: Nonparametric classification procedure based on
  dd-plot.
\newblock {\em JASA}, 107(498):737--753.

\bibitem[Lin, 2004]{lin-2004-rouge}
Lin, C.-Y. (2004).
\newblock {ROUGE}: A package for automatic evaluation of summaries.
\newblock In {\em Text Summarization Branches Out}, pages 74--81.

\bibitem[Liu, 1990]{liu1990}
Liu, R.~Y. (1990).
\newblock On a notion of data depth based on random simplices.
\newblock {\em The Annals of Statistics}, 18(1):405--414.

\bibitem[Liu, 1992]{Liu92}
Liu, R.~Y. (1992).
\newblock {\em Data Depth and Multivariate Rank Tests}, page 279–294.
\newblock North-Holland, Amsterdam.

\bibitem[Liu and Singh, 1993]{LiuSingh}
Liu, R.~Y. and Singh, K. (1993).
\newblock A quality index based on data depth and multivariate rank tests.
\newblock {\em Journal of the American Statistical Association},
  88(421):252--260.

\bibitem[Liu et~al., 2019a]{pavlo2018}
Liu, X., Mosler, K., and Mozharovskyi, P. (2019a).
\newblock Fast computation of tukey trimmed regions and median in dimension p
  {$>$} 2.
\newblock {\em Journal of Computational and Graphical Statistics},
  28(3):682--697.

\bibitem[Liu and Zuo, 2014]{LiuZ14}
Liu, X. and Zuo, Y. (2014).
\newblock Computing projection depth and its associated estimators.
\newblock {\em Statistics and Computing}, 24(1):51--63.

\bibitem[Liu and Lapata, 2019]{liu2019text}
Liu, Y. and Lapata, M. (2019).
\newblock Text summarization with pretrained encoders.
\newblock {\em arXiv preprint arXiv:1908.08345}.

\bibitem[Liu et~al., 2019b]{liu2019roberta}
Liu, Y., Ott, M., Goyal, N., Du, J., Joshi, M., Chen, D., Levy, O., Lewis, M.,
  Zettlemoyer, L., and Stoyanov, V. (2019b).
\newblock Roberta: A robustly optimized bert pretraining approach.
\newblock {\em arXiv preprint arXiv:1907.11692}.

\bibitem[MacKay, 2003]{mackay2003}
MacKay, D. J.~C. (2003).
\newblock {\em Information theory, inference and learning algorithms}.
\newblock Cambridge university press.

\bibitem[Mairesse et~al., 2010]{mairesse2010phrase}
Mairesse, F., Gasic, M., Jurcicek, F., Keizer, S., Thomson, B., Yu, K., and
  Young, S. (2010).
\newblock Phrase-based statistical language generation using graphical models
  and active learning.
\newblock In {\em Proceedings of the 48th Annual Meeting of the Association for
  Computational Linguistics}, pages 1552--1561.

\bibitem[McNamee and Dang, 2009]{mcnamee2009overview}
McNamee, P. and Dang, H.~T. (2009).
\newblock Overview of the tac 2009 knowledge base population track.
\newblock In {\em Proceedings of the Text Analysis Conference (TAC)},
  volume~17, pages 111--113.

\bibitem[Melamed et~al., 2003]{melamed2003precision}
Melamed, I.~D., Green, R., and Turian, J. (2003).
\newblock Precision and recall of machine translation.
\newblock In {\em Companion Volume of the Proceedings of HLT-NAACL 2003-Short
  Papers}, pages 61--63.

\bibitem[Mikolov et~al., 2013]{word2vec}
Mikolov, T., Chen, K., Corrado, G., and Dean, J. (2013).
\newblock Efficient {E}stimation of {W}ord {R}epresentations in {V}ector
  {S}pace.
\newblock {\em arXiv preprint arXiv:1301.3781}.

\bibitem[Mosler, 2013]{mosler}
Mosler, K. (2013).
\newblock Depth statistics.
\newblock {\em Robustness and complex data structures}.

\bibitem[Mosler and Mozharovskyi, 2021]{MoslerM21}
Mosler, K. and Mozharovskyi, P. (2021).
\newblock Choosing among notions of depth for multivariate data.
\newblock {\em Statistical Science}.
\newblock In press.

\bibitem[Mozharovskyi et~al., 2015]{MozharovskyiML15}
Mozharovskyi, P., Mosler, K., and Lange, T. (2015).
\newblock Classifying real-world data with the ${DD\alpha}$-procedure.
\newblock {\em Advances in Data Analysis and Classification}, 9(3):287--314.

\bibitem[Mukherjee et~al., 2020]{mukherjee2020outlierrobust}
Mukherjee, D., Guha, A., Solomon, J., Sun, Y., and Yurochkin, M. (2020).
\newblock Outlier-robust optimal transport.
\newblock {\em arXiv preprint arXiv:2012.07363}.

\bibitem[Nagy, 2019]{nagycharacterize}
Nagy, S. (2019).
\newblock Halfspace depth does not characterize probability distributions.
\newblock {\em Statistical Papers}, 26(3):1135--1139.

\bibitem[Nagy and Dvořák, 2021]{illumination}
Nagy, S. and Dvořák, J. (2021).
\newblock Illumination depth.
\newblock {\em Journal of Computational and Graphical Statistics},
  30(1):78--90.

\bibitem[Nagy et~al., 2020]{NagyDM20}
Nagy, S., Dyckerhoff, R., and Mozharovskyi, P. (2020).
\newblock Uniform convergence rates for the approximated halfspace and
  projection depth.
\newblock {\em Electronic Journal of Statistics}, 14(2):3939--3975.

\bibitem[Narayan et~al., 2018]{narayan2018ranking}
Narayan, S., Cohen, S.~B., and Lapata, M. (2018).
\newblock Ranking sentences for extractive summarization with reinforcement
  learning.
\newblock {\em arXiv preprint arXiv:1802.08636}.

\bibitem[Nenkova et~al., 2007]{nenkova2007pyramid}
Nenkova, A., Passonneau, R., and McKeown, K. (2007).
\newblock The pyramid method: Incorporating human content selection variation
  in summarization evaluation.
\newblock {\em ACM Transactions on Speech and Language Processing (TSLP)},
  4(2):4--es.

\bibitem[Nenkova and Passonneau, 2004]{nenkova2004evaluating}
Nenkova, A. and Passonneau, R.~J. (2004).
\newblock Evaluating content selection in summarization: The pyramid method.
\newblock In {\em Proceedings of the human language technology conference of
  the north american chapter of the association for computational linguistics:
  Hlt-naacl 2004}, pages 145--152.

\bibitem[Nieto-Reyes and Battey, 2016]{NietoRB16}
Nieto-Reyes, A. and Battey, H. (2016).
\newblock A topologically valid definition of depth for functional data.
\newblock {\em Statistical Science}, 31(1):61--79.

\bibitem[Oja, 1983]{Oja}
Oja, H. (1983).
\newblock Descriptive statistics for multivariate distributions.
\newblock {\em Statistics and Probability Letters}.

\bibitem[Paindaveine and bever, 2013]{localdepth}
Paindaveine, D. and bever, G.~V. (2013).
\newblock From depth to local depth: A focus on centrality.
\newblock {\em Journal of the American Statistical Association},
  108(503):1105--1119.

\bibitem[Paindaveine and Van~Bever, 2018]{paindaveine2018}
Paindaveine, D. and Van~Bever, G. (2018).
\newblock Halfspace depths for scatter, concentration and shape matrices.
\newblock {\em The Annals of Statistics}, 46(6B):3276--3307.

\bibitem[Panaretos and Zemel, 2019]{panaretosW}
Panaretos, V.~M. and Zemel, Y. (2019).
\newblock Statistical aspects of wasserstein distances.
\newblock {\em Annual Review of Statistics and Its Application}, 6(1):405--431.

\bibitem[Papineni et~al., 2002]{bleu}
Papineni, K., Roukos, S., Ward, T., and Zhu, W.-J. (2002).
\newblock {B}leu: a method for automatic evaluation of machine translation.
\newblock In {\em Proceedings of the 40th Annual Meeting of the Association for
  Computational Linguistics}, pages 311--318.

\bibitem[Paty and Cuturi, 2019]{paty19}
Paty, F.-P. and Cuturi, M. (2019).
\newblock Subspace robust {W}asserstein distances.
\newblock In {\em Proceedings of the 36th International Conference on Machine
  Learning}, volume~97, pages 5072--5081.

\bibitem[Pennington et~al., 2014]{pennington2014glove}
Pennington, J., Socher, R., and Manning, C. (2014).
\newblock {G}lo{V}e: {G}lobal {V}ectors for {W}ord {R}epresentation.
\newblock In {\em Proceedings of the 2014 EMNLP ({EMNLP})}, pages 1532--1543.
  ACL.

\bibitem[Perez-Beltrachini et~al., 2016]{perez2016building}
Perez-Beltrachini, L., Sayed, R., and Gardent, C. (2016).
\newblock Building rdf content for data-to-text generation.
\newblock In {\em The 26th International Conference on Computational
  Linguistics (COLING 2016)}.

\bibitem[Peters et~al., 2018]{elmo}
Peters, M.~E., Neumann, M., Iyyer, M., Gardner, M., Clark, C., Lee, K., and
  Zettlemoyer, L. (2018).
\newblock Deep contextualized word representations.
\newblock In {\em Proc. of NAACL}.

\bibitem[Peyré and Cuturi, 2019]{Peyre}
Peyré, G. and Cuturi, M. (2019).
\newblock Computational optimal transport.
\newblock {\em Foundations and Trends® in Machine Learning}, 11(5-6):355--607.

\bibitem[Pokotylo et~al., 2019]{PokotyloMD19}
Pokotylo, O., Mozharovskyi, P., and Dyckerhoff, R. (2019).
\newblock Depth and depth-based classification with {R-Package} ddalpha.
\newblock {\em Journal of Statistical Software, Articles}, 91(5):1--46.

\bibitem[Rabin et~al., 2012]{sliced}
Rabin, J., Peyr{\'e}, G., Delon, J., and Bernot, M. (2012).
\newblock Wasserstein barycenter and its application to texture mixing.
\newblock In Bruckstein, A.~M., ter Haar~Romeny, B.~M., Bronstein, A.~M., and
  Bronstein, M.~M., editors, {\em Scale Space and Variational Methods in
  Computer Vision}, pages 435--446, Berlin, Heidelberg. Springer Berlin
  Heidelberg.

\bibitem[Rachev, 1991]{rachev1991probability}
Rachev, S. (1991).
\newblock {\em Probability Metrics and the Stability of Stochastic Models}.
\newblock Wiley Series in Probability and Statistics - Applied Probability and
  Statistics Section. Wiley.

\bibitem[Raffel et~al., 2019]{raffel2019exploring}
Raffel, C., Shazeer, N., Roberts, A., Lee, K., Narang, S., Matena, M., Zhou,
  Y., Li, W., and Liu, P.~J. (2019).
\newblock Exploring the limits of transfer learning with a unified text-to-text
  transformer.
\newblock {\em arXiv preprint arXiv:1910.10683}.

\bibitem[Ramsay et~al., 2019]{IRW}
Ramsay, K., Durocher, S., and Leblanc, A. (2019).
\newblock Integrated rank-weighted depth.
\newblock {\em Journal of Multivariate Analysis}, 173:51--69.

\bibitem[Rankel et~al., 2013]{rankel2013decade}
Rankel, P.~A., Conroy, J., Dang, H.~T., and Nenkova, A. (2013).
\newblock A decade of automatic content evaluation of news summaries:
  Reassessing the state of the art.
\newblock In {\em Association for Computational Linguistics (ACL)}, pages
  131--136.

\bibitem[Rousseeuw and Hubert, 1999]{rousseeuw1999}
Rousseeuw, P.~J. and Hubert, M. (1999).
\newblock Regression depth.
\newblock {\em Journal of the American Statistical Association},
  94(446):388--402.

\bibitem[Rousseeuw and Hubert, 2018]{rousseeuw2018}
Rousseeuw, P.~J. and Hubert, M. (2018).
\newblock Anomaly detection by robust statistics.
\newblock {\em WIREs Data Mining and Knowledge Discovery}, 8(2):1236.

\bibitem[Rousseeuw and Rutz, 1999]{rousseeuwruts}
Rousseeuw, P.~J. and Rutz, I. (1999).
\newblock The depth function of a population distribution.
\newblock {\em Metrika}, 49(3):213--244.

\bibitem[Rousseeuw and Struyf, 1998]{RousseeuwS98}
Rousseeuw, P.~J. and Struyf, A. (1998).
\newblock Computing location depth and regression depth in higher dimensions.
\newblock {\em Statistics and Computing}, 8(3):193--203.

\bibitem[Rényi, 1961]{renyi1961}
Rényi, A. (1961).
\newblock On measures of entropy and information.
\newblock In {\em Proceedings of the 4th Berkeley Symposium on Mathematical
  Statistics and Probability, Volume 1: Contributions to the Theory of
  Statistics}, pages 547--561, Berkeley, Calif. University of California Press.

\bibitem[Schneider, 1993]{schneider1993}
Schneider, R. (1993).
\newblock {\em Convex Bodies: The Brunn-Minkowski Theory}.
\newblock Cambridge University Press, Cambridge.

\bibitem[See et~al., 2017]{see2017get}
See, A., Liu, P.~J., and Manning, C.~D. (2017).
\newblock Get to the point: Summarization with pointer-generator networks.
\newblock {\em arXiv preprint arXiv:1704.04368}.

\bibitem[Serfling, 2006]{ser2006}
Serfling, R. (2006).
\newblock Depth functions in nonparametric multivariate inference.
\newblock {\em DIMACS Series in Discrete Mathematics and Theoretical Computer
  Science}, 72.

\bibitem[Shannon, 1948]{shannon}
Shannon, C.~E. (1948).
\newblock A mathematical theory of communication.
\newblock {\em The Bell System Technical Journal}, 27(3):379--423.

\bibitem[Shi and Malik, 2000]{spectralclustering}
Shi, J. and Malik, J. (2000).
\newblock Normalized cuts and image segmentation.
\newblock {\em IEEE Transactions on Pattern Analysis \& Machine Intelligence},
  22(08):888--905.

\bibitem[Snover et~al., 2006]{snover2006study}
Snover, M., Dorr, B., Schwartz, R., Micciulla, L., and Makhoul, J. (2006).
\newblock A study of translation edit rate with targeted human annotation.
\newblock In {\em Proceedings of the 7th Conference of the Association for
  Machine Translation in the Americas: Technical Papers}, pages 223--231.

\bibitem[Specia et~al., 2010]{specia2010machine}
Specia, L., Raj, D., and Turchi, M. (2010).
\newblock Machine translation evaluation versus quality estimation.
\newblock {\em Machine translation}, 24(1):39--50.

\bibitem[Sriperumbudur et~al., 2012]{sriperumbudurIPM}
Sriperumbudur, B.~K., Fukumizu, K., Gretton, A., Schölkopf, B., and Lanckriet,
  G. R.~G. (2012).
\newblock {On the empirical estimation of integral probability metrics}.
\newblock {\em Electronic Journal of Statistics}, 6:1550 -- 1599.

\bibitem[Staerman, 2022]{phdguigui}
Staerman, G. (2022).
\newblock {\em Functional anomaly detection and robust estimation}.
\newblock PhD thesis, Institut polytechnique de Paris.

\bibitem[Staerman et~al., 2022]{staerman2022functional}
Staerman, G., Adjakossa, E., Mozharovskyi, P., Hofer, V., Gupta, J.~S., and
  Cl{\'e}men{\c{c}}on, S. (2022).
\newblock Functional anomaly detection: a benchmark study.
\newblock {\em arXiv preprint arXiv:2201.05115}.

\bibitem[Staerman et~al., 2021a]{staerman2020ot}
Staerman, G., Laforgue, P., Mozharovskyi, P., and d'Alch{\'e} Buc, F. (2021a).
\newblock When ot meets mom: Robust estimation of wasserstein distance.
\newblock In {\em Proceedings of The 24th International Conference on
  Artificial Intelligence and Statistics}, volume 130, pages 136--144.

\bibitem[Staerman et~al., 2020]{staerman2020}
Staerman, G., Mozharovskyi, P., and Cl\'emençon, S. (2020).
\newblock The area of the convex hull of sampled curves: a robust functional
  statistical depth measure.
\newblock In {\em Proceedings of the 23nd International Conference on
  Artificial Intelligence and Statistics}, volume 108, pages 570--579.

\bibitem[Staerman et~al., 2019]{FIF}
Staerman, G., Mozharovskyi, P., Cl\'{e}mençon, S., and d'Alch\'{e} Buc, F.
  (2019).
\newblock Functional isolation forest.
\newblock In {\em Proceedings of The 11th Asian Conference on Machine
  Learning}.

\bibitem[Staerman et~al., 2021b]{AIIRW}
Staerman, G., Mozharovskyi, P., and Clémençon, S. (2021b).
\newblock Affine-invariant integrated rank-weighted depth: Definition,
  properties and finite sample analysis.
\newblock {\em arXiv preprint arXiv:2106.11068}.

\bibitem[Stahel, 1981]{stahel}
Stahel, W.~A. (1981).
\newblock Breakdown of covariance estimators.
\newblock Technical report, Fachgruppe für Statistik, ETH, Zürich.

\bibitem[Stanchev et~al., 2019]{stanchev2019eed}
Stanchev, P., Wang, W., and Ney, H. (2019).
\newblock Eed: Extended edit distance measure for machine translation.
\newblock In {\em Proceedings of the Fourth WMT (Volume 2: Shared Task Papers,
  Day 1)}, pages 514--520.

\bibitem[Stummer and Vajda, 2012]{stummer2012}
Stummer, W. and Vajda, I. (2012).
\newblock On bregman distances and divergences of probability measures.
\newblock {\em IEEE Transactions on Information Theory}, 58(3):1277 -- 1288.

\bibitem[Tukey, 1975]{Tukey75}
Tukey, J.~W. (1975).
\newblock Mathematics and the picturing of data.
\newblock In James, R., editor, {\em Proceedings of the International Congress
  of Mathematicians}, volume~2, pages 523--531. Canadian Mathematical Congress.

\bibitem[Villani, 2003]{Villani}
Villani, C. (2003).
\newblock {\em Topics in Optimal Transportation}.
\newblock Graduate Studies in Mathematics Series. American Mathematical
  Society, New York.

\bibitem[Wang et~al., 2020]{wang2020heterogeneous}
Wang, D., Liu, P., Zheng, Y., Qiu, X., and Huang, X. (2020).
\newblock Heterogeneous graph neural networks for extractive document
  summarization.
\newblock {\em arXiv preprint arXiv:2004.12393}.

\bibitem[Wang et~al., 2016]{wang2016character}
Wang, W., Peter, J.-T., Rosendahl, H., and Ney, H. (2016).
\newblock Character: Translation edit rate on character level.
\newblock In {\em Proceedings of the First WMT: Volume 2, Shared Task Papers},
  pages 505--510.

\bibitem[Wen et~al., 2015]{wen2015semantically}
Wen, T.-H., Ga{\v{s}}i{\'c}, M., Mrk{\v{s}}i{\'c}, N., Su, P.-H., Vandyke, D.,
  and Young, S. (2015).
\newblock Semantically conditioned {LSTM}-based natural language generation for
  spoken dialogue systems.
\newblock In {\em Proceedings of the 2015 Conference on Empirical Methods in
  Natural Language Processing}, pages 1711--1721.

\bibitem[Witon et~al., 2018]{witon2018disney}
Witon, W., Colombo, P., Modi, A., and Kapadia, M. (2018).
\newblock Disney at iest 2018: Predicting emotions using an ensemble.
\newblock In {\em Proceedings of the 9th Workshop on Computational Approaches
  to Subjectivity, Sentiment and Social Media Analysis}, pages 248--253.

\bibitem[Wolf et~al., 2019]{wolf2019huggingface}
Wolf, T., Debut, L., Sanh, V., Chaumond, J., Delangue, C., Moi, A., Cistac, P.,
  Rault, T., Louf, R., Funtowicz, M., et~al. (2019).
\newblock Huggingface's transformers: State-of-the-art natural language
  processing.
\newblock {\em arXiv preprint arXiv:1910.03771}.

\bibitem[Yoon et~al., 2020]{yoon2020learning}
Yoon, W., Yeo, Y.~S., Jeong, M., Yi, B.-J., and Kang, J. (2020).
\newblock Learning by semantic similarity makes abstractive summarization
  better.
\newblock {\em arXiv preprint arXiv:2002.07767}.

\bibitem[Zhang et~al., 2020]{zhang2020pegasus}
Zhang, J., Zhao, Y., Saleh, M., and Liu, P. (2020).
\newblock Pegasus: Pre-training with extracted gap-sentences for abstractive
  summarization.
\newblock In {\em Proceedings of the 37th International Conference on Machine
  Learning}, volume 119, pages 11328--11339.

\bibitem[Zhang et~al., 2019]{zhang2019bertscore}
Zhang, T., Kishore, V., Wu, F., Weinberger, K.~Q., and Artzi, Y. (2019).
\newblock Bertscore: Evaluating text generation with bert.
\newblock {\em arXiv preprint arXiv:1904.09675}.

\bibitem[Zhao et~al., 2019]{zhao2019moverscore}
Zhao, W., Peyrard, M., Liu, F., Gao, Y., Meyer, C.~M., and Eger, S. (2019).
\newblock Moverscore: Text generation evaluating with contextualized embeddings
  and earth mover distance.
\newblock {\em arXiv preprint arXiv:1909.02622}.

\bibitem[Zhong et~al., 2020]{zhong2020extractive}
Zhong, M., Liu, P., Chen, Y., Wang, D., Qiu, X., and Huang, X. (2020).
\newblock Extractive summarization as text matching.
\newblock {\em arXiv preprint arXiv:2004.08795}.

\bibitem[Zhong et~al., 2019]{zhong2019searching}
Zhong, M., Liu, P., Wang, D., Qiu, X., and Huang, X. (2019).
\newblock Searching for effective neural extractive summarization: What works
  and what's next.
\newblock {\em arXiv preprint arXiv:1907.03491}.

\bibitem[Zhou et~al., 2018]{zhou2018neural}
Zhou, Q., Yang, N., Wei, F., Huang, S., Zhou, M., and Zhao, T. (2018).
\newblock Neural document summarization by jointly learning to score and select
  sentences.
\newblock {\em arXiv preprint arXiv:1807.02305}.

\bibitem[Zuo, 2003]{zuo}
Zuo (2003).
\newblock Projected based depth functions and associated medians.
\newblock {\em The annals of statistics}, 31(5):1460--1490.

\bibitem[Zuo and Serfling, 2000]{ZuoSerfling00}
Zuo, B. and Serfling, R. (2000).
\newblock General notions of statistical depth function.
\newblock {\em The Annals of Statistics}, 28(2):461--482.

\end{thebibliography}
%

\newpage
\appendix
\onecolumn

\clearpage

This Appendix is organized as follows:
\vspace{0.2cm}
\begin{itemize}
\item Appendix~\ref{sec:app:pre} contains additional notations as well as useful preliminary results.
\vspace{0.1cm}
\item Appendix~\ref{sec:app:preuves} contains the proofs of the propositions/theorems provided in the paper.
\vspace{0.1cm}
\item Appendix~\ref{approxalgo} contains approximation algorithms to compute halfspace/projection/AI-IRW depth.
\vspace{0.1cm}
\item Appendix~\ref{sec:app:toy} contains additional synthetic experiments.
\vspace{0.1cm}
\item Appendix~\ref{sec:suplementary_nlp} contains details on experimental settings of NLP applications.

\end{itemize}

\section{PRELIMINARY RESULTS}\label{sec:app:pre}
First, we introduce additional notations and recall some lemmas, used in the subsequent proofs.
\subsection{Hausdorff Distance}\label{hauss}
The Hausdorff  distance between two bounded subspaces  $\mathcal{K}_1,\mathcal{K}_2$ of $\mathbb{R}^d$ is defined as

\begin{equation*} \label{hauss}
d_{\mathcal{H}}(\mathcal{K}_1,\mathcal{K}_2)=\max \left\{ \underset{x\in \mathcal{K}_1}{\sup}\; \underset{y\in \mathcal{K}_2}{\inf}\; ||x-y|| ,\; \; \underset{y\in \mathcal{K}_2}{\sup} \;\underset{x\in \mathcal{K}_1}{\inf}\; ||x-y|| \right\}.
\end{equation*}

Furthermore, if $\mathcal{K}_1$ and $\mathcal{K}_2$ are convex bodies (i.e.  non empty compact convex sets), the Hausdorff distance can be reformulated with support functions of $\mathcal{K}_1,\mathcal{K}_2$:
\begin{equation*}
d_{\mathcal{H}}(\mathcal{K}_1,\mathcal{K}_2)= \supS \big| h_{\mathcal{K}_1}(u) -  h_{\mathcal{K}_2}(u) \big|,
\end{equation*}

where $h_{\mathcal{K}_1}(u)=\sup \{ \langle u,x \rangle , \; \; x\in \mathcal{K}_1\} $.

\subsection{Quantile regions}
Let $u\in \mathbb{S}^{d-1}$ and $X\sim \mu$ where $\mu \in \mathcal{M}_1(\mathcal{X})$ with $\mathcal{X}\subset \mathbb{R}^d$. We define the  $(1-\beta)$ directional quantile of a distribution $\mu$ in the direction $u$ as 
\begin{align}
q_{\mu, u}^{{\scriptscriptstyle 1-\beta}}= \inf \left\{ t\in \mathbb{R}: \; \; \; \mathbb{P}\left(\langle u,X \rangle\leq t \right)\geq 1-\beta \right\}
\end{align}

and the upper $(1-\beta)$ quantile set of $\mu$
\begin{align}
Q_{\mu}^{{\scriptscriptstyle1-\beta}} = \left\{ x\in \mathbb{R}^d: \; \; \langle u,x \rangle \leq q_{\mu, u}^{{\scriptscriptstyle 1-\beta}} , \; \; \;  \forall \; u\in \mathbb{S}^{d-1} \right\}.
\end{align}

\subsection{Auxiliary results}\label{aux}
We now recall useful results, so as to characterize the halfspace depth regions. 

\begin{lemma}[\citealp{bruneltukey}, Lemma 1]\label{brunellemma}
Let $\mu \in \mathcal{M}_1(\mathcal{X})$, for any $\beta \in (0,1)$, it holds:
$D_{\mu}^{\beta}=Q_{\mu}^{{\scriptscriptstyle1-\beta}} $.
\end{lemma}

\begin{lemma}[\citealp{bruneltukey}, Proposition 1] \label{brunellemma2} Let $\mu \in \mathcal{M}_1(\mathcal{X})$ with a $(1-\beta)$ directional quantile $q_{\mu, u}^{{\scriptscriptstyle 1-\beta}}$ for any $u\in \mathbb{S}^{d-1}$. Assume that $u\mapsto q_{\mu, u}^{{\scriptscriptstyle 1-\beta}}$ are sublinear, i.e., $q_{\mu, u+\lambda v}^{{\scriptscriptstyle 1-\beta}}\leq q_{\mu, u}^{{\scriptscriptstyle 1-\beta}} + \lambda~ q_{\mu, v}^{{\scriptscriptstyle 1-\beta}}, \; \; \forall \; \lambda >0$. Then for any $u\in \mathbb{S}^{d-1}$, it holds $h_{Q_{\mu,u}^{{\scriptscriptstyle1-\beta}}}(u)=q_{\mu, u}^{{\scriptscriptstyle 1-\beta}}$.
\end{lemma}

\begin{lemma}\label{dimensionone}
Let $d=1$ and $X^{1}\sim \mu_1,\;Y^{1} \sim \nu_1$ be two random variables where $\mu_1,\nu_1$ are univariate probability distributions. Denoting by $F^{-1}_{X^1}$ the quantile function of $X^1$, then the depth-trimmed region based pseudo-metric (associated with the halfspace depth) is defined as 
\begin{align*}
    DR_{p,\varepsilon}^p(\mu_1,\nu_1)= 2\int_{\varepsilon/2}^{1/2} \max \Big\{ &| F^{-1}_{X^1} (q) -F^{-1}_{Y^1}(q) |^p, \;  | F^{-1}_{X^1} (1-q) -F^{-1}_{Y^1}(1-q) |^p \Big \} \; \mathrm{d}q.
\end{align*}
\end{lemma}

\begin{proof}
In dimension one,  the halfspace depth of any $t\in \mathbb{R}$ w.r.t. $\mu_1$ and $\nu_1$ boils down to

\begin{align*}
    D(t,\mu_1) = \min \Big \{ F_{X^1} (t), 1- F_{X^1} (t) \Big \} \quad  \text{and} \quad    D(t,\nu_1) = \min \Big \{ F_{Y^1} (t), 1- F_{Y^1} (t) \Big \},
\end{align*}

and for any $\gamma \in [0,1]$, its upper-level sets  to intervals

\begin{align}\label{region}
    D^{\gamma}_{\mu_1} = [F_{X^1}^{-1}(\gamma), \;  F^{-1}_{X^1}(1-\gamma)] \quad  \text{and} \quad    D^{\gamma}_{\nu_1} = [F_{Y^1}^{-1}(\gamma), \;  F^{-1}_{Y^1}(1-\gamma)].
\end{align}

Now, the quantile function $\alpha (\beta, .)$ can be explicitly derived as function of $\beta \in [0,1]$:

\begin{align*}
\alpha (\beta, \mu_1)&= \sup \Big \{ \gamma \in [0,1]: \; \;  \mu_1 \Big ([F_{X^1}^{-1}(\gamma), \;  F^{-1}_{X^1}(1-\gamma)] \Big )  \geq  \beta\Big \} \\& =\sup \Big \{ \gamma \in [0,1]: \; \;  1-2\gamma  \geq  \beta\Big \} \\& =\frac{1-\beta}{2}.
\end{align*}
Following the same reasoning, it holds $\alpha (\beta, \nu_1)=\frac{1-\beta}{2}$. Further, by change of variables 

\begin{align*}
    \int_{0}^{1-\varepsilon} d_{\mathcal{H}} \left( D^{(1-\beta) /2}_{\mu_1}, D^{(1-\beta) /2}_{\nu_1}\right)^p~\mathrm{d}\beta = 2\int_{\varepsilon/2}^{1/2} d_{\mathcal{H}} \left( D^{q}_{\mu_1}, D^{q}_{\nu_1}\right)^p~\mathrm{d}q.
\end{align*}

Combining \Cref{region} and the Hausdorff distance definition recalled in Section~\ref{hauss} lead to the result.

\end{proof}

\section{TECHNICAL PROOFS}\label{sec:app:preuves}

We now prove the main results stated in the paper.

\subsection{Proof of Proposition \ref{distance}}

For any $0\leq \beta \leq 1-\varepsilon  $ with $\varepsilon \in (0,1]$, and any $\mu \in \mathcal{M}_1(\mathcal{X})$, $\nu \in \mathcal{M}_1(\mathcal{Y})$, $D^{\alpha(\beta)}_{\mu},D^{\alpha(\beta)}_{\nu}$ are non-empty compact subsets of $\mathbb{R}^d$ due to the properties (\textbf{D2-D3}). The Hausdorff distance $d_{\mathcal{H}}$, recalled in Section~\ref{hauss}, is known to be a distance on the space of non-empty compact sets which implies that $DR_{p,\varepsilon}$ satisfies positivity, symmetry and the triangle inequality (thanks to Minkowski inequality). If $\mu=\nu$ then $D^{\alpha(\beta)}_{\mu}=D^{\alpha(\beta)}_{\nu},\quad \forall \; \beta \in [0, 1-\varepsilon]  $ which leads to $DR_{p,\varepsilon}(\mu,\nu)=0$. The reverse is not true. $DR_{p,\varepsilon}(\mu,\nu)=0$ implies $D^{\alpha(\beta)}_{\mu}=D^{\alpha(\beta)}_{\nu}, \quad \forall \; \beta \in [0, 1-\varepsilon]  $ that not leads to $\mu=\nu$. Indeed, convex depth regions do not characterize probability distributions in general (see \citealp{nagycharacterize} for the halfspace depth)  that would be the first step in order to prove the previous entailment.



\subsection{Proof of Proposition \ref{affine}}
Let $A \in \mathbb{R}^{d\times d}$ be a non-singular matrix and $b\in \mathbb{R}^d$ such that $g: x\mapsto Ax+b$. Then, it holds:

\begin{align}\label{cc}
~DR^p_{p,\varepsilon}(g_{\sharp}\mu, g_{\sharp}\nu) &= \int_{0}^{1-\varepsilon} \left[d_{\mathcal{H}}(D_{g_{\sharp}\mu}^{\alpha(\beta)}, D_{g_{\sharp}\nu}^{\alpha(\beta)})\right]^p \;\mathrm{d}\beta \nonumber \\& \overset{(i)}{=}  \int_{0}^{1-\varepsilon} \left[d_{\mathcal{H}}(AD_{\mu}^{\alpha(\beta)}+b, AD_{\nu}^{\alpha(\beta)}+b)\right]^p \;\mathrm{d}\beta,
\end{align}
where ($i$) holds because any data depth satisfies (\textbf{D1}) by definition. Furthermore,
\begin{align*}
d_{\mathcal{H}}(AD_{\mu}^{\alpha(\beta)}+b, AD_{\nu}^{\alpha(\beta)}+b)&=\max \left\{ \underset{x\in D_{\mu}^{\alpha(\beta)}}{\sup}\; \underset{y\in D_{\nu}^{\alpha(\beta)}}{\inf}\; ||Ax-Ay|| ,\; \; \underset{y\in D_{\nu}^{\alpha(\beta)}}{\sup} \;\underset{x\in D_{\mu}^{\alpha(\beta)}}{\inf}\; ||Ax-Ay|| \right\} \\& \overset{(ii)}{=} \max \left\{ \underset{x\in D_{\mu}^{\alpha(\beta)}}{\sup}\; \underset{y\in D_{\nu}^{\alpha(\beta)}}{\inf}\; ||x-y|| ,\; \; \underset{y\in D_{\nu}^{\alpha(\beta)}}{\sup} \;\underset{x\in D_{\mu}^{\alpha(\beta)}}{\inf}\; ||x-y||  \right\} \\& =d_{\mathcal{H}}(D_{\mu}^{\alpha(\beta)}, D_{\nu}^{\alpha(\beta)}),
\end{align*} 

where ($ii$) holds by virtue of hypothesis $AA^\top =I_d$. Replacing it in (\ref{cc}) yields the desired results.
\subsection{Proof of Proposition \ref{general}}

\noindent {\bf First assertion.}
Denote $Z_1,Z_2$ two random variables following $\mu^*,\nu^*$ respectively. Assume that $X,Y,Z_1,Z_2$ are defined on the probability space $\left(\Omega, \mathcal{A},\mathbb{P} \right)$. For any $x\in \mathbb{R}^d$ and $\beta\in [0,1-\varepsilon]$,

\begin{align*}
x\in D_{\mu}^{\alpha(\beta)} \Longleftrightarrow  HD_{\mu}(x) \geq \alpha(\beta) & \Longleftrightarrow  \forall \;   u\in \mathbb{S}^{d-1}, \; \; \;  \mathbb{P}\left( \langle u,X\rangle \; \leq \langle u,x\rangle \right) \; \geq \;  \alpha(\beta) \\& \Longleftrightarrow \forall \;   u\in \mathbb{S}^{d-1}, \; \; \; \mathbb{P}\left( \langle u,Z_1 + \mathbf{m}_1\rangle \; \leq \langle u,x\rangle \right) \; \geq  \; \alpha(\beta) \\& \Longleftrightarrow \forall \;   u\in \mathbb{S}^{d-1}, \; \; \; \mathbb{P}\left( \langle u,Z_1\rangle \; \leq \langle u,x-\mathbf{m}_1\rangle \right) \; \geq \; \alpha(\beta) \\& \Longleftrightarrow  x-\mathbf{m}_1 \in D^{\alpha(\beta)}_{\mu^*} 
\end{align*}

The same reasoning holds for $\nu$ and $\nu^*$. Following this, for any $\beta\in [0,\;1-\varepsilon]$ and $u\in \mathbb{S}^{d-1}$, it holds:

\begin{align*}
h_{D^{\alpha(\beta)}_{\mu}} (u)= h_{D^{\alpha(\beta)}_{\mu^*}} (u) - \langle u,\mathbf{m}_1 \rangle \qquad \text{and} \qquad h_{D^{\alpha(\beta)}_{\nu}} (u)= h_{D^{\alpha(\beta)}_{\nu^*}} (u) - \langle u,\mathbf{m}_2 \rangle
\end{align*}
 Thus it holds:
 \begin{align}\label{eq:upper}
 DR^2_{2,\varepsilon}(\mu,\nu) &= \intR \supS \Big| h_{D^{\alpha(\beta)}_{\mu^*}} (u) - \langle u,\mathbf{m}_1 \rangle   -h_{D^{\alpha(\beta)}_{\nu^*}} (u) + \langle u,\mathbf{m}_2 \rangle\Big|^2~\mathrm{d}\beta \nonumber \\&\leq \supS \big| \langle u, \mathbf{m}_1-\mathbf{m}_2\rangle \big|^2 +\intR \supS \big|h_{D^{\alpha(\beta)}_{\mu^*}} (u)- h_{D^{\alpha(\beta)}_{\nu^*}} (u) \big|^2~\mathrm{d}\beta \nonumber  \\& \qquad +2  \supS \big| \langle u, \mathbf{m}_1-\mathbf{m}_2\rangle \big| \intR \supS \big| h_{D^{\alpha(\beta)}_{\mu^*}} (u)-h_{D^{\alpha(\beta)}_{\nu^*}} (u) \big|~\mathrm{d}\beta \nonumber \\&
  =||\mathbf{m}_1-\mathbf{m}_2||^2 + DR^2_{2,\varepsilon}(\mu^*,\nu^*)+ 2||\mathbf{m}_1-\mathbf{m}_2|| DR_{1,\varepsilon}(\mu^*,\nu^*).
 \end{align}

On the other side, we have:
 \begin{align}\label{eq:lower}
 DR^2_{2,\varepsilon}(\mu,\nu) &\geq \supS \big| \langle u, \mathbf{m}_1-\mathbf{m}_2\rangle \big|^2 +\intR \supS \big|h_{D^{\alpha(\beta)}_{\mu^*}} (u)- h_{D^{\alpha}_{\nu^*}} (u) \big|^2 ~\mathrm{d}\beta  \nonumber\\& \qquad -2  \supS \big| \langle u, \mathbf{m}_1-\mathbf{m}_2\rangle \big| \intR \supS \big| h_{D^{\alpha(\beta)}_{\mu^*}} (u)-h_{D^{\alpha(\beta)}_{\nu^*}} (u) \big|~\mathrm{d}\beta  \nonumber\\&
  =||\mathbf{m}_1-\mathbf{m}_2||^2 + DR^2_{2,\varepsilon}(\mu^*,\nu^*)- 2||\mathbf{m}_1-\mathbf{m}_2|| DR_{1,\varepsilon}(\mu^*,\nu^*).
 \end{align}

Combining (\ref{eq:upper}) and (\ref{eq:lower}) lead to the desired result.

\bigskip

\noindent {\bf Second assertion.}
For any $u\in \mathbb{S}^{d-1}$, the ($1-\alpha(\beta)$) quantiles of  random variables $\langle u,X\rangle$ and $\langle u,Y\rangle$ such that $\langle u,X\rangle \sim \mathcal{N}( \langle u,\mathbf{m}_1 \rangle , u^\top \mathbf{\Sigma}_1 u )$ and  $\langle u,Y\rangle \sim \mathcal{N}( \langle u,\mathbf{m}_2 \rangle , u^\top \mathbf{\Sigma}_2 u )$ are defined by

\begin{equation*}
q_{ \mu, u}^{{\scriptscriptstyle 1-\alpha(\beta)}}= \langle u,\mathbf{m}_1 \rangle + \Phi^{-1}(1-\alpha(\beta)) \sqrt{u^\top \mathbf{\Sigma}_1 u} \qquad q_{ \nu, u}^{{\scriptscriptstyle 1-\alpha(\beta)}}= \langle u,\mathbf{m}_2 \rangle + \Phi^{-1}(1-\alpha(\beta)) \sqrt{u^\top \mathbf{\Sigma}_2 u},
\end{equation*} 

where $\Phi$ is the cumulative distribution function of the univariate standard Gaussian distribution. Now, to apply Lemma \ref{brunellemma2}, it is sufficient to prove that directional quantiles are sublinear. It holds using subadditivity of the square root function. Indeed, for any $u,v\in \mathbb{S}^{d-1}$ and $\lambda >0$, we have:

{\small
\begin{align*}
\langle u+\lambda v,\mathbf{m}_1 \rangle + \Phi^{-1}(1-\alpha(\beta))  \sqrt{(u+\lambda v)^{\top} \mathbf{\Sigma}_1 (u+\lambda v)}&=\langle u,\mathbf{m}_1 \rangle + \lambda \langle v,\mathbf{m}_1 \rangle +\Phi^{-1}(1-\alpha(\beta)) \sqrt{(u+\lambda v)^{\top} \mathbf{\Sigma}_1 (u+\lambda v)} \\&\leq \langle u,\mathbf{m}_1 \rangle + \lambda \langle v,\mathbf{m}_1 \rangle +\Phi^{-1}(1-\alpha(\beta))\left[ \sqrt{u^{\top} \mathbf{\Sigma}_1 u} +\lambda \sqrt{v^{\top} \mathbf{\Sigma}_1 v} \right] \\& = q_{ \mu, u}^{{\scriptscriptstyle 1-\alpha(\beta)}} + \lambda \; q_{ \mu, v}^{{\scriptscriptstyle 1-\alpha(\beta)}} .
\end{align*}}

The same reasoning holds for  $\nu$. Applying  Lemma \ref{brunellemma} and Lemma \ref{brunellemma2}, for any $u\in \mathbb{S}^{d-1}$, we have $ h_{D^{\alpha(\beta)}_{\mu}}(u)=q_{\mu,u}^{{\scriptscriptstyle 1-\alpha(\beta)}}$ and 
$ h_{D^{\alpha(\beta)}_{\nu}}(u)=q_{\nu,u}^{{\scriptscriptstyle 1-\alpha(\beta)}}$. It follows:
 

\begin{align*}
DR_{1,\varepsilon}(\mu,\nu) &= \intR d_{\mathcal{H}} \left( D_{\mu}^{\alpha(\beta)},D_{\nu}^{\alpha(\beta)} \right) \; \mathrm{d}\beta  = \intR  \supS \big|h_{D^{\alpha(\beta)}_{\mu}}(u)-h_{D^{\alpha(\beta)}_{\nu}}(u) \big| ~\mathrm{d}\beta \\&
 =  \intR  \supS \Big| \langle u, \mathbf{m}_1- \mathbf{m}_2 \rangle+ \Phi^{-1}(1-\alpha(\beta)) \left[\sqrt{u^\top \mathbf{\Sigma}_1 u} - \sqrt{u^\top \mathbf{\Sigma}_2 u}\right] \Big| ~d\beta
\\& \leq|| \mathbf{m}_1 - \mathbf{m}_2 || +  \intR \supS \Big|\Phi^{-1}(1-\alpha(\beta)) \left[\sqrt{u^\top \mathbf{\Sigma}_1 u} - \sqrt{u^\top \mathbf{\Sigma}_2 u} \right] \Big|  ~\mathrm{d}\beta 
\\& = || \mathbf{m}_1 - \mathbf{m}_2 || +  C_{\varepsilon}~ \supS \big| \sqrt{u^\top \mathbf{\Sigma}_1 u} - \sqrt{u^\top \mathbf{\Sigma}_2 u} \big| ,
\end{align*}
 with $C_{\varepsilon}= \intR\big|\Phi^{-1}(1-\alpha(\beta))\big| ~\mathrm{d}\beta$. The lower bound is obtained by means the same reasoning. Notice that

\begin{align*}
 || \mathbf{m}_1 - \mathbf{m}_2 ||&= \;\supS \big| \langle u,\mathbf{m}_1 - \mathbf{m}_2 \rangle\big| =  \intR  \supS \big| \langle u,\mathbf{m}_1 - \mathbf{m}_2 \rangle\big| ~\mathrm{d}\beta  .
\end{align*}

Introducing $h_{D^{\alpha(\beta)}_{\mu}}(u),h_{D^{\alpha(\beta)}_{\nu}}(u)$ and using triangular inequality, subadditivity of the supremum and linearity of the integral, we obtain:
\begin{align*}
 || \mathbf{m}_1 - \mathbf{m}_2 ||&\leq DR_{1,\varepsilon}(\mu,\nu) + C_{\varepsilon}~ \supS \big| \sqrt{u^\top \mathbf{\Sigma}_1 u} - \sqrt{u^\top \mathbf{\Sigma}_2 u} \big| ,
\end{align*}

which ends the proof.

\subsection{Proof of Proposition \ref{Breakdown}}

For $DR_{p,\varepsilon}$ to break down at $\mathcal{S}_n$, it needs to have at least one trimmed-region that breaks down. Then the breakdown point of $DR_{p,\varepsilon}$ is higher than the minimum of the breakdown point of each region. Indeed, we have

\begin{align*}
BP(DR_{p,\varepsilon}, \mathcal{S}_n) &= \min \left\{ \dfrac{o}{n+ o}: \; \underset{Z_1,\ldots, Z_o }{\sup } \;  DR_{p,\varepsilon} \left(\hat{\mu}_{n+o}, \hat{\mu}_n \right) = +\infty  \right\} \\& \geq \underset{ \beta \in [0, 1-\varepsilon]}{\min } \; \min  \left\{ \dfrac{o}{n+ o}: \; \underset{Z_1,\ldots, Z_o }{\sup } \;  d_{\mathcal{H}} \left(D^{\alpha(\beta,\hat{\mu}_{n+o})}_{\hat{\mu}_{n+o}},D^{\alpha(\beta,\hat{\mu}_n)}_{\hat{\mu}_n} \right) = +\infty \right\}\\& =  \underset{\beta \in [0, 1-\varepsilon]}{\min } \; BP(D^{\alpha(\beta, \hat{\mu}_n)}_{\hat{\mu}_n}, \mathcal{S}_n).
\end{align*}

Now applying Lemma 3.1 in \citet{DonohoGasko} and Theorem 4 in \citet{illumination}, a lower bound of the breakdown point of each halfspace region, for every $\beta \in [0,1-\varepsilon]$, is given by 

\begin{align*}
  BP(D^{\alpha(\beta,\hat{\mu}_n)}_{\hat{\mu}_n}, \mathcal{S}_n) \geq\left\{
  \begin{array}{@{}ll@{}}
    \frac{\lceil n\alpha(1-\varepsilon,\hat{\mu}_n)/ (1-\alpha(1-\varepsilon,\hat{\mu}_n))\rceil}{n +\lceil n\alpha(1-\varepsilon,\hat{\mu}_n) / (1-\alpha(1-\varepsilon,\hat{\mu}_n))\rceil } & \text{if} \;  \alpha(1-\varepsilon,\hat{\mu}_n) \leq  \frac{\alpha_{\max}(\hat{\mu}_n) }{1+\alpha_{\max}(\hat{\mu}_n) },\\[0.4cm]
    \frac{\alpha_{\max}(\hat{\mu}_n) }{1+\alpha_{\max}(\hat{\mu}_n) } & \text{otherwise},
  \end{array}\right. 
\end{align*}

where $\alpha_{\mathrm{max}}( \hat{\mu}_n)=\underset{x\in \mathbb{R}^d}{\max} \; \;  HD_{\hat{\mu}_n}(x)$.




\section{APPROXIMATION ALGORITHMS}\label{approxalgo}

In this part, we display the approximation algorithms of the halfspace depth (see Algorithm \ref{algo::HD}), the projection depth (see Algorithm \ref{algo::PD}) and the AI-IRW depth (see Algorithm \ref{algo::AIIRW}, proposed in \citealp{AIIRW}) used in the first step of the Algorithm \ref{alg:DRWappr}.

\begin{algorithm}[H]
\caption{Approximation of the halfspace depth}
\textit{Initialization:} $\mathbf{X} \in \mathbb{R}^{n\times d}$, $K$.
      \begin{algorithmic}[1]
      \STATE Construct $\mathbf{U}\in \mathbb{R}^{d \times K}$ by sampling uniformly $K$ vectors  $U_1,\ldots,U_K$ in $\mathbb{S}^{d-1}$
      \STATE Compute $\mathbf{M}=\mathbf{XU}$
      \STATE Compute the rank value $\sigma(i,k)$, the rank of index $i$ in $\mathbf{M}_{:,k}$ for every $i\leq n$ and $k\leq K$
      \STATE Set $D_i=\underset{k\leq K}{\min} \; \; \sigma(i,k)$ for every $i\leq n$\\
 \textbf{Output}: $D,\mathbf{M}$
      \end{algorithmic}
      \label{algo::HD}
\end{algorithm}
\begin{algorithm}[H]
\caption{Approximation of the projection depth}
\textit{Initialization:} $\mathbf{X} \in \mathbb{R}^{n\times d}$, $K$.
      \begin{algorithmic}[1]
      \STATE Construct $\mathbf{U}\in \mathbb{R}^{d \times K}$ by sampling uniformly $K$ vectors  $U_1,\ldots,U_K$ in $\mathbb{S}^{d-1}$
      \STATE Compute $\mathbf{M}=\mathbf{XU}$
      \STATE Find  $\;\mathbf{M}_{\text{med},k}$ the median value of $\mathbf{M}_{:,k}$,  $\; \forall \; k\leq K$
      \STATE Compute  $\text{MAD}_k=\text{median}\{ \big \lvert \mathbf{M}_{i,k}-\mathbf{M}_{\text{med},k} \big \rvert, \; i\leq n \}$ for $k\leq K$
      \STATE Compute $\mathbf{V}$ s.t. $\mathbf{V}_{i,k}=\big \lvert \mathbf{M}_{i,k}-\mathbf{M}_{\text{med},k} \big \rvert  / \text{MAD}_k $
      \STATE Set $D_i=\underset{k\leq K}{\min} \; \; 1/(1+\mathbf{V}_{i,k})$ for every $i\leq n$\\
 \textbf{Output}: $D,\mathbf{M}$
      \end{algorithmic}
      \label{algo::PD}
\end{algorithm}

\begin{algorithm}[H]
\caption{Approximation of the AI-IRW depth}
\textit{Initialization:} $\mathbf{X} \in \mathbb{R}^{n\times d}$, $K$.
      \begin{algorithmic}[1]
      \STATE Construct $\mathbf{U}\in \mathbb{R}^{d \times K}$ by sampling uniformly $K$ vectors  $U_1,\ldots,U_K$ in $\mathbb{S}^{d-1}$
      \STATE Compute $\widehat{\Sigma}$ using any estimator
      \STATE Perform Cholesky or SVD on $\widehat{\Sigma}$  to obtain $\widehat{\Sigma}^{-1/2}$
      \STATE Compute $\mathbf{V}= \widehat{\Sigma}^{-1/2}\mathbf{U}/ ||\widehat{\Sigma}^{-1/2}\mathbf{U} ||$
      \STATE Compute $\mathbf{M}=\mathbf{XV}$
      \STATE Compute the rank value $\sigma(i,k)$, the rank of index $i$ in $\mathbf{M}_{:,k}$ for every $i\leq n$ and $k\leq K$
      \STATE Set $D_i=\frac{1}{K} \sum_{k=1}^{K} \; \; \sigma(i,k)$ for every $i\leq n$\\
 \textbf{Output}: $D,\mathbf{M}$
      \end{algorithmic}
      \label{algo::AIIRW}
\end{algorithm}





\section{ADDITIONAL EXPERIMENTS}\label{sec:app:toy}

\subsection{Illustration of data depth contours}

\begin{figure}[!h]
\begin{center}
	\includegraphics[trim=0cm 0cm 0cm 0cm,scale=0.35]{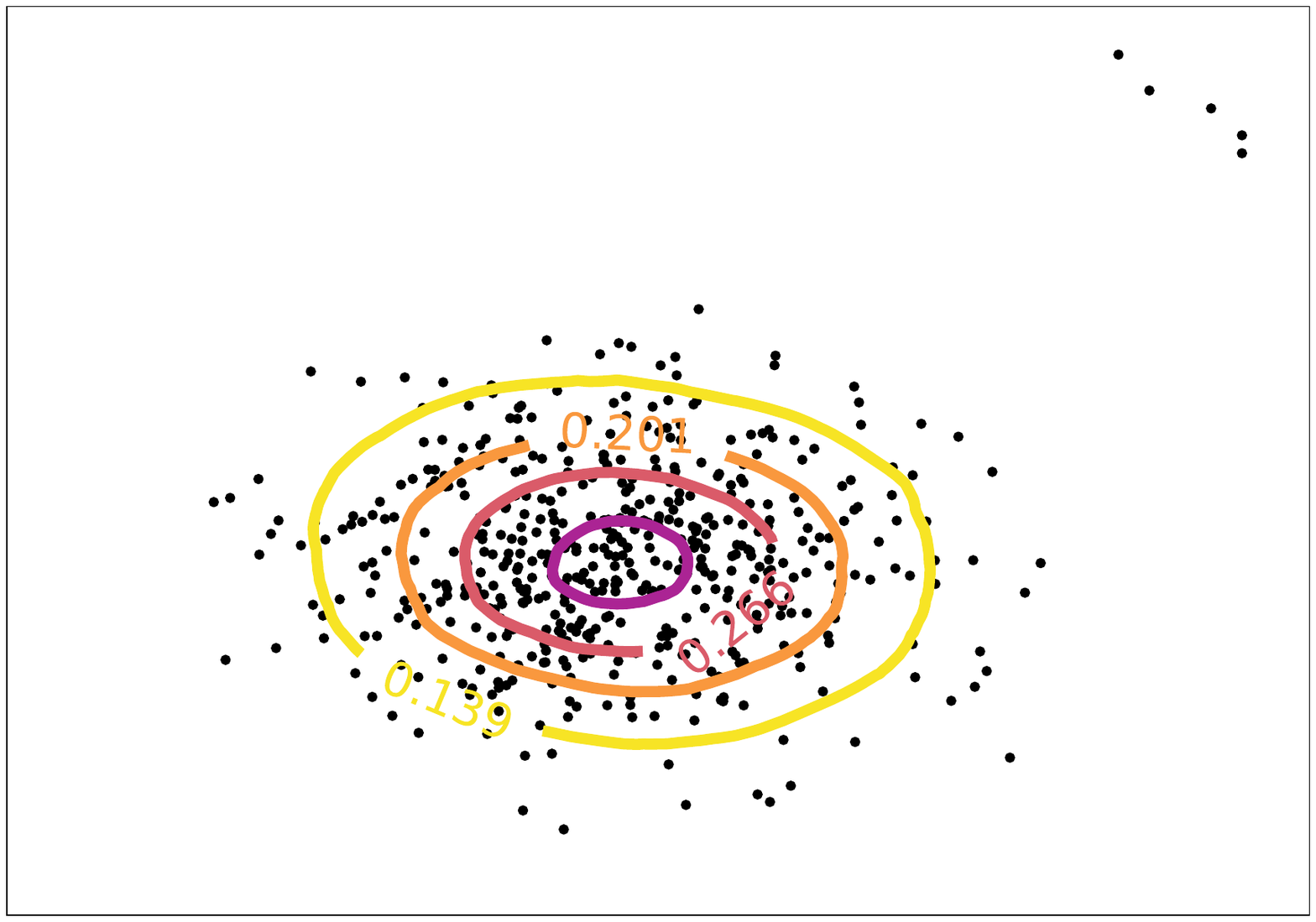}
	\end{center}
	\caption{AI-IRW depth contours for a bivariate sample contaminated with outliers.}
	\label{fig:TukeySample}
\end{figure}

Figure~\ref{fig:TukeySample}, which plots a family of AI-IRW (using MCD estimator) depth  induced trimmed-contours for a dataset contaminated with outliers, illustrates its robustness.

\subsection{Illustration of the depth trimmed-regions based pseudo-Metric}

\begin{figure}[!h]
\begin{center}
	\includegraphics[width= \textwidth]{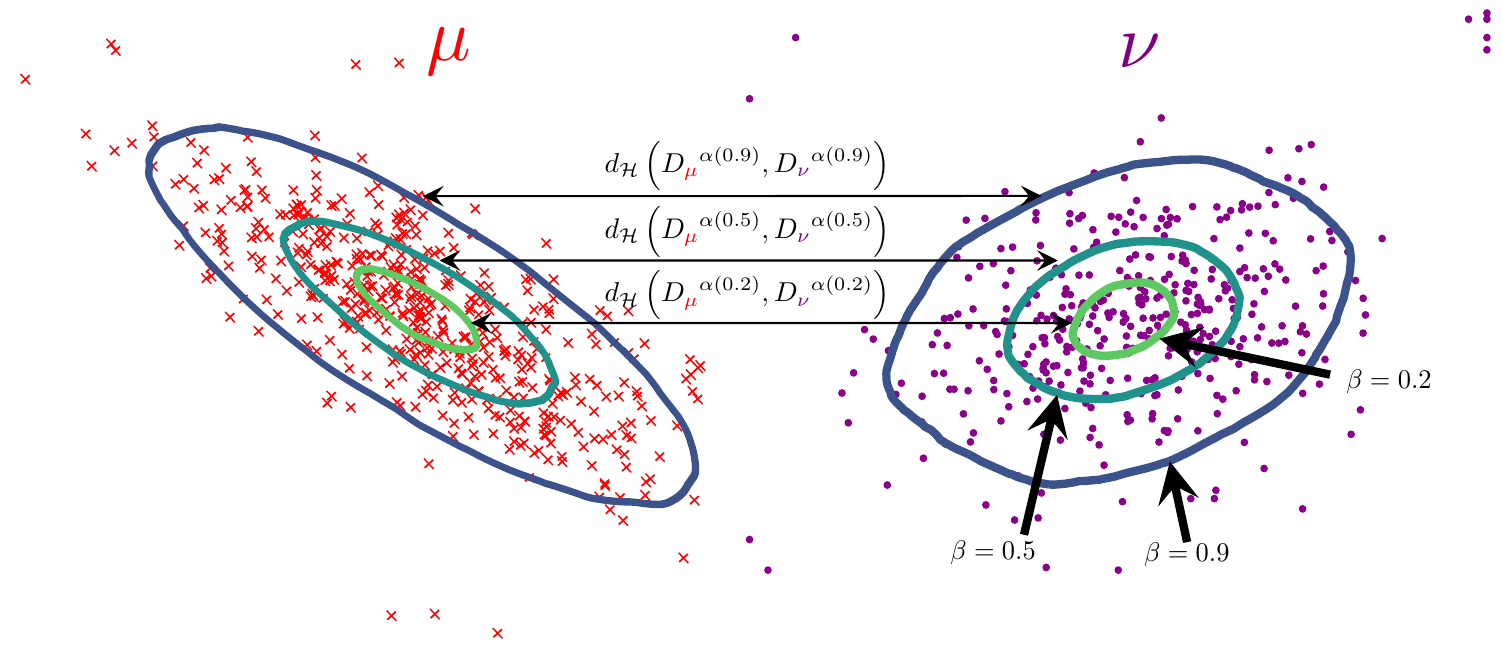}
	\end{center}
	\caption{Illustration of the principle of the depth trimmed-regions based pseudo-metric.}
	\label{fig:baseline}
\end{figure}

Figure~\ref{fig:baseline}, which plots a family of (approximated) AI-IRW depth induced trimmed-regions for two datasets contaminated with outliers,  illustrates the key idea of our proposed pseudo-metric.

\subsection{Empirical analysis of statistical rates}

\begin{figure}[!h]
\begin{center}
\begin{tabular}{cc}
\includegraphics[trim=0cm 0cm 0cm 0cm, scale=0.30]{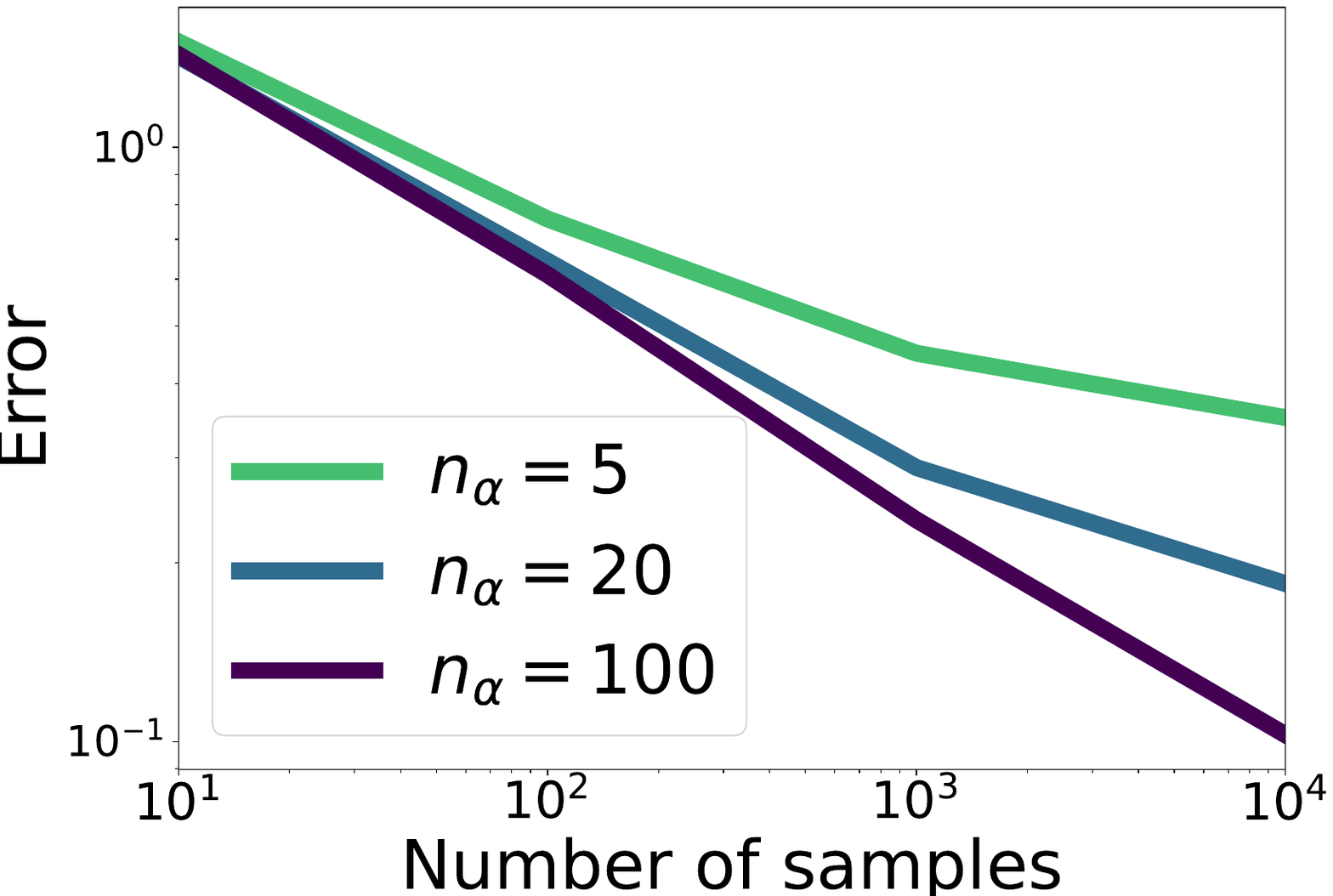} &\hspace{2cm}
\includegraphics[trim=10cm 0cm 0cm 0cm, scale=0.30]{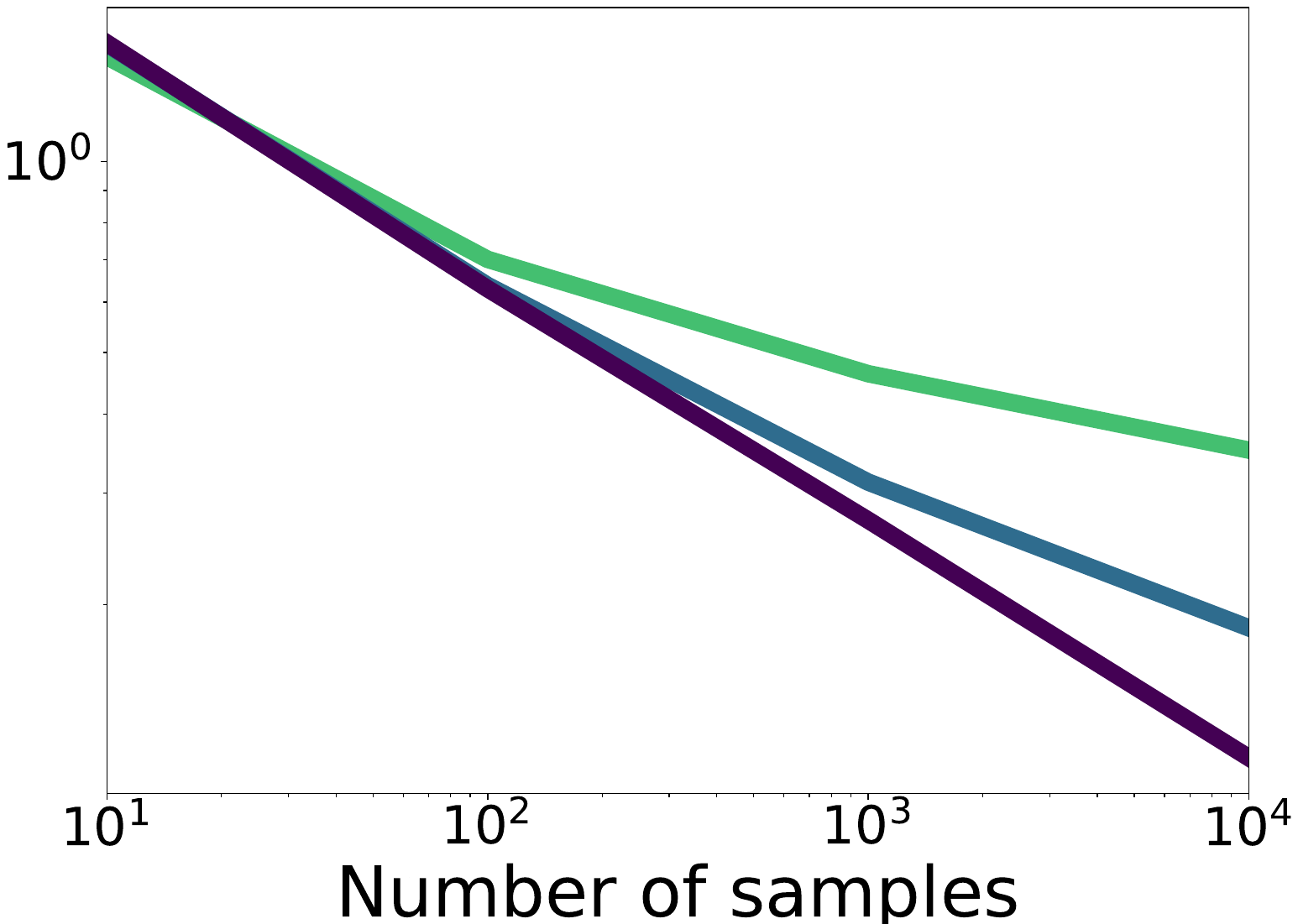}
\end{tabular}
\end{center}
\caption{Empirical analysis of statistical convergence rates. Resulting error of the proposed pseudo-metric when increasing the sample size using the projection depth (left) and the halfspace depth (right) for various $n_{\alpha}$ parameters.}
\label{fig:stat}
\end{figure}

Deriving theoretical finite-sample analysis may appear to be challenging for the proposed pseudo-metric. Thus, we numerically investigate the statistical convergence speed of $DR_{2,\varepsilon}$. To that end, we simulate two samples $\mathbf{X}$ and $\mathbf{Y}$ from two standard Gaussian distributions in dimension two with varying sample sizes. We compute the $DR_{2,\varepsilon}$ between $\mathbf{X}$ and $\mathbf{Y}$ with $n_{\alpha}\in \{ 5, 20, 100\}$ using the halfspace and the projection depths. Our proposed metric is computed with a high number of directions $K=10000$ to isolate the statistical error. We report the estimation error (averaged over 10 runs, the true value of $DR_{2,\varepsilon}$ being equal to zero) in Figure~\ref{fig:stat}. The experiment suggests that the statistical rates should be in $O(n^{1/4})$.

\subsection{The influence of the parameter $\varepsilon$}\label{sec:app:eps}

\begin{figure}[!h]
\begin{center}
\begin{tabular}{cc}
\includegraphics[trim=0cm 0cm 0cm 0cm, scale=0.30]{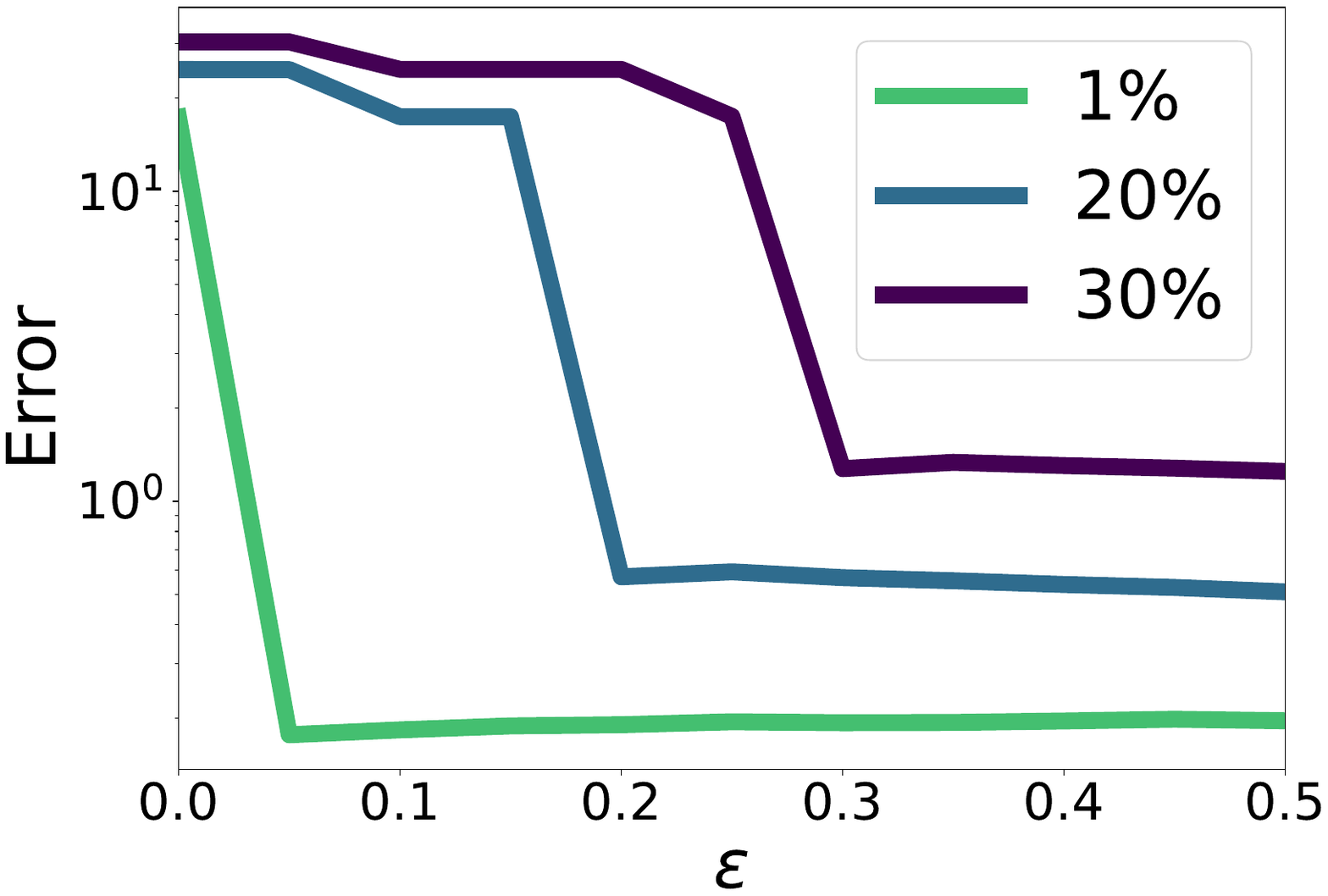} &\hspace{2cm}
\includegraphics[trim=10cm 0cm 0cm 0cm, scale=0.30]{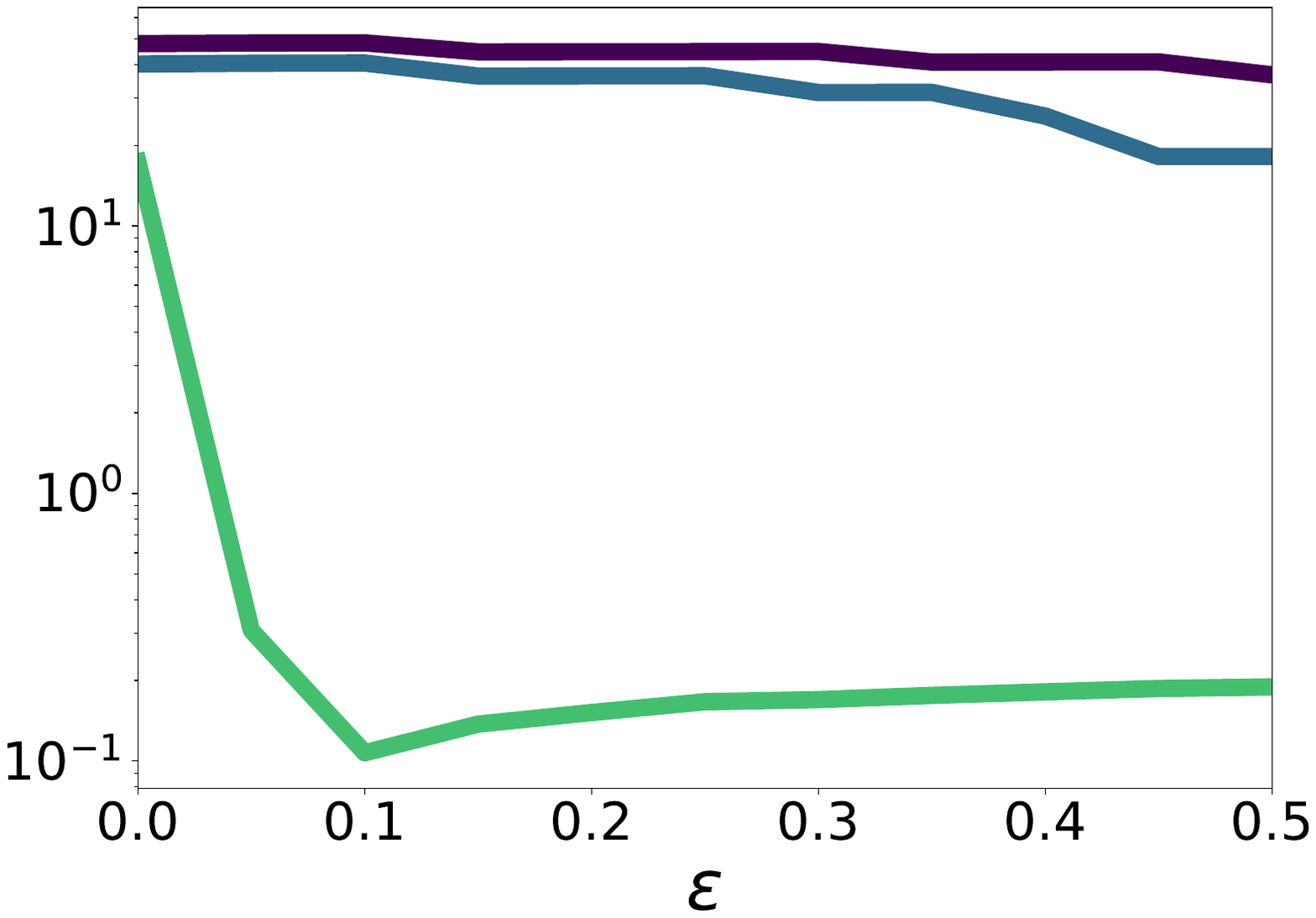}
\end{tabular}
\end{center}
\caption{Influence of the parameter $\varepsilon$ on the robustness of the proposed pseudo-metric with a robust depth function (the projection depth, left) and a non-robust one (the halfspace depth, right) for various proportion of anomalies.}
\label{fig:eps}
\end{figure}

The parameter $\varepsilon$ plays the role of the robust tuning parameter of $DR_{2,\varepsilon}$. In this part, we complete our theoretical results provided in Section~\ref{sec:robust}. We assess the robustness of our pseudo-metric making varying the parameter $\varepsilon$. Precisely, we simulate two normal samples $\mathbf{X}$ and $\mathbf{Y}$ from two standard Gaussian distributions in dimension two with a sample size of $10000$. From that, we construct abnormal samples with a proportion of anomalies equal to $\{1\%, 10\%, 20\%  \}$. To that end, we choose a proportion of normal samples and replace their first (for $\mathbf{X}$) and second (for $\mathbf{Y}$) coordinates as follows: $X_{\text{anom}}=30+50Z $ and  $Y_{\text{anom}}=-30-50Z $  where $Z$ follows a uniform distribution on $[0,1]$; leading to points far from the normal distributions. Thus, we compute $DR_{2,\varepsilon}$ with both robust and non-robust data depths, i.e. the projection and halfspace depths between $\mathbf{X}$ and $\mathbf{Y}$ being used as a benchmark. Further, we compute  $DR_{2,\varepsilon}$ between abnormal samples and report mean error (comparing values obtained between normal samples and values obtained between abnormal samples; averaged over ten runs) on Figure~\ref{fig:eps}. First, when computing with a robust depth function, we can see that the robustness of the proposed pseudo-metric relies directly on the parameter $\varepsilon$. This is shown by the presence of an elbow when the parameter $\varepsilon$ reaches the level of the proportion of anomalies. In contrast, we can see that for a non-robust depth function such as the halfspace depth, our proposed pseudo-metric becomes non-robust once the abnormal proportion is higher than $1\%$, leading to a poorly robust depth. This experiment then confirms our theoretical results on the Breakdown Point of $DR_{p,\varepsilon}$ displayed in Propostion~\ref{Breakdown}. The parameter $\varepsilon$ provides robustness to our pseudo-metric when combined with a robust depth function.

\subsection{The choice of the parameter $n_{\alpha}$}
\begin{figure}[!h]
\begin{center}
\begin{tabular}{ccc}
\includegraphics[trim=2cm 0cm 0cm 0cm, scale=0.20]{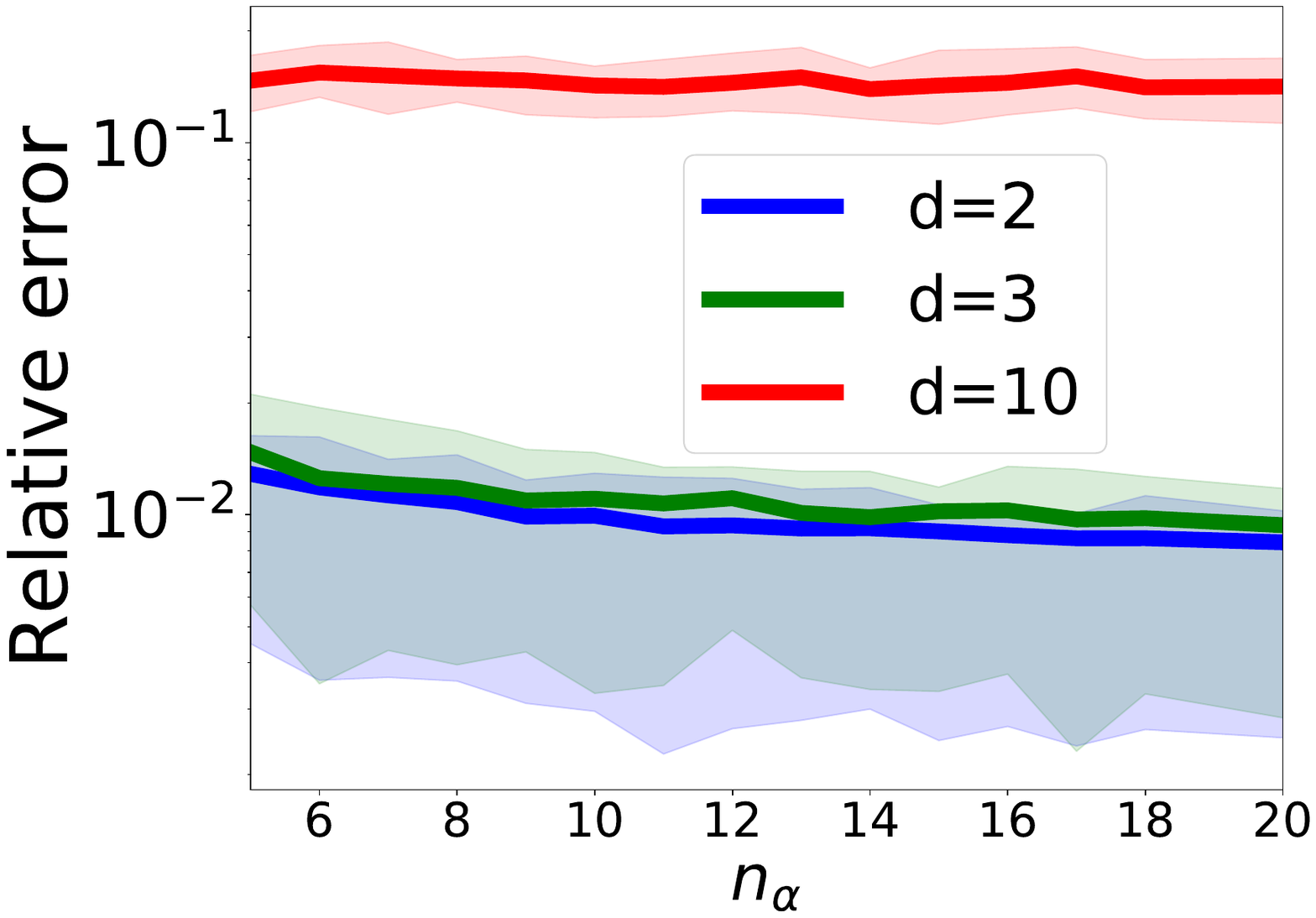}&\includegraphics[trim=2.5cm 0cm 0cm 0cm, scale=0.20]{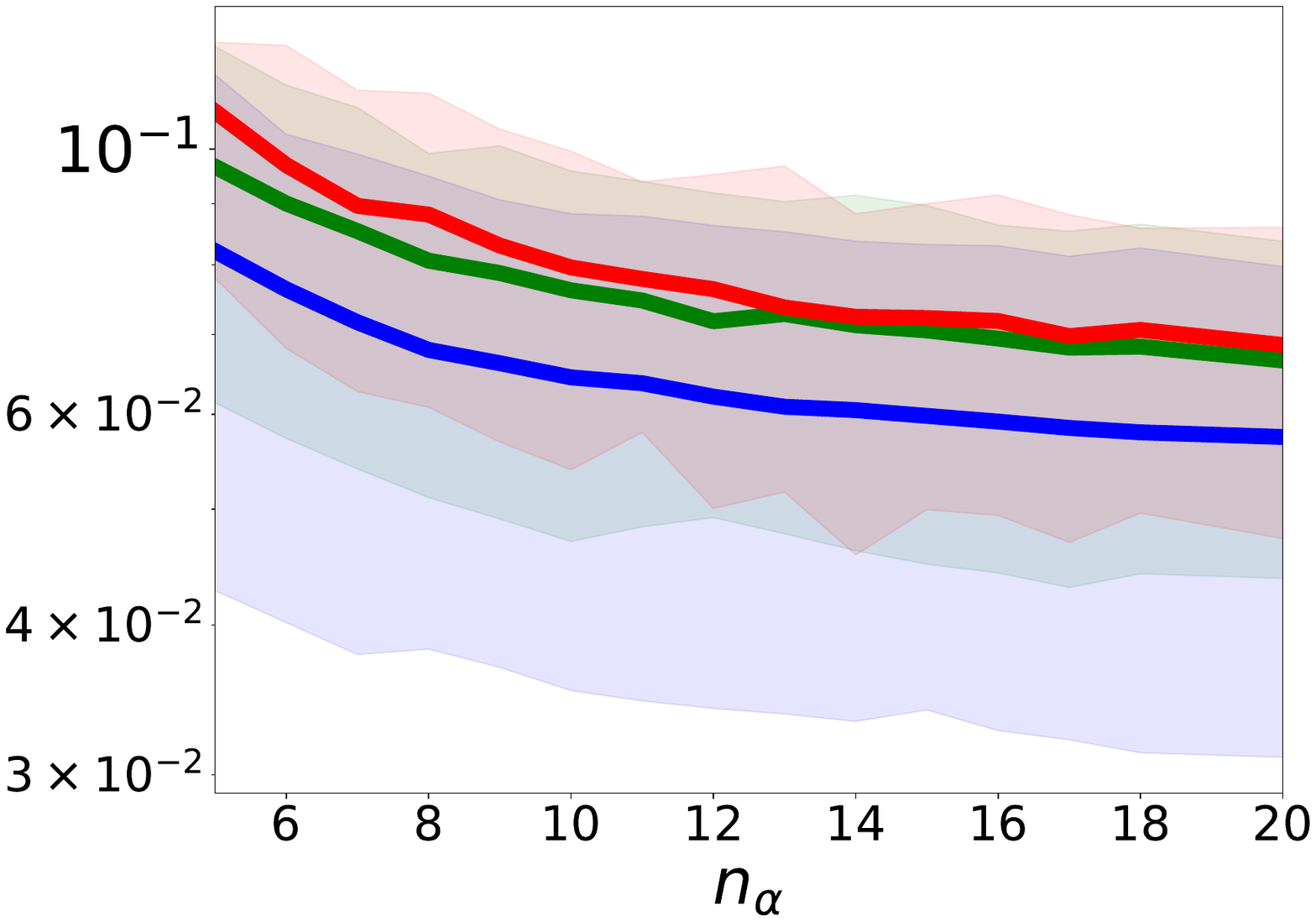}   & \includegraphics[trim=2.61cm 0cm 0cm 0cm, scale=0.20]{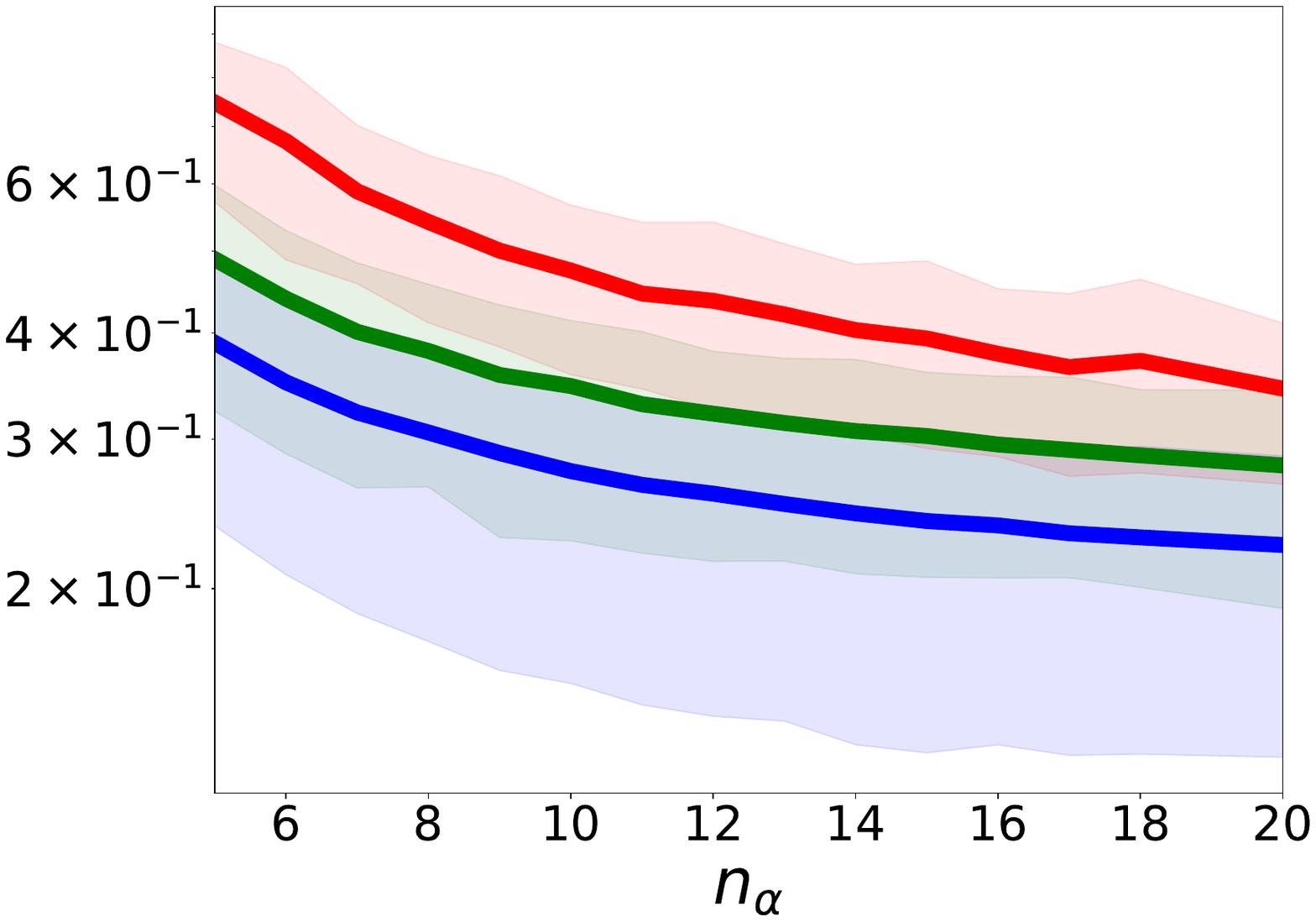}
\end{tabular}
\end{center}
\caption{Relative approximation error (averaged over $100$ repetitions, y-axis in log scale) of $DR_{p,\varepsilon}$ for elliptically symmetric \emph{Cauchy} (left), \emph{Student}-$t_2$ (middle) and \emph{Gaussian} (right) distributions for differing numbers of $n_{\alpha}$.}
\label{fig:parameter}
\end{figure}

Proposition \ref{general} allows to derive a closed form expression for $DR_{2,\varepsilon}(\mu,\nu)$ when $\mu,\nu$ are Gaussian distributions with the same variance-covariance matrix. 
In order to investigate the quality of the approximation on light-tailed and heavy-tailed distributions, we focus on computing $DR_{2,0.1}$ (with $K=500$) for varying number of $n_{\alpha}$ between a sample of 1000 points stemming from $\mu \sim \mathcal{N}(\mathbf{0}_d,\Sigma)$ for $d\in \{2,3,10 \}$, $\Sigma$ drawn from the Wishart distribution (with parameters ($d,I_d$)) on the space of definite matrices and three different samples (which yields nine settings). These three samples are constructed from 1000 observations stemming from elliptically symmetric \emph{Cauchy}, \emph{Student}-$t_2$ and \emph{Gaussian} distributions all centered at $\mathbf{7}_d$. Results that report the averaged approximation error and the 25-75\% empirical quantile intervals are depicted in Figure \ref{fig:parameter}. They show that $DR_{p,\varepsilon}$ converges slowly  for \emph{Cauchy}  with growing $n_{\alpha}$, while it converges with small $n_{\alpha}$ for  \emph{Gaussian}  and \emph{Student}-$t_2$ distributions.
 
\subsection{Robustness to outliers}

\begin{figure}[!h]
\begin{center}
\begin{tabular}{cc}
\includegraphics[trim=0cm 0cm 0cm 0cm, scale=0.20]{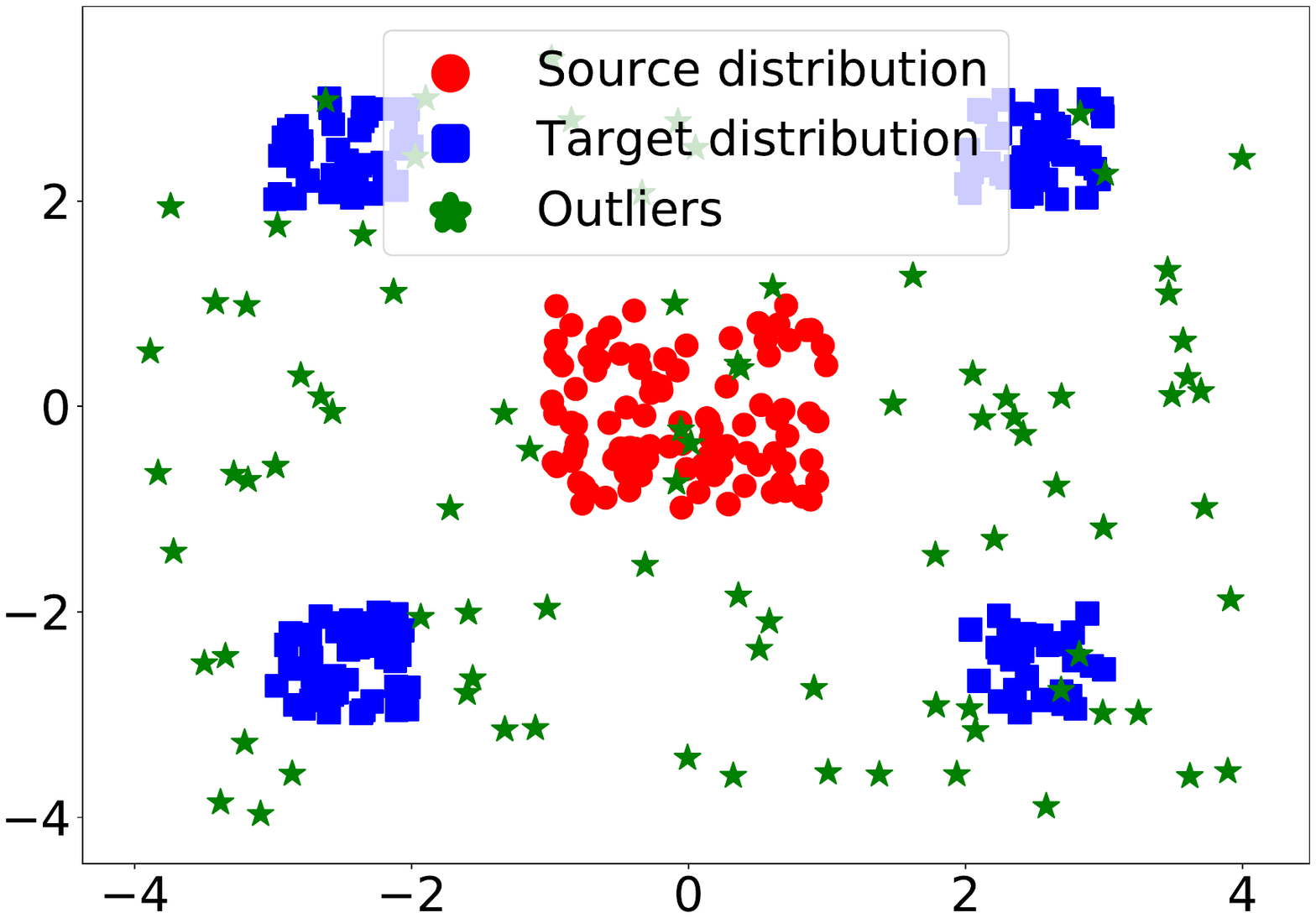} &\hspace{2cm}
\includegraphics[trim=0cm 0cm 0cm 0cm, scale=0.20]{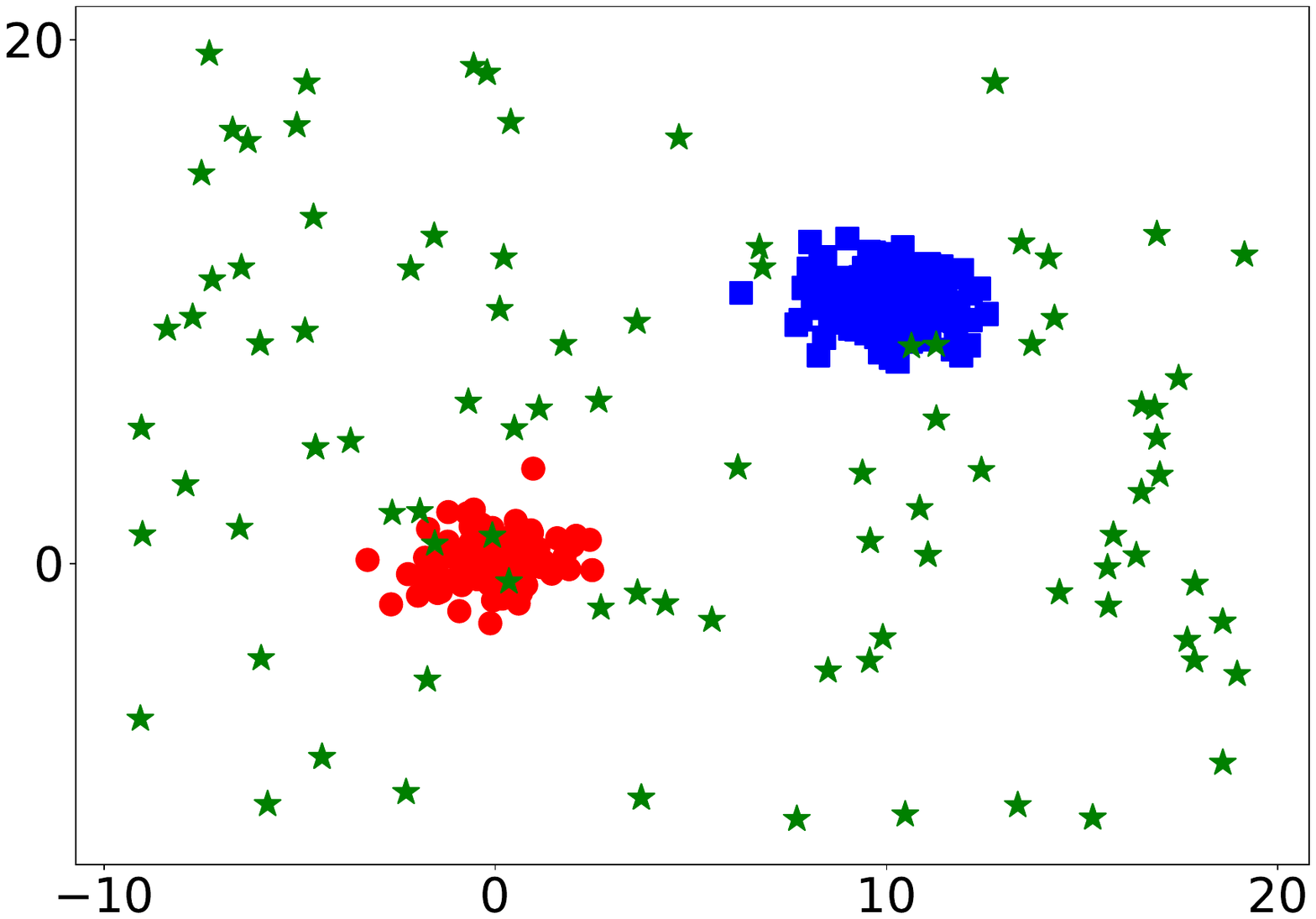} 
\end{tabular}
\end{center}
\caption{datasets related to robustness experiments depicted in Section~\ref{NUM} with $20\% $ of outliers for \emph{fragmented hypercube} (left) and \emph{Gaussian} (right).}
\label{robustness-datasets}
\end{figure}

Datasets on which experiments regarding "Robustness to outliers" in Section~\ref{NUM} have been performed are displayed in Figure~\ref{robustness-datasets}.

\section{APPLICATIONS TO NLP}\label{sec:suplementary_nlp}
In this section, we gather details on experimental settings and additional results on the automatic evaluation of natural language generation (NLG). 
\subsection{Extended related works on automatic evaluation of NLG}
Many metrics have been recently introduced for the automatic evaluation of text generation. In this work, we rely on untrained metrics. These metrics can be grouped into two categories: string-based metrics that depend on the string representation of the input texts to compute the similarity score and embedding-based metrics that rely on a continuous representation of the texts.

String matching metrics can be divided into two categories: N-gram matching and edit distance-based metrics. Perhaps the most used N-gram matching metrics are {BLEU}, {ROUGE} and {METEOR}. Edit distance-based metrics (e.g. TER;  \citealp{snover2006study}) measure the distance as the number of basic operations such as ‘edit'/‘delete'/‘insert'. Variants of {TER} include  CHARACTERE \citep{wang2016character}, CDER \citep{leusch-etal-2006-cder}, {EED} \citep{stanchev2019eed}. String-based metrics fail to produce meaningful scores in the case of paraphrases, especially if no common n-grams are found between the candidate and the reference text.

The second category of untrained metrics (namely embedding-based metrics) achieves state-of-the-art performance in many NLG evaluation tasks and has been introduced to address the issues mentioned above. Originally introduced for the widely used words embedding \citep{garcia2019token,colombo2019affect,colombo2020guiding,colombo2021beam} such as  Word2Vec \citep{word2vec} or Glove \citep{pennington2014glove}, this class of metrics has leveraged recently introduced contextualized word representations (CWR). CWR such as BERT, ELMO \citep{elmo}, HILAMOD \citep{chapuis2020hierarchical,chapuis2021code} or ROBERTA \citep{liu2019roberta} are popular in NLP \citep{witon2018disney} as they achieve SOTA performance on many tasks. The two most popular metrics are MoverScore and BertScore.
\subsection{Evaluation} 
For the task of evaluation of text generation, we assume that we have access to a dataset $ \{T_{R_i}, \{T_{G_i}^{j},h(T_{G_i}^{j})\}_{j=1}^{n_S} \}_{i=1}^{n_T}$ where $T_{G_i}^j$ represents the $i$-th  generated text by the $j$-th natural generation system, and $h(T_{G_i}^j)$ represents score assigned by the human annotator\footnote{In practice an averaged score is considered as each sentence is annotated by 3 different annotators. The considered datasets directly provide the aggregated score.} to $T_{G_i}^j$, and $T_{R_i}$ is the reference text. $n_T$ is the number of available texts, and $n_S$ is the number of different systems. \\

To assess the relevance of
an evaluation metric $\mathfrak{M}$, the correlation with the human judgment is considered one of the most important criteria \citep{banerjee2005meteor,koehn2009statistical,chatzikoumi2020evaluate}. To measure this correlation, two evaluation strategies are commonly adopted and built on top of a classical correlation measure, denoted $C$, e.g. Kendall ($\tau$; \citealp{kendall1938new}),  Pearson ($r$; \citealp{leusch2003novel}) or Spearman ($\rho$; \citealp{melamed2003precision}).
\begin{itemize}
    \item \emph{The text level correlation} ($C_{text}$) measures the ability of the metric to distinguish between badly and well generated text. Formally, $C_{text}$ is defined as follows:
    \begin{align} \label{se}
    C_{text} =&  \frac{1}{N_T} \sum_{i=1}^{n_T} C ( \mathbf{M}^{text}_i, \mathbf{H}^{text}_i ), \\
    \mathbf{M}^{text}_i =& \big{[}\mathfrak{M}(T_{R_i},T_{C_i}^1),\cdots,\mathfrak{M}(T_{R_i},T_{C_i}^{n_S})\big{]} ,\nonumber \\
    \mathbf{H}^{text}_i =& \big{[}{h}(T_{C_i}^1),\cdots,{h}(T_{C_i}^{n_S})\big{]}.\nonumber
\end{align}
    
    \item \emph{The system level correlation} ($C_{sys}$) assesses the ability of a metric to distinguish between good and bad systems. Formally, $C_{sys}$ is defined as follows:
    \begin{align}\label{sy}
    C_{sys} =& C(\mathbf{M}^{sys}, \mathbf{H}^{sys}),\\
    \mathbf{M}^{sys} =& \bigg{[}\frac{1}{{n_T}}\sum_{i=1}^{n_T} \mathfrak{M}(T_{R_i},T_{C_i}^1),\cdots, \frac{1}{{n_T}}\sum_{i=1}^{n_T} \mathfrak{M}(T_{R_i},T_{C_i}^{n_S})\bigg{]},\nonumber
    \\
    \mathbf{H}^{sys} =& \bigg{[}\frac{1}{{n_T}}\sum_{i=1}^{n_T} {h}(T_{C_i}^1),\cdots,\frac{1}{{n_T}}\sum_{i=1}^{n_T} {h}(T_{C_i}^{n_S})\bigg{]},\nonumber
\end{align}
\end{itemize}

We refer the reader to \citet{bhandari2020re} for further details on the evaluation of text generation.

\subsection{Results on Data2text}

In this section, we gather further details and results on data2text automatic evaluation.

\subsubsection{Task description}
In WebNLG 2020, the goal is to create new efficient generation algorithms that can verbalise knowledge-based fragments. These algorithms are called Knowledge Base Verbalizers \citep{gardent2017creating} and are used during the micro-planning phase of NLG systems \citep{ferreira2018enriching}. WebNLG has been gathered to be more representative of the progress of recent NLG systems than previously existing task-oriented dialogue datasets (see e.g. SFHOTEL \citep{wen2015semantically} and BAGEL \citep{mairesse2010phrase}). As previously mentioned for the data2text task we work on the WebNLG2020 challenge \citep{gardent2017creating,perez2016building}. Data and system performances can be found in \url{https://webnlg-challenge.loria.fr/}. The task consists in mapping RDF triples to natural language (RDF format is used for many application including FOAF (see \url{http://www.foaf-project.org/}). For WebNLG 2020, the triplets are extracted from DBpedia \citep{auer2007dbpedia}. Data have been made freely available from the authors at \url{https://gitlab.com/shimorina/webnlg-dataset/-/tree/master/release_v3.0}. To compose this dataset, 15 systems (both symbolic and neural-based) have been used. The final dataset is composed of over 3k samples of human annotations (see \url{https://webnlg-challenge.loria.fr/files/WebNLG-2020-Presentation.pdf} for more details).\\

\textbf{Example}: Given the following triplet \texttt{(John\_Blaha birthDate 1942\_08\_26) (John\_Blaha birthPlace San\_Antonio) (John\_Blaha job Pilot)} the ground-truth reference is \texttt{John Blaha, born in San Antonio on 1942-08-26, worked as a pilot}. 

\subsubsection{Results} We gather in \Cref{tab:full_web_nlg_sys} complete results on the WebNLG tasks including results on ROUGE-2. To compare $DR_{p,\varepsilon}$ (with $\varepsilon=0.01$, $n_\alpha=5$, $p=2$) with the different metrics (i.e.  Wasserstein, Sliced-Wasserstein, MMD), we work on Roberta-based model from the HuggingFace hub \citep{wolf2019huggingface} and extract representation from the 11th layer. From \Cref{tab:full_web_nlg_sys}, we observe a similar behavior from BertScore and MoverScore. This similarity has also been reported in a different setting in the previous work of \citet{zhao2019moverscore}. Overall, we observe that $DR_{p,\varepsilon}$ is always among its group's top-scoring metrics and achieves the best overall results on several configurations. It is worth noticing that $DR_{p,\varepsilon}$ only relies on information available in the candidate and the reference text. In contrast, BertScore and MoverScore use IDF information computed on every dataset.

\begin{table}
\centering
\begin{tabular}{lrrr|rrr|rrr}\hline  
  &  \multicolumn{3}{c}{Correctness} & \multicolumn{3}{c}{Data Coverage} & \multicolumn{3}{c}{Relevance} \\
 \cmidrule(lr){2-4} \cmidrule(lr){5-7} \cmidrule(lr){8-10}  &  $r$  &  $\tau$ &  $\rho$    &  $r$ &  $\tau$ &  $\rho$    &  $r$  & $\tau$ &  $\rho$    \\
  $DR_{p,\varepsilon}$& \textbf{89.4}  &  \textbf{80.0} & \textbf{92.6} & \textbf{84.2}  &   \underline{58.3} & \underline{72.3} & \textbf{86.2} & \underline{62.7} & \underline{72.9} \\
   Wasserstein& 86.2 &  73.0 & 86.7   & 80.4 &  45.3 & 62.3  & 83.8 & 51.3 & 67.6 \\
  Sliced-Wasserstein& 86.1  &  73.0 & 85.8 & 80.9  &  45.5 & 60.0& 82.0 &  51.3 & 68.2 \\
  MMD& 25.4  &  71.7 & 8.3 & 19.1  & 45.3 & 10.0& 26.1 &  51.3 & 15.0 \\\hline
BertScore &      \underline{85.5} & \underline{73.3} &       83.4 &      
74.7 & \underline{53.3} &     \underline{68.2} &   
\underline{83.3}& \underline{65.0}  &     \textbf{79.4}    \\       
MoverScore &      84.1 &  \underline{73.3} &      \underline{84.1} &         
\underline{78.7} &   \underline{53.3}  &   66.2 &    
82.1 &   \underline{65.0} &   77.4  \\\hline  
BLEU &      77.6 &   60.0 &     66.3 &       
55.7 &  36.6 &    50.2 &      
63.0 &  51.6 &    65.2    \\
ROUGE-1 &      80.6 &  65.0&     65.0 &      
76.5 &  \textbf{60.3} &    \textbf{76.3} &     
64.3 &  56.7 &    {69.2}   \\
ROUGE-2 &      73.6 &   58.3 &     63.3 &          54.7 &     35.0 &  43.1 &     
62.0 &  46.7 &    60.8   \\
METEOR &      \underline{86.5} &  \underline{70.0} &     {66.3} &     
\underline{77.3} &  46.6 &    50.2 &   
\underline{82.1} &  58.6 &    65.2   \\
TER &      79.6 &  58.0 &      \underline{78.3} &        69.7 & 38.0 &     58.2 &        75.0 &   \textbf{77.6} &   \underline{70.2}  \\
\bottomrule\end{tabular}
\caption{WebNLG 2020 (full results): absolute correlation at the system level with three human judgment criteria. Best overall results are indicated in bold, best results in their group are underlined.}\label{tab:full_web_nlg_sys}
\end{table}

\subsection{Results on summarization}\label{ssec:suplementary_summarization}
In this section, we gather experimental details and results on the automatic evaluation of the text summarization task.
\subsubsection{Task description}
Text summarization has attracted much attention in recent years \citep{zhang2020pegasus}. Two types of models exist: \emph{extractive} and \emph{abstractive}. In extractive summarization, the system copies chunks of informative fragments from the input texts, whereas, in abstractive summarization, the system generates novel words.  In this section, we describe our experimental setting. We present the tasks and the baseline metrics used for the automatic evaluation of summarization. We work with the dataset from \citet{bhandari2020re} for this task. This dataset has been introduced to solve several flaws \citep{rankel2013decade} present in existing summarization datasets such as TAC \citep{dang2008overview,mcnamee2009overview}. The dataset has been annotated using the pyramid score \citep{nenkova2007pyramid,nenkova2004evaluating} and automatically built from the CNN/Daily News \citep{bhandari2020re}. It gathers 11\,490 summaries coming from  11 extractive systems \citep{see2017get,chen2018fast,raffel2019exploring,gehrmann-etal-2018-bottom,dong2019unified,liu2019text,lewis2019bart,yoon2020learning} and 14 abstractive systems \citep{zhou2018neural,narayan2018ranking,kedzie2018content,zhong2019searching,liu2019text,dong2019unified,wang2020heterogeneous,zhong2020extractive}. 
\\

\textbf{Example}: The goal is to assign a similarity score between a reference text: 
``\emph{Manchester United take on Manchester City on Sunday. Match will begin at 4 pm local time at United's Old Trafford home. Police have no objections to kick-off being so late in the afternoon. Last late afternoon weekend kick-off in the Manchester derby saw 34 fans arrested at Wembley in 2011 fa cup semi-final}'' and the text generated by a NLG system: ``\emph{Manchester Derby takes place at Old Trafford on Sunday afternoon police have no objections to the late afternoon kick-off both sides are challenging for a top-four spot in the Premier League the man in charge of patrolling the sell-out clash has no such fears}''.

\subsubsection{Results} We gather in \Cref{tab:summarization_results}, the results on the summarization task. We use a bert-based uncased model and rely on the representations extracted from the 9th layer (similarly to BertScore). For this experiment the following parameters are used: $\varepsilon=0.01$, $n_\alpha=5$, $p=2$. For this task, we can reproduce results from \citet{bhandari2020re} where the different behavior regarding the extractive and the abstractive systems is also observed. In this experiment, we observe that $DR_{p,\varepsilon}$ can achieve stronger results than other metrics based on Wasserstein, Sliced-Wasserstein and MMD. We also observe that $DR_{p,\varepsilon}$  outperforms MoverScore and BertScore on extractive systems (on $r$ and $\tau$). We believe these results support our approach.

\begin{table}[t!]
\centering

\begin{tabular}{cccc|ccc}\hline
    & \multicolumn{3}{c}{Abstractive} & \multicolumn{3}{c}{Extractive} \\
\cmidrule(lr){2-4} \cmidrule(lr){5-7} & $r$ & $\tau$  & $\rho$		   & $r$ & $\tau$ & $\rho$	 \\
  $DR_{p,\varepsilon}$&\underline{72.1} & \underline{72.1}  & 70.1 & \underline{91.5} & \textbf{91.5}  & \underline{69.2}\\
   Wasserstein& 71.0 & 70.4  & \underline{71.1} & 74.2 & 74.2  & 40.0 \\
  Sliced-Wasserstein& 70.1 & 68.7   & 71.0& 72.4 & 73.9  & \underline{69.2} \\
  MMD& 68.2 & 67.5  & 67.9 &  75.6 & 75.6  & 56.1 \\\hline
 BertScore &	71.7 &	\underline{71.9}  & 72.0 &	70.9&	72.9	  & \textbf{73.8}  \\
MoverScore  &	\underline{72.4} &	\underline{71.9}  & \underline{73.0}  & \underline{76.1} &	\underline{76.1}   & 47.4  \\\hline
ROUGE-1 &	\textbf{73.5} &	73.0    & \textbf{74.4} & 	72.2&	\underline{74.0}	  & \underline{69.1}  \\
ROUGE-2	&	73.0 &	\textbf{73.5}  & 73.0 &	55.1&	53.2	  & \underline{69.1} \\
\texttt{JS-2} & 68.9 &6.8  & 69.8 &	\textbf{92.9} & 5.5    & 19.0\\
\hline
    \end{tabular}
    \caption{Summarization: absolute correlation coefficients (using Pearson ($r$), Spearman ($\rho$) and Kendall ($\tau$) coefficient) between different metrics on text summarization. Best overall results are indicated in bold, best results in their group are underlined.}
    \label{tab:summarization_results}
\end{table}

\end{document}